\documentclass[final]{anthology-ch} 

\usepackage{booktabs}
\usepackage{graphicx}
\usepackage{array}
\usepackage{caption}
\usepackage{wasysym}
\usepackage{comment}
\usepackage{float} 
\usepackage{comment}
\usepackage{amsmath}    
\usepackage{amssymb}

\title{Computing the Formal and Institutional Boundaries of Contemporary Genre and Literary Fiction}

\author[]{Natasha Johnson}[
  orcid=0009-0001-7388-9053
]

\affiliation{}{Language Technologies Institute, Carnegie Mellon University, Pittsburgh, USA}

\keywords{computational literary studies, genre, gender, literary prestige}

\pubyear{2025}
\pubvolume{1}
\pagestart{1}
\pageend{1}
\conferencename{Proceedings of Conference XXX}
\conferenceeditors{Editor1 Editor2}
\doi{00000/00000}  

\begin{document}

\maketitle

\begin{abstract}
Though the concept of genre has been a subject of discussion for millennia, the relatively recent emergence of genre fiction has added a new layer to this ongoing conversation. While more traditional perspectives on genre have emphasized form, contemporary scholarship has invoked both formal and institutional characteristics in its taxonomy of genre, genre fiction, and literary fiction. This project uses computational methods to explore the soundness of genre as a formal designation as opposed to an institutional one. Pulling from Andrew Piper’s CONLIT dataset of Contemporary Literature, we assemble a corpus of literary and genre fiction, with the latter category containing romance, mystery, and science fiction novels. We use Welch's ANOVA to compare the distribution of narrative features according to author gender within each genre and within genre versus literary fiction. Then, we use logistic regression to model the effect that each feature has on literary classification and to measure how author gender moderates these effects. Finally, we analyze stylistic and semantic vector representations of our genre categories to understand the importance of form and content in literary classification. This project finds statistically significant formal markers of each literary category and illustrates how female authorship narrows and blurs the target for achieving literary status.
\end{abstract}

\section{Introduction} 

Scholars have been using, debating, and reinventing the term “genre” for millennia. In fact, in 1976, Todorov admitted that discussing genre might be considered “an obviously anachronistic pastime” \cite{duff-2000}. However, the emergence of genre fiction 
over the past century has added new layers to this age-old conversation.

When scholars such as Frye, Jameson, and Rosen have considered form alongside genre, they have touched on features including narrative structure, character, and the presence or lack of “elevated” or “non-elevated” language. Scholars emphasizing the institutional nature of genre have highlighted its relation to author gender and literary distribution. This project uses computational methods to further examine the formal character of genre and literary fiction and the impact that author gender has on these boundaries. Through a distant reading of Andrew Piper’s CONLIT corpus of contemporary fiction, we seek to gain a deeper and more empirical understanding of the formal boundaries of contemporary genre fiction; the distinctions between genre, genre fiction, and literary fiction; and how this varies by author gender. 

We begin by using Welch's ANOVA to examine narrative structure, character, and elevated language within and between genre and literary fiction. Then, we use logistic regression to explore how author gender affects the role that these features play in literary classification. Finally, we use distance-based statistical tests to measure the stylistic and semantic distance between our categories.

\section{Related Work}

\subsection {Debates on Genre}
Northrop Frye is one of the earliest literary scholars whose writings on genre are still frequently referenced in current criticism. In Anatomy of Criticism \cite{frye-1957}, Frye presents a string of classically inspired theories about mode, symbolism, myth, and genre. His book is very abstract and distant from traditional literary scholarship, but his identification of four fundamental plot structures has been revisited by scholars such as Hayden White \cite{white-1971} and Kurt Vonnegut \cite{vonnegut-2004}. 

Fredric Jameson \cite{jameson-1975} builds upon Frye’s work in “Magical Narratives: Romance as Genre,” exploring genre both as a formal designation and as a broader social framework. Jameson proposes that genres are contracts between writers and readers that establish expectations regarding the content and form of novels and also provide a means of categorizing books. He suggests two approaches to the practice of contemporary genre: the syntactic, which explores the form of genre, and the semantic, which explores the meaning of genre within society at large.

Jeremy Rosen \cite{rosen-2019} also engages with both form and society in his discussion of genre, but he first makes the key contribution of distinguishing between “genre,” “genre fiction,” and “literary fiction.” He defines genre as “an existing literary framework or recipe that writers may adapt according to their needs,” which touches upon formal elements similar to Frye’s work and Jameson’s syntactic approach.  Rosen suggests that authors of literary fiction make use of the same genre conventions that popular genre fiction follows, but also clearly try to create distance between their work and popular genre fiction through methods such as using elevated literary devices or critiquing popular culture within the text itself. Despite these formal practices, Rosen argues that literary fiction and genre fiction should be thought of primarily as “relational subfields of production, circulation, and reception," as opposed to formal designations. Rosen's work complements Ken Gelder's stance on genre and genre fiction. Gelder \cite{gelder-2004} suggests that genre not only provides a formal framework for genre fiction, but that it furthermore determines a book's publishing, marketing, and consumption. Gelder emphasizes the social and industrial characteristics of genre fiction by referring to it as ``popular fiction."

Mark McGurl \cite{mcgurl-2021} similarly challenges the validity of formal distinctions between genre and literary fiction by, in Goldstone’s words, making high-literary assumptions about genre and simultaneously deflating the pretensions of literary fiction \cite{goldstone-2023A}. However, McGurl also suggests that digital distribution of literature undermines once-strong institutional divides. Others such as Theodore Martin \cite{martin-2017}, Andrew Hoberek \cite{hoberek-2007}, and Marc Verboord \cite{verboord-2012} have also spoken of a narrowing gap between genre and literary fiction, but they have attributed this to a widespread elevation in the status of genre fiction and popular culture. 

\paragraph{Formal Elements of Genre and Literary Fiction.} Despite differing views on the importance of form in the classification of genre and literary fiction relative to institutional factors, scholars often attribute a consistent set of formal elements to literary fiction: elegant prose, novelty, emphasis on character over plot, complex structure, serious subject matter, and slow pacing \cite{rosen-2019, saricks-2009}. In contrast, genre fiction is often stated to be plot driven, highly standardized, fast-paced, and stylistically simple \cite{duff-2000, radway-1991}.

\paragraph{Institution and Genre.}
Scholarship discussing the institutional nature of genre and literary fiction has offered varied perspectives on the specific institutions at play. Some scholars have emphasized literary fiction's resistance to mass culture \cite{mcgurl-2001} and distance from the marketplace \cite{rosen-2019}. Others have centered differences in the publication, distribution, and consumption of literary and genre fiction \cite{rosen-2019}. Andrew Hoberek has cited gender as the primary force at play, stating that ``[g]ender continues to mark the outer limits of what counts as literary," with genre fiction often being associated with women's writing \cite{hoberek-2017}. Verboord \cite{verboord-2012} has also emphasized the status-based system which underlies the literary field and the systematic marginalization of women in fields of artistic production.

\subsection {Empirical Analyses}
A number of empirical studies have responded to theories posed above. Algee-Hewitt et al. \cite{algee-hewitt-2016} compared canonical works (often equated with literary fiction) to archival works (other works of fiction) from 1800 - 1900 and found the canonical to have a much higher entropy but lower type token ratio than archival works. In looking at reading difficulty within books from the Chigaco Corpus (from 1880 - 2000), Bizzoni et al. \cite{bizzoni-2023} found that there was a small but statistically significant difference in the reading difficulty of literary versus non-literary novels. When considering the same corpus, Lassen at al. \cite {lassen-2024} found that machine learning models exhibit male-favored gender bias when attempting to classify texts based on canonicity.

Anderson et al. \cite{anderson-2001} analyzed 604 documents from the British National Corpus, published before 2001, and found significant differences in the linguistic and stylistic footprints male and female authors---namely that women use more pronouns, whereas men use more noun specifiers; and women are more involved, whereas men are more informational. However, it is difficult to know whether these findings apply to contemporary fiction, because the British National Corpus contains both fiction and non-fiction, and Anderson et al. observed that the aforementioned female stylistic footprints are broadly correlated with fiction, whereas the male foorprints are associated with nonfiction.

Verboord \cite{verboord-2012} looked at US, French, and German literature between 1960 and 2009 and found that women are more represented on best seller lists than among literary award winners. In line with the work of Tuchman and Fortin and Radway, he discusses how the artistic products of women are held in lower artistic status than those of men \cite{radway-1991} and how men routinely enter previously women-dominated fields that are rising in prestige and edge women out \cite{tuchman-1984}. 

While each of these analyses offers insights into stylistic and institutional qualities of fiction, most only consider a limited range of stylistic or institutional features, without exploring ways in which these might interact. Furthermore, none of the studies focus on books written in the past two decades, after the widespread growth of the internet. If the aforementioned claims about waning genre divides are true, as supported by the empirical analyses of Sharma et al. \cite{sharma-2020}, examining fiction from the twenty-first century is necessary in order to understand where current literary and institutional boundaries lie. 
\section{Data}

The data we examine in this project comes from Piper’s CONLIT dataset of Contemporary Literature \cite{piper-2022}. This dataset includes data and metadata from 2,754 Anglophone books written between 2001 and 2021 inclusive. The corpus includes 234 mystery novels, 208 romance novels, and 223 science fiction novels, all hand-tagged. We remove the 4 books tagged as both science fiction and mystery and keep the remaining as our genre fiction books.

\begin{table}[h]
  \centering 
  \small
  \begin{tabular}{lc}
    \toprule
    \textbf{Category} & \textbf{Dataset Count}\\
    \midrule
    Science Fiction & 222 \\
    Mystery & 231 \\
    Romance & 208 \\
    Literary Fiction & 191 \\
    Genre Fiction & 661 \\
    \bottomrule
  \end{tabular}
  \caption{Book categories explored in this project and the total number of works present within our dataset for each category}
  \label{tab:counts}
\end{table}

While there are no books explicitly tagged as literary fiction, the “Prizelists” category includes 258 works of fiction that have won or been shortlisted for the Faulkner Award, National Book Award, Man Booker Prize, Giller Prize, and Governor General’s Award. The former four of these awards are given to works of high literary value, which we would colloquially refer to as literary fiction; in fact, Rosen cites the selection criteria for the Man Booker Prize as standards for literary fiction \cite{rosen-2019}. The Governor General’s Award does not make this distinction and has been given to works that would generally be considered genre fiction. We remove the books nominated for this award to produce a group of 191 novels, which we consider our literary fiction. The final book counts for this project can be seen in Table \ref{tab:counts}.

For each of the books in CONLIT, Piper has shared features such as the narrative speed, topical heterogeneity, and Tuldava score (a commonly used measure of reading difficulty). The features were identified using methods described by Toubia et al. \cite{toubia-2021}, Ciaramita and Altün \cite{ciaramita-2006}, and Bamman et al. \cite{bamman-2014}. The features we consider in this project are explained in Table \ref{tab:feature_descriptions}. Piper also shares the top 10,000 unigrams\footnote{In the context of this project, unigrams are equivalent to words.} that appear in at least 18\% of the books in the corpus, alongside their frequency counts within each book. We consider these as well.

\begin{table}
  \centering
  \small
  \begin{tabular}{ll}
    \toprule
    \textbf{Feature} & \textbf{Description} \\
    \midrule
    Average Speed & Measure of narrative pace \\
    Minimum Speed & Measure of narrative distance \\
    Circuitousness & Measure of narrative non-linearity \\
    Topical Heterogeneity & Measure of the semantic spread of book sections \\
    Event Count & Estimated number of diegetic events \\
    Total Characters & Total number of named characters appearing >5 times \\
    Protagonist Concentration & Percentage of all character mentions by main character \\
    Probability of First Person & 1 if first-person, 0 if third person\footnotemark \\
    Token Count & Number of tokens (words and punctuation) in a book \\
    Avg. Word Length & Average length of all words per book \\
    Avg. Sentence Length & Average length of all sentences per book \\
    Tuldava Score & Reading difficulty measure \\
    \bottomrule
  \end{tabular}
  \caption{Feature descriptions for narrative analysis} 
  \label{tab:feature_descriptions}
\end{table}
\footnotetext{For calculations involving the probability of 1st person (i.e. Welch ANOVA testing and regression modeling focusing on this feature) probability of 1st person is treated as a binary feature, and the 22 books with values in between 0 and 1 are excluded.}
\section{Methodology}
\subsection{Feature comparison with Welch's ANOVA}
We begin by measuring the significance of narrative features in genre classification. We take the mean of each feature within each category (science fiction, mystery, romance, literary fiction, and genre fiction) and use Welch's ANOVA \cite{moder-2010} to test whether these values are significantly different between the individual genres (science fiction, mystery, romance) and between genre and literary fiction. \footnote{For more information Welch's ANOVA and how our data meets its assumptions, see Appendix \ref{appdx:welch}.}

For clarity, we group the features into the categories of narrative structure, language, and character. Within narrative structure, we include narrative pace, distance, circuitousness, topical heterogeneity, and event count. Within language, we consider token count, average sentence length, average word length, Tuldava score, and the point of view (probability of 1st person). For character, we include character count and the protagonist concentration.

Then, we divide our corpus by author gender (excluding the 5 books with an unspecified author gender) and repeat the same tests for male and female authors. Book counts for male and female authors can be seen in Table \ref{tab:counts_by_gender}. Since only 2 of the books written by male authors are romance books, we exclude these books as well, as they do not constitute a large enough sample size to meet the normality assumption of Welch ANOVA (see Appendix \ref{appdx:welch}). In order to ensure that our observations about female-versus-male-authored books are not skewed by the presence of romance novels only in the former category, we run Welch ANOVA twice for the female authors---once with romance books included, and once with romance books excluded.

\begin{table}[h]
  \centering 
  \small
  \begin{tabular}{lcc}
    \toprule
    \textbf{Category} & \textbf{Female} & \textbf{Male} \\
    \midrule
    SF & 72 & 150\\
    MY & 117 & 110\\
    ROM & 205 & 2\\
    Lit & 80 & 111\\
    Genre & 394 & 262\\
    \midrule
    Total & 474 & 373 \\
    \bottomrule
  \end{tabular}
  \caption{Book counts within dataset by author gender}
  \label{tab:counts_by_gender}
\end{table}

\subsection{Modeling the impact of gender}
To further examine the impact of author gender, we measure whether author gender affects the effect that literary features have on a book's classification as genre or literary fiction. Using a subset of features from Table \ref{tab:feature_descriptions}, we fit a logistic regression model that incorporates the main effects of each feature and author gender, as well as the interaction affects of the features and author gender. Prior to modeling, we normalize each non-binary feature using z-score standardization. The model is represented by the following formula:

\[
\log\left(\frac{P(\text{Literary})}{1 - P(\text{Literary})}\right) = \beta_0 + \delta G + \sum_{i=1}^p \beta_i X_i + \sum_{i=1}^p \gamma_i (X_i \times G)
\]
where
\[
\begin{aligned}
& P(\text{Literary}) \text{ is the probability of literary classification}, \\
& X_i \text{ are the features}, \\
& G \text{ is the binary gender indicator (e.g., } 0 \text{ for Male, } 1 \text{ for Female)}, \\
& \beta_0 \text{ is the intercept}, \\
& \delta \text{ is the coefficient for gender}, \\
& \beta_i \text{ is the coefficient for feature } i, \\
& \gamma_i \text{ is the coefficient for the interaction between feature } i \text{ and gender}.
\end{aligned}
\]

To formally test whether the interaction term significantly improves the model, we fit a second model without the interaction term, represented by the following formula:

\[
\log\left(\frac{P(\text{Literary})}{1 - P(\text{Literary})}\right) = \beta_0 + \delta G + \sum_{i=1}^p \beta_i X_i
\]
We then compare both models using a likelihood ratio test \cite{newsom-2021}.

\paragraph{Feature Selection.}
Analysis of Variance Inflation Factors (VIFs) reveals high multicollinearity among our features (see Appendix \ref{appdx:logistic_output}). To help us determine which features to exclude, we fit a logistic regression model for each feature individually, represented by the following formula:
\[ \log \left( \frac{P(\text{Literary} = 1)}{1 - P(\text{Literary} = 1)} \right) = \beta_0 + \beta_i X_i + \delta G + \gamma_i (X_i \cdot G)
\] We eliminate those features with $p$ values > 0.05.\footnote{For results of univariate logistic regression, see Appendix \ref{appdx:logistic_output}.} The features without significant impact are total number of characters, narrative distance, and topical heterogeneity. Then, since average sentence length and Tuldava score still have VIFs > 10, we eliminate Tuldava score, which has the higher VIF of the two (due to its additional overlap with average word length). The remaining features we use for our multivariate logistic regression are protagonist concentration, average word length, average sentence length, token count, pace, circuitousness, point of view, and event count.

\paragraph{Probabilities.} To further illustrate how author gender affects the relationship between narrative feature and literary classification, we predict the probability of literary classification given author gender and a range of values for each selected feature. We calculate these probabilities by taking the sigmoid of the log-odds (logit) for each gender along a range of values for each feature \cite{li-2022}. When reporting these results, we convert the scaled feature values back to their original values for easier interpretation. 

\subsection{Measuring Stylistic and Semantic Distance with Unigrams}
Since our previous analyses rely on formal attributes of our texts, we use the books' unigrams to gauge whether the differences between our categories are primarily formal or semantic. We repeat a series of statistical tests measuring the distance between individual genres and genre versus literary fiction, focusing once on style and once on content.

\begin{figure}[t!]
  \centering
  \includegraphics[width=0.49\linewidth]{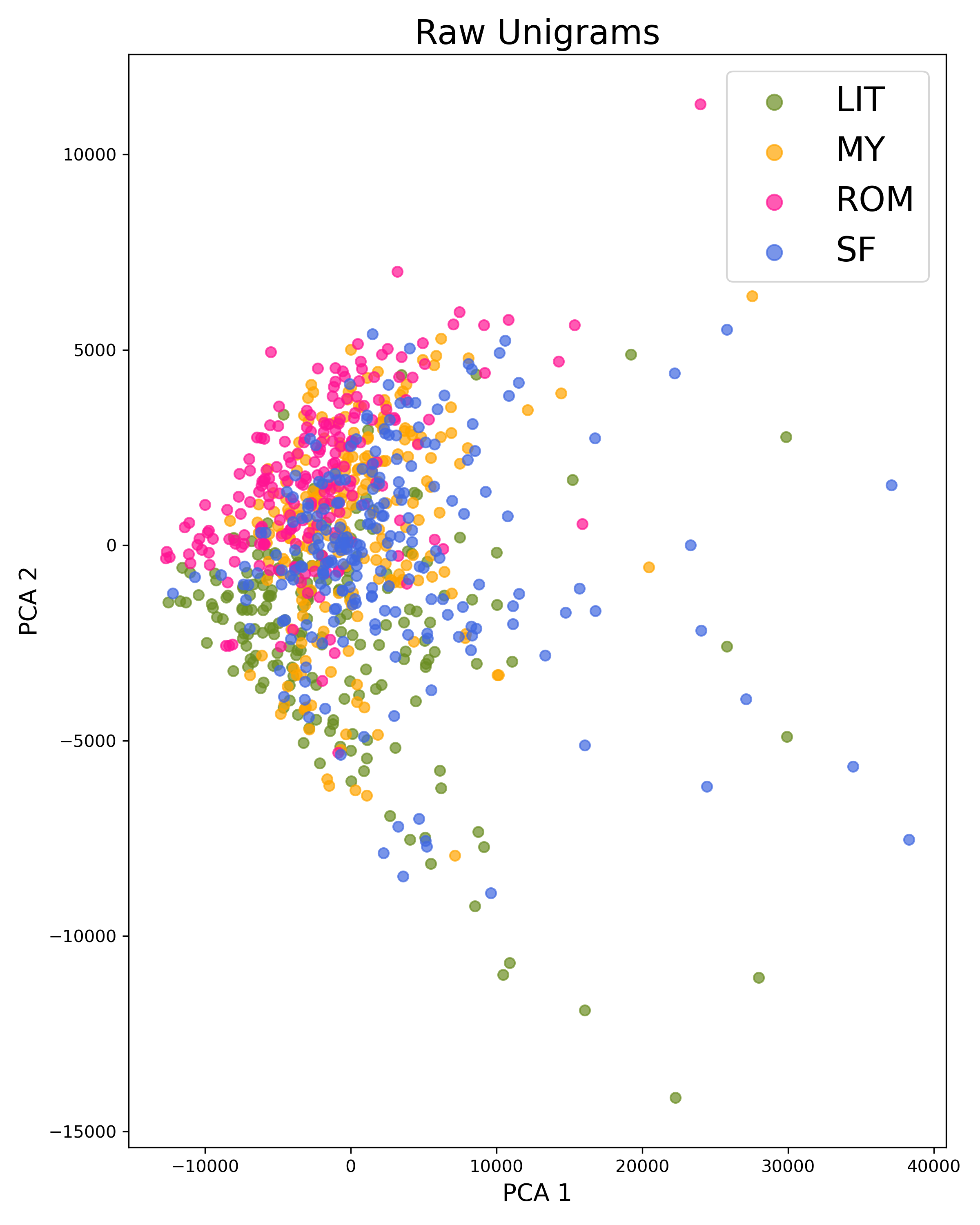}
  \includegraphics[width=0.49\linewidth]
  {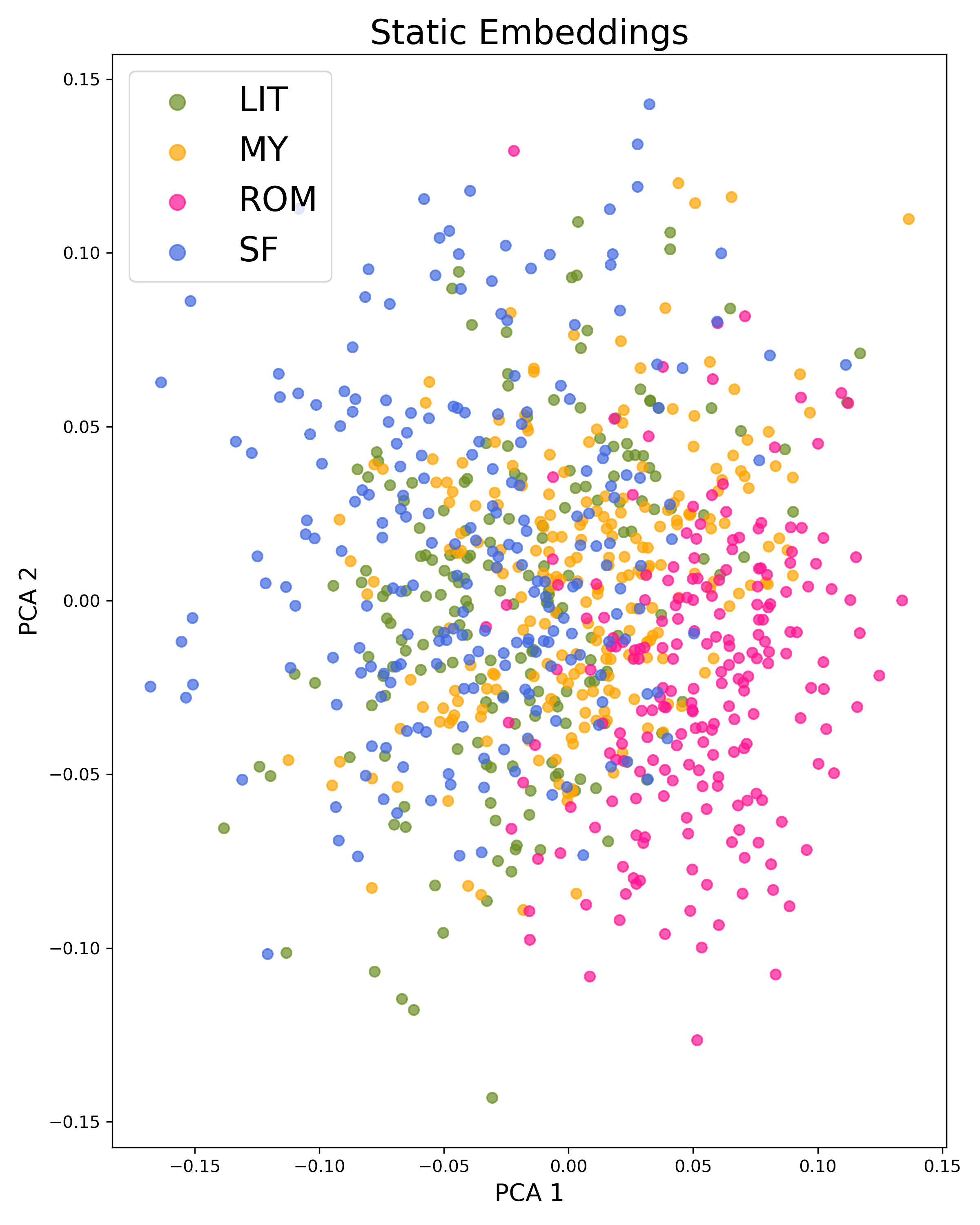}
    \caption{PCA projections of books based on raw-unigram (left) and static-embedding vector representations (right). Each point corresponds to a book. Color indicates genre category.}
  \label{fig:unigram_plots}
\end{figure}

When measuring stylistic differences, we use use the raw unigram counts as feature vectors. When measuring semantic distance, we construct feature vectors from static embeddings of the unigrams. Using fastText embeddings \cite{mikolov2018advances} trained on Common Crawl, we represent each book as the TF-IDF weighted-average of its unigram embeddings and then apply L2-normalization to account for differences in book lengths. We plot the raw-unigram and static-embedding representations of our corpus, applying PCA to reduce dimensionality (see Figure \ref{fig:unigram_plots}).


Before we can measure distances between categories, we must use PERMDISP test to measure the dispersion within each category and determine whether the categorical variances are too large for traditional PERMANOVA tests, which require homogeneity of dispersion \cite{anderson-2006}. Because our PERMDISP tests reveal heteroscedasticity between the raw unigrams of literary and genre fiction, we use $W^*_d$ and TwT tests---non-parametric, distance-based tests of group difference, which are robust against non-homogeneous dispersions \cite{hamidi-2019}---to measure the distances between our vectors in each category. The results of these tests can be seen in Tables \ref{tab:permdisp} and \ref{tab:wd}.

\section{Results}

\subsection{Feature Welch ANOVA Testing.}
Tables \ref{tab:anova_combined}, \ref{tab:anova_combined_female}, \ref{tab:anova_combined_female_no_rom}, and  \ref{tab:anova_combined_male} contains the means, $F$-statistics and effect sizes resulting from our Welch ANOVA tests.\footnote{For complete Welch ANOVA output, see Appendix \ref{appdx:anova_output}. For histograms illustrating ANOVA results, see Appendix \ref{appdx:histograms}.} \footnote{Note that when comparing Welch ANOVA results across different features and categories, effect sizes ($\eta^2_p$) provide a more meaningful comparison than $F$.} When looking at books written by both male and female authors, science fiction, romance, and mystery are significantly different ($p<0.05$) across all the features. Likewise, literary and genre fiction are significantly different in protagonist concentration, sentence length, word length, Tuldava score, number of events, circuitousness, narrative distance, and topical heterogeneity. They are not significantly different in their token counts, number of characters, narrative speed, and point of view. Splitting the dataset by author gender changes the significant features. 

\begin{table} [H]
  \centering
  \small
  \renewcommand{\arraystretch}{1.2}
  \setlength{\tabcolsep}{4pt} 
  \begin{tabular}{|>{\raggedright\arraybackslash}p{3.2cm}|c|c|c|c|c|c|c|c|c|c|c|}
    \hline
    & \multicolumn{5}{c|}{\textbf{Inter-Genre}} & \multicolumn{4}{c|}{\textbf{Genre vs Literary Fiction}} \\
    \hline
    \textbf{\shortstack[l]{Feature}} & \textbf{\shortstack{\\SF \\Mean}} & \textbf{\shortstack{\\MY \\Mean}} & \textbf{\shortstack{\\ROM \\Mean}} & $\boldsymbol{F}$  & $\boldsymbol{\eta^2_p}$ & \textbf{\shortstack{\\Genre \\Mean}} & \textbf{\shortstack{\\Lit \\Mean}} & $\boldsymbol{F}$ & $\boldsymbol{\eta^2_p}$ \\
    \hline
    Narrative Pace & 2.25 & 2.27 & 2.33 & 62.07* & 0.15 & 2.28 & 2.29 & 1.32 & 0.00 \\
    \hline
    Narrative Distance & 2.48 & 2.50 & 2.56 & 93.61* & 0.22 & 2.51 & 2.49 & 15.71* & 0.02 \\
    \hline
    Circuitousness & 0.23 & 0.24 & 0.23 & 4.98* & 0.02 & 0.23 & 0.20 & 38.25* & 0.06 \\
    \hline
    Topical Heterogeneity & 3.01 & 3.01 & 3.04 & 6.63* & 0.02 & 3.02 & 2.98 & 12.71* & 0.02 \\
    \hline
    Event Count & 7286 & 6659 & 6575 & 3.49* & 0.01 & 6844 & 5657 & 21.90* & 0.03 \\
    \hline
    Character Count & 52.96 & 48.27 & 27.89 & 113.14* & 0.15 & 43.43 & 49.25 & 3.51 & 0.01 \\
    \hline
    Protagonist Conc. & 0.24 & 0.23 & 0.31 & 34.03* & 0.09 & 0.26 & 0.21 & 25.64* & 0.03 \\
    \hline
    Written in 1st Person & 0.37 & 0.51 & 0.61 & 12.91* & 0.04 & 0.49 & 0.57 & 3.63 & 0.00 \\
    \hline
    Token Count & 144272 & 124636 & 103573 & 34.41* & 0.10 & 124604 & 120939 & 0.52 & 0.00 \\
    \hline
    Avg Word Length & 4.24 & 4.12 & 3.99 & 138.90* & 0.32 & 4.12 & 4.17 & 15.99* & 0.01 \\
    \hline
    Avg Sentence Length & 14.53 & 13.82 & 13.37 & 14.10* & 0.04 & 13.92 & 16.72 & 113.77* & 0.17 \\
    \hline
    Tuldava Score & 3.50 & 3.31 & 3.16 & 58.28* & 0.16 & 3.33 & 3.66 & 123.95* & 0.14 \\
    \hline
  \end{tabular}
  \caption{Welch’s ANOVA statistics comparing inter-genre (SF, MY, ROM) and genre vs literary fiction for combined author gender. Values are rounded to two decimal places, excluding event count and token count, which are rounded to zero decimal places. * indicates significance ($p<0.05$). For standard deviations and degrees of freedom, see Appendix \ref{appdx:anova_output}.}
  \label{tab:anova_combined}
\end{table}

\begin{table} [H]
  \centering
  \small
  \renewcommand{\arraystretch}{1.2}
  \setlength{\tabcolsep}{4pt} 
  \begin{tabular}{|>{\raggedright\arraybackslash}p{3.2cm}|c|c|c|c|c|c|c|c|c|c|}
    \hline
    & \multicolumn{5}{c|}{\textbf{Inter-Genre}} & \multicolumn{4}{c|}{\textbf{Genre vs Literary Fiction}} \\
    \hline
    \textbf{\shortstack[l]{Feature}} & \textbf{\shortstack{\\SF \\Mean}} & \textbf{\shortstack{\\MY \\Mean}} & \textbf{\shortstack{\\ROM \\Mean}} & $\boldsymbol{F}$  & $\boldsymbol{\eta^2_p}$ & \textbf{\shortstack{\\Genre \\Mean}} & \textbf{\shortstack{\\Lit \\Mean}} & $\boldsymbol{F}$ & $\boldsymbol{\eta^2_p}$ \\
    \hline
    Narrative Pace & 2.28 & 2.29 & 2.33 & 16.52*& 0.08 & 2.31 & 2.30 & 0.18& 0.00 \\
    \hline
    Narrative Distance & 2.50 & 2.52 & 2.56 & 33.82* & 0.14 & 2.54 & 2.50 & 13.28* & 0.04 \\
    \hline
    Circuitousness & 0.22 & 0.23 & 0.23 & 2.91 & 0.01 & 0.23 & 0.19 & 12.90* & 0.05 \\
    \hline
    Topical Heterogeneity & 3.02 & 3.03 & 3.04 & 1.31 & 0.01 & 3.03 & 2.99 & 8.06* & 0.03 \\
    \hline
    Event Count & 7186 & 6742 & 6583 & 1.08 & 0.01 & 6740& 5596  & 9.10* & 0.02 \\
    \hline
    Character Count & 49.97 & 45.21 & 27.76 & 61.16* & 0.24 &45.65& 37.00   & 4.62* & 0.02 \\
    \hline
    Protagonist Conc. & 0.24 & 0.23 & 0.31 & 27.78* & 0.12 & 0.27 & 0.22 & 17.83* & 0.04 \\
    \hline
    Written in 1st Person & 0.39 & 0.54 & 0.60 & 4.90* & 0.02 & 0.54 & 0.51 & 0.32 & 0.00 \\
    \hline
    Token Count & 138125 & 124244 & 103533 & 19.39* & 0.10 & 116005& 117351  & 0.04 & 0.00 \\
    \hline
    Avg Word Length & 4.18 & 4.08 & 3.99 & 49.00* & 0.22 & 4.05 & 4.15 & 37.72* & 0.06 \\
    \hline
    Avg Sentence Length & 14.54 & 13.61 & 13.38 & 7.08* & 0.04 & 13.66 & 16.61 & 68.80* & 0.18 \\
    \hline
    Tuldava Score & 3.44 & 3.25 & 3.16 & 20.40* & 0.11 & 3.24 & 3.64 & 100.36* & 0.19 \\
    \hline
  \end{tabular}
  \caption{Welch’s ANOVA statistics comparing inter-genre (SF, MY, ROM) and genre vs literary fiction for female authors. Values are rounded to two decimal places, excluding event count and token count, which are rounded to zero decimal places. * indicates significance ($p<0.05$). For standard deviations and degrees of freedom see Appendix \ref{appdx:anova_output}.}
  \label{tab:anova_combined_female}
\end{table}

\begin{table} [H]
  \centering
  \small
  \renewcommand{\arraystretch}{1.2}
  \setlength{\tabcolsep}{4pt}
  \begin{tabular}{|>{\raggedright\arraybackslash}p{3.2cm}|c|c|c|c|c|c|c|c|c|}
    \hline
    & \multicolumn{4}{c|}{\textbf{Inter-Genre}} & \multicolumn{4}{c|}{\textbf{Genre vs Literary Fiction}} \\
    \hline
    \textbf{\shortstack[l]{Feature}} & \textbf{\shortstack{\\SF \\Mean}} & \textbf{\shortstack{\\MY \\Mean}} & $\boldsymbol{F}$ & $\boldsymbol{\eta^2_p}$ & \textbf{\shortstack{\\Genre \\Mean}} & \textbf{\shortstack{\\Lit \\Mean}} & $\boldsymbol{F}$  & $\boldsymbol{\eta^2_p}$ \\
    \hline
    Narrative Pace & 2.28 & 2.29  & 0.02 & 0.00 & 2.29 & 2.30 & 2.70 & 0.01 \\
    \hline
    Narrative Distance & 2.50 & 2.52  & 4.68* & 0.02 & 2.51 & 2.50 & 1.90 & 0.01 \\
    \hline
    Circuitousness & 0.22 & 0.23  & 5.25* & 0.03 & 0.23 & 0.19 & 11.22* & 0.06 \\
    \hline
    Topical Heterogeneity & 3.02 & 3.03  & 0.47 & 0.00 & 3.02 & 2.99 & 5.54* & 0.03 \\
    \hline
    Event Count & 7186 & 6742  & 1.30 & 0.01 & 6911 & 5596 & 11.20* & 0.05 \\
    \hline
    Character Count & 49.97 & 45.21  & 1.75 & 0.01 & 47.03 & 45.65 & 0.11 & 0.00 \\
    \hline
    Protagonist Conc. & 0.24 & 0.23  & 0.87 & 0.00 & 0.23 & 0.22 & 1.90 & 0.01 \\
    \hline
    Written in 1st Person & 0.39 & 0.54 & 4.21* & 0.02 & 0.48 & 0.51 & 0.16 & 0.00 \\
    \hline
    Token Count & 138125 & 124244  & 4.34* & 0.03 & 129532 & 117351 & 2.92 & 0.01 \\
    \hline
    Avg Word Length & 4.18 & 4.08  & 19.47* & 0.10 & 4.12 & 4.15 & 2.95 & 0.01 \\
    \hline
    Avg Sentence Length & 14.54 & 13.61  & 7.69* & 0.04 & 13.96 & 16.61 & 49.77* & 0.19 \\
    \hline
    Tuldava Score & 3.44 & 3.25  & 15.72* & 0.08 & 3.32 & 3.64 & 53.16* & 0.17 \\
    \hline
  \end{tabular}
  \caption{Welch’s ANOVA statistics comparing inter-genre (SF \& MY) and genre vs literary fiction for female authors. Values are rounded to two decimal places, except for token and event count, which are rounded to zero decimal places. * indicates significance ($p<0.05$). For standard deviations and degrees of freedom, see Appendix \ref{appdx:anova_output}.}
  \label{tab:anova_combined_female_no_rom}
\end{table}

\begin{table} [H]
  \centering
  \small
  \renewcommand{\arraystretch}{1.2}
  \setlength{\tabcolsep}{4pt}
  \begin{tabular}{|>{\raggedright\arraybackslash}p{3.2cm}|c|c|c|c|c|c|c|c|c|}
    \hline
    & \multicolumn{4}{c|}{\textbf{Inter-Genre}} & \multicolumn{4}{c|}{\textbf{Genre vs Literary Fiction}} \\
    \hline
    \textbf{\shortstack[l]{Feature}} & \textbf{\shortstack{\\SF \\Mean}} & \textbf{\shortstack{\\MY \\Mean}} & $\boldsymbol{F}$ & $\boldsymbol{\eta^2_p}$ & \textbf{\shortstack{\\Genre \\Mean}} & \textbf{\shortstack{\\Lit \\Mean}} & $\boldsymbol{F}$ & $\boldsymbol{\eta^2_p}$ \\
    \hline
    Narrative Pace & 2.24 & 2.24  & 0.11 & 0.00 & 2.24 & 2.28 & 19.23* & 0.05 \\
    \hline
    Narrative Distance & 2.47 & 2.49  & 6.02* & 0.02 & 2.48 & 2.48 & 0.22 & 0.00 \\
    \hline
    Circuitousness & 0.23 & 0.25  & 7.99* & 0.03 & 0.24 & 0.20 & 31.53* & 0.09 \\
    \hline
    Topical Heterogeneity & 3.00 & 2.99  & 0.52 & 0.00 & 3.00 & 2.98 & 1.85 & 0.01 \\
    \hline
    Event Count & 7335 & 6595  & 4.57* & 0.01 & 7022 & 5702 & 13.87* & 0.04 \\
    \hline
    Character Count & 54.39 & 51.36  & 0.63 & 0.00 & 53.11 & 51.85 & 0.07  & 0.00 \\
    \hline
    Protagonist Conc. & 0.24 & 0.23  & 0.42 & 0.00 & 0.24 & 0.21 & 4.64* & 0.01 \\
    \hline
    Written in 1st Person & 0.36 & 0.49  & 4.46* & 0.02 & 0.42 & 0.61 & 13.20* & 0.03 \\
    \hline
    Token Count & 147223 & 125185  & 10.92* & 0.04 & 137899 & 123525 & 3.71  & 0.01 \\
    \hline
    Avg Word Length & 4.27 & 4.15  & 45.41* & 0.14 & 4.22 & 4.18 & 5.99* & 0.01 \\
    \hline
    Avg Sentence Length & 14.53 & 14.08  & 2.62 & 0.01 & 14.34 & 16.80 & 42.09* & 0.14 \\
    \hline
    Tuldava Score & 3.53 & 3.37  & 15.54* & 0.05 & 3.47 & 3.67 & 22.97* & 0.06 \\
    \hline
  \end{tabular}
  \caption{Welch’s ANOVA statistics comparing inter-genre (SF \& MY) and genre vs literary fiction for male authors. Values are rounded to two decimal places, except for token and event count, which are rounded to zero decimal places. * indicates significance ($p<0.05$). For standard deviations and degrees of freedom, see Appendix \ref{appdx:anova_output}.}
  \label{tab:anova_combined_male}
\end{table}

\subsection{Modeling the Effect of Author Gender}
Table \ref{tab:regression_outputs} shows the results of our logistic regression modeling. The following features have an independently significant effect ($p<0.05$) on a book's classification as literary or genre fiction: protagonist concentration, average word length, average sentence length, and circuitousness. Furthermore, female authorship has a significant effect on the impact of average word length on literary classification. It is possible that other features or author interactions do affect literary classification but that a logistic regression model is not expressive enough to capture these effects at a 95\% confidence interval. For example, if we look at female authorship independent of feature usage, our coefficient (-0.53) and odds ratio (0.59) suggest a potential decrease in the odds of being classified as literary compared to male authors (potentially 41\% lower odds), but this potential effect is not statistically significant at the 95\% confidence interval ($p$ = 0.13, 95\% CI [–1.22, 0.15]). Therefore, we cannot draw definite conclusions about the effect of author gender on literary classification, independent of literary features.

The results of our Likelihood Ratio test $(\chi^2(7) = 20.12,\ p = 0.0053)$ confirm that including the 7 author-feature interaction terms in our model significantly improves fit compared to the model without interaction terms. 

\begin{figure}[t!]
  \centering
  \includegraphics[width=0.45\linewidth]{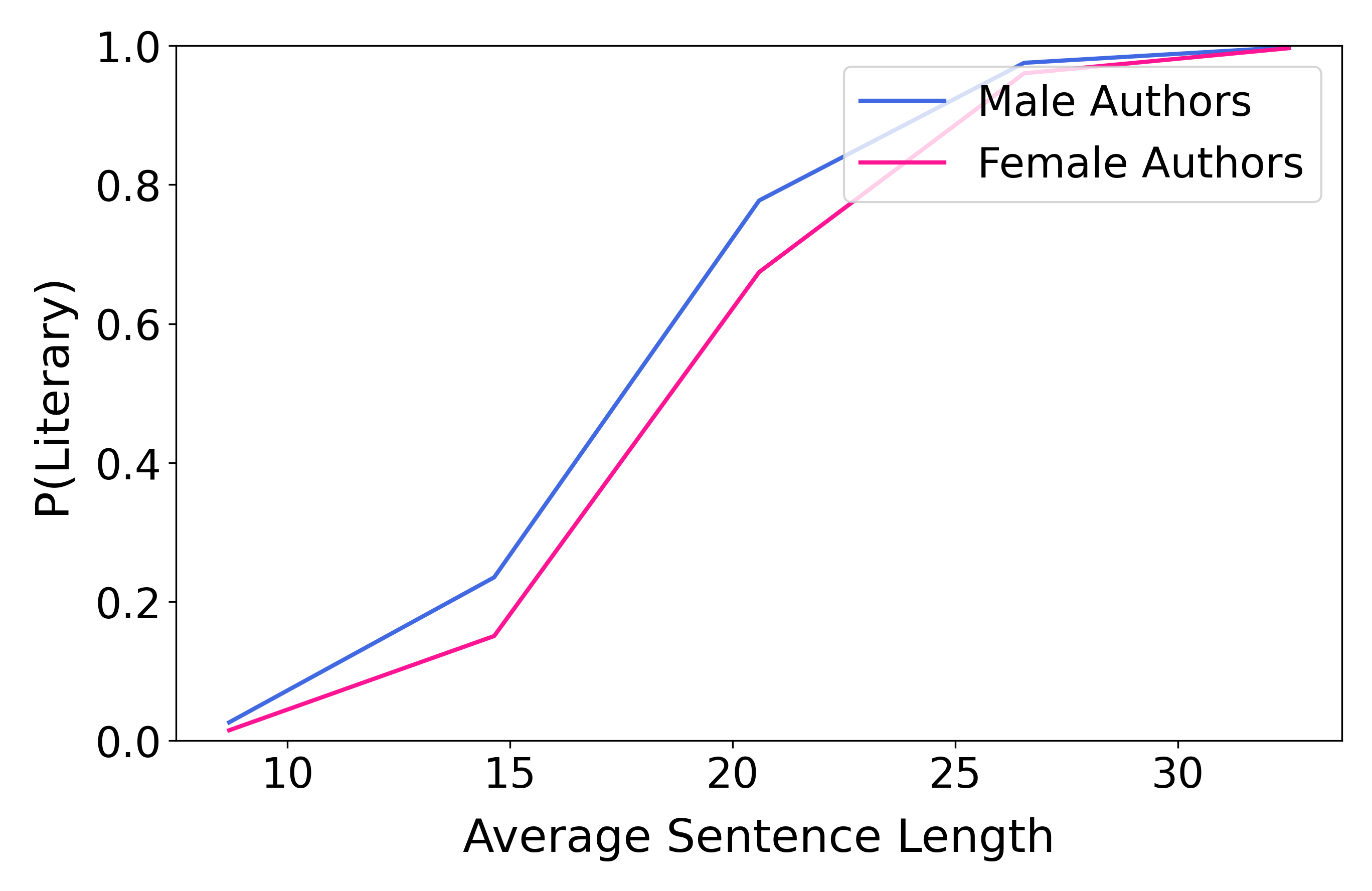}
  \includegraphics[width=0.45\linewidth]{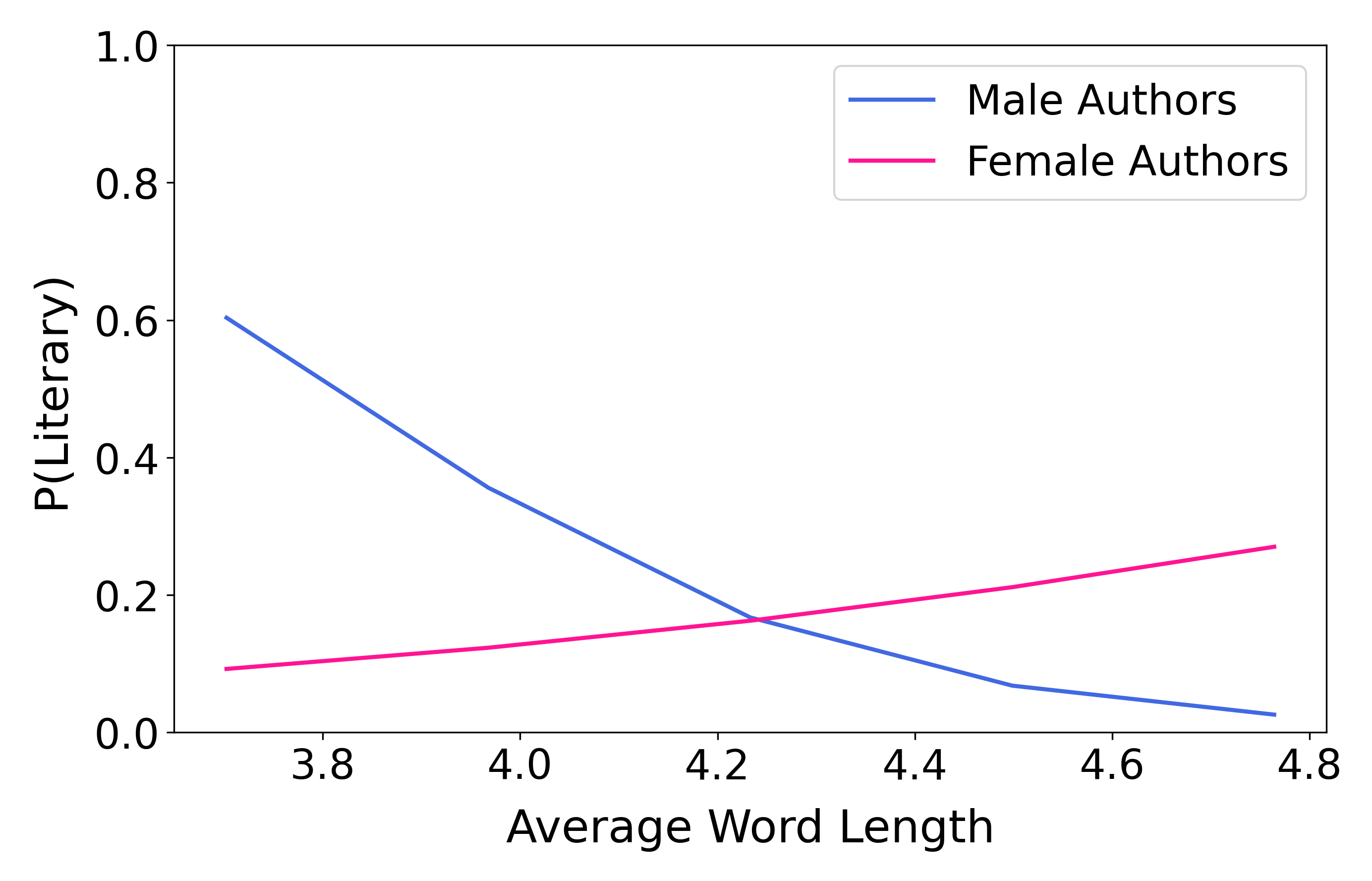}
  \centering
  \includegraphics[width=0.45\linewidth]{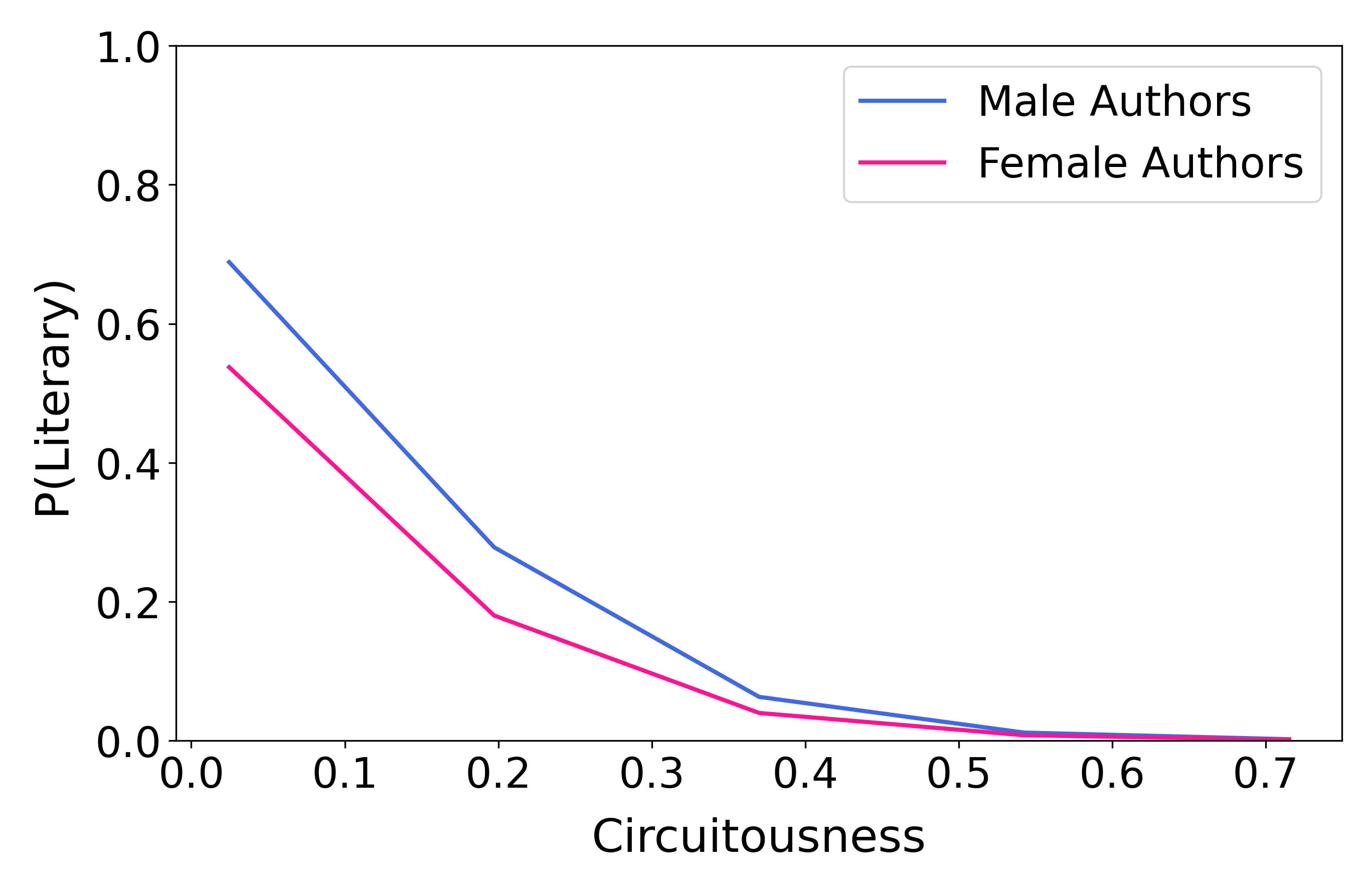}
  \includegraphics[width=0.45\linewidth]{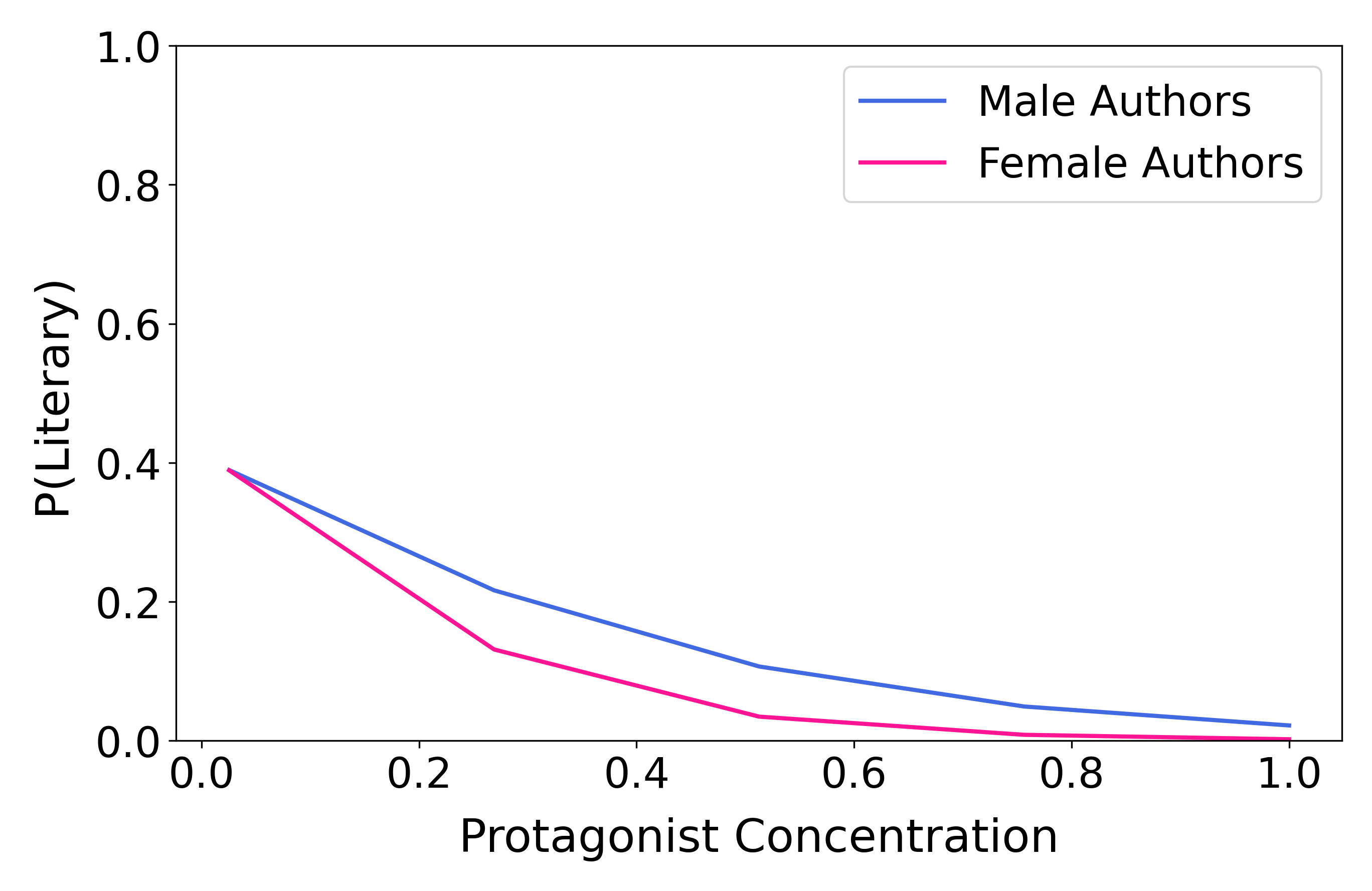}
    \caption{Plots depicting the probability of literary classification relative to feature usage for features with a significant impact ($p<0.05$) on model results. Note that the only the interaction effect which is significant ($p<0.05$) at a 95\% confidence interval is the interaction effect between author gender and average word length.}
  \label{fig:interaction_plots}
\end{figure}

Figure \ref{fig:interaction_plots} illustrates the probability of a book being classified as literary fiction based on author gender and feature usage, for the features which have a significant effect ($p<0.05$) on classification. Average sentence length is positively correlated with literary classification, whereas circuitousness and protagonist concentration are inversely correlated with literary classification. For male authors, average word length is inversely correlated with literary classification, whereas for female authors, the correlation between average word length and literary classification is positive. 

\subsection{Unigram Comparison}
Table \ref{tab:permdisp} shows the results of the PERMDISP tests assessing the dispersion within individual genres and within genre and literary fiction, for both our raw unigram and static embeddings. Literary and genre fiction are significantly different in their raw unigram dispersions ($p$ = 0.2813), and thus it would not be possible to compare them using a method that assumes homogeneity in this area.

\begin{table}[H]
  \centering 
  \small
  \begin{tabular}{lcccc|cccc}
    \toprule
    & \multicolumn{4}{c|}{\textbf{Raw Unigrams}} & \multicolumn{4}{c}{\textbf{Static Embeddings}} \\
    \cmidrule(lr{0.0pt}){2-5}\cmidrule(l{0.0pt}r){6-9}
    \textbf{\shortstack[l]{\\Comparison\\Categories}} 
      & $\boldsymbol{F}$ & $\boldsymbol{p}$ & $\boldsymbol{\eta^2_p}$ & \textbf{\shortstack{Mean Dist.\\to Centroid}} 
      & $\boldsymbol{F}$ & $\boldsymbol{p}$ & $\boldsymbol{\eta^2_p}$ & \textbf{\shortstack{Mean Dist.\\to Centroid}}\\
    \midrule
    SF vs MY vs ROM 
      & 5.03 & 0.00 & 0.02 & \shortstack{MY: 0.0437\\ROM: 0.0403\\SF: 0.0341} 
      & 10.39 & 0.00 & 0.03 & \shortstack{MY: 0.0050\\ROM: 0.0050\\SF: 0.0062}\\
    Literary vs Genre  
      & 1.01 & 0.28 & 0.00 & \shortstack{Genre: 0.0447\\Lit: 0.0420} 
      & 7.98 & 0.01 & 0.01 & \shortstack{Genre: 0.0064\\Lit: 0.0055}\\
    \bottomrule
  \end{tabular}
  \captionof{table}{Results of PERMDISP tests with 999 reps. Mean distance to centroid has been rounded to four decimal places. All other values have been rounded to two decimal places. When comparing results across categories, effect size ($\eta^2_p$) is more meaningful than $F$.}
  \label{tab:permdisp}
\end{table}

Table \ref{tab:wd} includes the results of our $W^*_d$ and $T^2_w$ tests, showing the distances between the means of the unigram and static-embedding vectors from within each category. Literary and genre fiction are much more distinct in their style ($\omega^2$ = 0.14) than in their content ($\omega^2$ = 0.05), whereas individual genres are more distinct in their content ($\omega^2$ = 0.29) than their style ($\omega^2$ = 0.16). Romance stands out as being the most distinct category in terms of both style and content. The $p$-values indicate that all these values are significant ($p<0.05$).

\begin{table}[H]
  \centering 
  \small
  \begin{tabular}{lccc|ccc}
    \toprule
        & \multicolumn{3}{c|}{\textbf{Raw Unigrams}} & \multicolumn{3}{c}{\textbf{Static Embeddings}} \\
    \cmidrule(lr{0.0pt}){2-4}\cmidrule(l{0.0pt}r){5-7}
    \textbf{Test} & $\boldsymbol{p}$  & \textbf{Distance-Based Statistic} & $\boldsymbol{\omega^2}$& $\boldsymbol{p}$  & \textbf{Distance-Based Statistic} & $\boldsymbol{\omega^2}$\\
    \midrule
    $W^*_d$ Inter-Genre & 0.00  &75.87 & 0.16 & 0.00 & 137.94 & 0.29\\
    Lit vs Genre $T^2_w$ & 0.00 & 163.86 & 0.14 & 0.00 & 53.82 & 0.05\\
    Lit vs SF $T^2_w$& 0.00  & 76.80 & 0.16 & 0.00 & 54.28 & 0.11\\
    Lit vs MY $T^2_w$& 0.00  & 93.70 & 0.18 & 0.00 & 38.94 & 0.08\\
    Lit vs ROM $T^2_w$& 0.00  & 247.48 & 0.38 & 0.00 & 166.41 & 0.30\\
    MY vs ROM $T^2_w$ &  0.00 & 50.57 & 0.10 & 0.00 & 116.40 & 0.21\\
    MY vs SF $T^2_w$ &0.00  &19.34  & 0.04 & 0.00 & 73.94 & 0.14\\
    ROM vs SF $T^2_w$  &0.00  &147.36 & 0.26 & 0.00 & 222.46 & 0.34 \\
    \bottomrule
  \end{tabular}
  \captionof{table}{Results of $W^*_d$ and $T^2_w$ tests with 999 reps. Values have been rounded to two decimal places. When comparing results across categories, effect size ($\omega^2$) is more meaningful than the distance-based statistic. See Appendix \ref{appdx:omega} for details on $\omega^2$ calculation.}
  \label{tab:wd}
\end{table}

\section{Discussion}

Our analyses reveals key formal characteristics of individual genres and of genre and literary fiction at large, while also highlighting how author gender affects writing practices and literary classification. 

\subsection{Characteristics of Individual Genres}
Each of the genres examined demonstrates strong conventions with regards to narrative structure, characters, and style.

Contemporary mystery novels feature more circuitous plots than science fiction or romance novels. Such non-linearity has long been considered a characteristic element of mystery novels, due to the fact that they intertwine two stories---that of the crime and that of its investigation \cite{Huhn}. Our mystery novels also feature the lowest protagonist concentration of the genre fiction categories, meaning that the protagonist makes up the lowest percentage of total character mentions. This cannot be attributed to a higher number of total characters or to a lower number of first-person narrators within mystery novels than within other genres (as neither is true). It may be reflective of the unimportance of character within mystery: Nearly a century ago, when speaking of the leisurely enjoyment of mystery novels by academics, Nicolson said, ``Character...troubles us little, though characters we have in abundance" \cite{nicolson-1946}. More recently, Goldman has spoken of the widely critiqued one-dimensionality of characters within mystery novels \cite{goldman-2011} (though he highlights specific examples of characters who he believes challenge this stereotype). Whether due to reader preferences, genre expectations, or the former affecting the latter, our results suggest that these claims about character still hold true for contemporary works in the genre.  

Romance novels cover more distance than mystery or science fiction, meaning that their ending states are the most removed from their starting states. This intuitively makes sense given that romance novels end with a happy couple (which is not how they often begin). Romance novels also have the fastest pace of all the genres and cover the most topics in each book section. Interestingly, fewer actual events occur in romance novels than in mystery or science fiction. One potential explanation for this is how protagonist-centered romance novels are: they feature the protagonist most out of any genre, they feature fewer side characters than other genres, and they are more likely to be written in first-person. It seems possible that romance novels prioritize revealing protagonist interiority over describing external events. This proximity to the protagonist likely aides in the ``reader’s vicarious emotional participation" in the romantic relationship, which is central to romance \cite{saricks-2009}. Furthermore, romance novels are by far the easiest genres to read, as indicated by their Tuldava scores, average word lengths, and average sentence lengths. Some have asserted that romance is not a respected because of its association with women \cite{fendley-2019}, and while many scholars have made strong arguments for this \cite{verboord-2012, radway-1991, tuchman-1984}, its syntactic simplicity might also be a contributing factor.

Overall, science fiction features the most events but covers the least narrative distance. Science fiction narratives feature more characters than other genres and are least likely to be written in first-person. Science fiction is consistently the most difficult-to-read genre as indicated by its mean Tuldava score, average word length, average sentence length, and total word count. Though the protagonist concentration within science fiction is slightly higher than in mystery novels, it is much lower than in romance novels. Similarity to with mystery novels, this could be attributed to a lack of emphasis on character, which has long-been discussed as a common characteristic of science fiction \cite{bereit}. 
    
These observations highlight the formulaic qualities of genre fiction, in line with existing scholarly literature \cite{duff-2000, radway-1991}. Though our unigram analyses do reveal stylistic distinctness between genres, they reveal larger semantic distinctness. This supports theories framing genres as ``regions of conceptual space" \cite{Evnine2015-EVNBII}\footnote{We cite Evnine's work because we use his exact wording to refer to a collection of theories of genre, not because he argues in favor of such theories.} and suggests that further study exploring story content within specific genres would be appropriate.

\paragraph{Gendered Differences within Genres.}

One of the most noteworthy observations about gendered practices within genre fiction begins with our corpus: Out of the 208 romance novels in the CONLIT dataset, only 2 were written by men. Though female authors have long been the primary producers of romance fiction \cite{verboord-2012}, this staggering of a difference far exceeds author gender ratios suggested by other researchers \cite{talbot-2023}.\footnote{It is important to note that while WordsRated identifies as a literary research group, their findings are not externally validated or peer-reviewed.} The difference could be attributed to the fact that the romance novels in the CONLIT dataset were drawn from a list of bestselling romance novels on Amazon, in which female authors might be more highly represented than in the romance genre more broadly. 

While romance novels are distinct from science fiction and mystery novels in terms of their form, content, and author gender, science fiction and mystery texts are quite similar across female and male authors. For the 6 features which are significantly different between genres for authors of both genders and for the 2 features which are significantly different between genres for one gender, the genre with the higher mean value is consistent across gender.

\subsection{Genre versus Literary Fiction}

The broader categories of genre and literary fiction exhibit noteworthy distinctions with regards to form, though not necessarily in the areas proposed by existing scholarship \cite{rosen-2019, saricks-2009, duff-2000, radway-1991}. Compared to genre fiction, literary fiction does contain longer and more difficult to read sentences, which supports the notion that literary fiction features more elevated language.

In contrast with the idea that literary fiction typically features slow, complex narratives, our analysis finds that literary fiction is not significantly slower-paced than genre fiction overall and is consistently more linear in its narratives. In fact, increased circuitousness (narrative non-linearity) is a significant predictor of genre fiction as opposed to literary fiction, as illustrated in Figure \ref{fig:interaction_plots}. Thus, the supposed elevated and complex character of literary fiction do not extend beyond language into more complex narrative structures.

Interestingly, literary fiction consistently has a lower protagonist concentration than any of our genres. Though it seemed reasonable to attribute the relatively low protagonist concentrations of mystery and science fiction to the unimportance of character within these genres, character development and complexity are considered key attributes of literary fiction \cite{saricks-2009}. Such an interpretation seems inaccurate in this context. However, cognitive literary studies scholars suggest a more plausible explanation: Kidd et al. \cite{kidd-2016} found that though genre and literary fiction contain equal amounts of social and cognitive content, within literary fiction, this content demonstrates more ``sophisticated explanations of one’s own and others’ behavior in terms of mental states." Perhaps the protagonists in literary fiction are mentioned less frequently than the protagonists in genre fiction, but each mention reveals more character complexity. Alternatively, perhaps the narrators or protagonists in literary fiction spend more time discussing other characters' mental states, resulting in less fewer mentions of the protagonist relative to mentions of other characters.   

Our unigram analyses (Table \ref{tab:wd}) show that literary fiction is much more defined by its style than by its content, and thus we believe that emphasizing form in our comparison of genre and literary fiction is appropriate.

\paragraph{Prestige by Author Gender.}

When interpreting our Welch ANOVA results for genre versus literary fiction, excluding romance novels can ensure that our interpretations are based on differences due to author gender, not to the presence or absence of romance fiction in the genre-fiction corpus. For science fiction, mystery, and literary fiction, the difference in feature distribution between genre and literary fiction is significant in 8 categories for male authors but only 5 categories for female authors (see Tables \ref{tab:anova_combined_female_no_rom} and \ref{tab:anova_combined_male}). This means that, for these genres, there are fewer formal characteristics associated with literary prestige for female authors than for male authors.

Author gender has an interesting impact on the role that elevated language plays in literary classification. Whether looking at only science fiction and mystery or all 3 genres, Tuldava score and average sentence length have a larger effect size for female authors than for male authors when comparing genre versus literary fiction. What is truly fascinating is that even though these two features are consistently correlated with increased literariness, the same cannot be said for average word length. While female authors become more literary as they use longer words, male authors become less literary (see Figure \ref{fig:interaction_plots}). This phenomena, combined with the higher effect sizes of Tuldava score and sentence length, suggests that female-authored books might be evaluated for literary quality more superficially than male-authored books. Male authors might be afforded the opportunity to demonstrate literary quality through a wider range of integrated narrative elements. The idea male authors have more latitude to experiment with form while still being considered literary is also supported by our Welch ANOVA tests: 10 out of 12 features have higher standard deviations within male-authored literary fiction than within female-authored literary fiction.

Together, our results suggest that for male authors, literary fiction is more distinct from genre fiction but more broadly defined, whereas for female authors, literary fiction is less distinct from genre fiction and more narrowly defined. These findings support Hoberek's claim that gender determines the outer limits of literary fiction \cite{hoberek-2017}. 

\subsection{Limitations}

It is important to keep in mind that no dataset can every be representative of an entire genre, nor can a feature set be truly representative of a book. Some specific limitations of this project are discussed below:
\paragraph{Narrative Features.} Our approaches were restricted by the fact that the corpus does not include the full texts for each book. While we were able to perform similarity comparison using the raw unigram vectors and static embeddings, static embeddings are not able to capture the same level of nuance and meaning that contextualized embeddings would be able to achieve. Furthermore, due to our not having access to the full bodies of the text, our analyses largely depend on the pre-extracted literary features included in the CONLIT dataset. Though the methods for extracting these features were validated by the researchers who developed them, they were not all evaluated on contemporary fiction, which might impact their effectiveness.

\paragraph{Genres.} Though science fiction, mystery, and romance make up a large portion of genre fiction, they are not representative of all genres. Based on our current analyses, romance appears to be an outlier genre. However, including other genres focused on emotional responses such as relationship fiction or horror \cite{saricks-2009} would likely reduce this effect.

\paragraph{Author Gender.} The books in the CONLIT dataset were manually labeled for author gender based on the perceived gender of the author. There is a chance that some of the authors published under pen names associated with the opposite gender \cite{mclean-2023}, which would have resulted in their gender being inaccurately represented within the dataset and within this project. Furthermore, due to data limitations, this project explores gender in a binary fashion, which is not reflective of the range of identities that authors hold.

\section{Conclusion}

This project highlights the formal characteristics of genre and literary fiction and illustrates how author gender affects literary boundaries. While many of our findings align with previously held notions about genre and form, some of our results undermine such assumptions.  We hope that our empirical analyses serve as a starting point, but not an end point, for theories about genre and prestige in twenty-first Anglophone fiction.

\section*{Acknowledgements}
I would like to thank Jeffrey Williams for his literature recommendations and feedback on the early stages of this project; the CHR reviewers for their comments and suggestions; Jake Hofgard for his suggestions regarding statistical modeling; and Andrew Piper for assembling and sharing the CONLIT dataset.

\printbibliography
\appendix

\section{Assumptions of Welch's ANOVA}
\label{appdx:welch}
For each Welch ANOVA, our book categories are the independent variables, and our feature is the dependent variable. All of the tests are one-way. A $p$-score $< 0.05$ indicates a significant difference.

Welch ANOVA tests rely on assumptions of Independence, Additivity, and Normality \cite{moder-2010}. Our data meets the Independence assumption because no book’s narrative speed, Tuldava score, etc. has the ability to impact other books' measurements in those areas. Furthermore, each book is only sampled one time. The Additivity assumption is met because we have removed books tagged in multiple categories. Because the smallest sample size we consider is 191, we can assume that the sampling distribution of any of these samples is normally distributed according to the Central Limit Theorem \cite{kwak-2017}, thus meeting the Normality assumption \cite{piovesana-2016}.

\section{Additional ANOVA Output}
\label{appdx:anova_output}

\begin{table}[H]
  \centering
  \small
  \renewcommand{\arraystretch}{1.2}
\setlength{\tabcolsep}{3.2pt} 
  \begin{tabular}{|>{\raggedright\arraybackslash}p{2.7cm}|c|c|c|c|c|c|c|c|c|c|c|}
    \hline
    \textbf{\shortstack[l]{Feature}} & \textbf{\shortstack{\\SF \\Mean}} & \textbf{\shortstack{\\MY \\Mean}} & \textbf{\shortstack{\\ROM \\Mean}} & \textbf{\shortstack{\\SF\\ STD}} & \textbf{\shortstack{\\MY \\STD}} & \textbf{\shortstack{\\ROM \\STD}} & \textbf{df1} & \textbf{df2} & $\boldsymbol{F}$ & $\boldsymbol{p}$ & $\boldsymbol{\eta^2_p}$ \\
    \hline
    Narrative Pace & 2.25 & 2.27 & 2.33 & 0.08 & 0.08 & 0.08 & 2 & 438.01 & 62.07 & 0.00 & 0.15 \\
    \hline
    \shortstack[l]{\\Narrative \\Distance} & 2.48 & 2.50 & 2.56 & 0.06 & 0.06 & 0.06 & 2 & 436.45 & 93.61 & 0.00 & 0.22 \\
    \hline
    Circuitousness & 0.23 & 0.24 & 0.23 & 0.05 & 0.05 & 0.05 & 2 & 437.18 & 4.98 & 0.01 & 0.02 \\
    \hline
    \shortstack[l]{\\Topical \\Heterogeneity} & 3.01 & 3.01 & 3.04 & 0.11 & 0.08 & 0.09 & 2 & 427.92 & 6.63 & 0.00 & 0.02 \\
    \hline
    Event Count & 7286 & 6659 & 6575 & 3361 & 1843 & 2849 & 2 & 399.33 & 3.49 & 0.03 & 0.01 \\
    \hline
    Character Count & 52.96 & 48.27 & 27.89 & 36.85 & 18.60 & 13.09 & 2 & 404.03 & 113.14 & 0.00 & 0.15 \\
    \hline
    \shortstack[l]{\\Protagonist \\Conc.} & 0.24 & 0.23 & 0.31 & 0.12 & 0.09 & 0.11 & 2 & 427.13 & 34.03 & 0.00 & 0.09 \\
    \hline
    \shortstack[l]{\\Written in \\1st Person} & 0.37 & 0.51 & 0.61 & 0.48 & 0.50 & 0.49 & 2 & 437.12 & 12.91 & 0.00 & 0.04 \\
    \hline
    Token Count & 144272 & 124636 & 103573 & 61930 & 36214 & 42327 & 2 & 417.69 & 34.41 & 0.00 & 0.10 \\
    \hline
    Avg Word Length & 4.24 & 4.12 & 3.99 & 0.17 & 0.13 & 0.15 & 2 & 429.10 & 138.90 & 0.00 & 0.32 \\
    \hline
    \shortstack[l]{\\Avg Sentence\\ Length} & 14.53 & 13.82 & 13.37 & 2.47 & 2.03 & 2.08 & 2 & 433.98 & 14.10 & 0.00 & 0.04 \\
    \hline
    Tuldava Score & 3.50 & 3.31 & 3.16 & 0.37 & 0.28 & 0.29 & 2 & 430.46 & 58.28 & 0.00 & 0.16 \\
    \hline
  \end{tabular}
  \caption{Welch’s ANOVA statistics for inter-genre comparison with combined author gender. Values are rounded to two decimal places, except for df1, event count, and token count, which are rounded to zero decimal places. $p<0.05$ indicates significance. }
  \label{tab:anova_genre_all}
\end{table}

\begin{table}[H]
  \centering
  \small
  \renewcommand{\arraystretch}{1.2}
  \setlength{\tabcolsep}{3.2pt}
  \begin{tabular}{|>{\raggedright\arraybackslash}p{2.7cm}|c|c|c|c|c|c|c|c|c|c|c|}
    \hline
    \textbf{\shortstack[l]{Feature}} & \textbf{\shortstack{\\SF \\Mean}} & \textbf{\shortstack{\\MY \\Mean}} & \textbf{\shortstack{\\ROM \\Mean}} & \textbf{\shortstack{\\SF\\ STD}} & \textbf{\shortstack{\\MY \\STD}} & \textbf{\shortstack{\\ROM \\STD}} & \textbf{df1} & \textbf{df2} & $\boldsymbol{F}$ & $\boldsymbol{p}$ & $\boldsymbol{\eta^2_p}$ \\
    \hline
    Narrative Pace & 2.28 & 2.29 & 2.33 & 0.07 & 0.08 & 0.08 & 2 & 181.15 & 16.52 & 0.00 & 0.08 \\
    \hline
    \shortstack[l]{\\Narrative \\Distance} & 2.50 & 2.52 & 2.56 & 0.05 & 0.06 & 0.06 & 2 & 190.72 & 33.82 & 0.00 & 0.14 \\
    \hline
    Circuitousness & 0.22 & 0.23 & 0.23 & 0.04 & 0.05 & 0.05 & 2 & 183.81 & 2.91 & 0.06 & 0.01 \\
    \hline
    \shortstack[l]{\\Topical \\Heterogeneity} & 3.02 & 3.03 & 3.04 & 0.08 & 0.07 & 0.09 & 2 & 189.49 & 1.31 & 0.27 & 0.01 \\
    \hline
    Event Count & 7186 & 6742 & 6583 & 3039 & 1659 & 2863 & 2 & 176.10 & 1.08 & 0.34 & 0.01 \\
    \hline
    Character Count & 49.97 & 45.21 & 27.76 & 27.60 & 16.62 & 13.13 & 2 & 149.01 & 61.16 & 0.00 & 0.24 \\
    \hline
    \shortstack[l]{\\Protagonist \\Conc.} & 0.24 & 0.23 & 0.31 & 0.09 & 0.09 & 0.11 & 2 & 190.24 & 27.78 & 0.00 & 0.12 \\
    \hline
    \shortstack[l]{\\Written in \\1st Person} & 0.39 & 0.54 & 0.60 & 0.49 & 0.50 & 0.49 & 2 & 180.02 & 4.90 & 0.01 & 0.02 \\
    \hline
    Token Count & 138125 & 124244 & 103533 & 50615 & 32159 & 42599 & 2 & 174.17 & 19.39 & 0.00 & 0.10 \\
    \hline
    Avg Word Length & 4.18 & 4.08 & 3.99 & 0.15 & 0.13 & 0.15 & 2 & 178.83 & 49.00 & 0.00 & 0.22 \\
    \hline
    \shortstack[l]{\\Avg Sentence\\ Length} & 14.54 & 13.61 & 13.38 & 2.30 & 2.09 & 2.08 & 2 & 174.75 & 7.08 & 0.00 & 0.04 \\
    \hline
    Tuldava Score & 3.44 & 3.25 & 3.16 & 0.34 & 0.28 & 0.29 & 2 & 172.19 & 20.40 & 0.00 & 0.11 \\
    \hline

  \end{tabular}
  \caption{Welch’s ANOVA statistics for inter-genre comparison with for female authors. Values are rounded to two decimal places, except for df1, event count, and token count, which are rounded to zero decimal places. $p<0.05$ indicates significance. }
  \label{tab:anova_genre_female}
\end{table}

\begin{table}
  \centering
  \small
  \renewcommand{\arraystretch}{1.2}
  \setlength{\tabcolsep}{5pt}
  \begin{tabular}{|>{\raggedright\arraybackslash}p{3.2cm}|c|c|c|c|c|c|c|c|c|c|}
    \hline
    \textbf{\shortstack[l]{Feature}} & \textbf{\shortstack{\\SF \\Mean}} & \textbf{\shortstack{\\MY \\Mean}} &  \textbf{\shortstack{\\SF \\STD}}&  \textbf{\shortstack{\\MY \\STD}}& \textbf{df1} & \textbf{df2} & $\boldsymbol{F}$ & $\boldsymbol{p}$ & $\boldsymbol{\eta^2_p}$ \\
    \hline
    Narrative Pace & 2.28 & 2.29  & 0.07 & 0.08& 1& 158.8&0.02 & 0.88 & 0.00 \\
    \hline
    Narrative Distance & 2.50 & 2.52 &0.05&0.06& 1 &155.95& 4.68 & 0.03 & 0.02 \\
    \hline
    Circuitousness & 0.22 & 0.23  &0.04 & 0.05 &1& 170.27& 5.25 & 0.02 & 0.03 \\
    \hline
    Topical Heterogeneity & 3.02 & 3.03 &0.08 & 0.07 &1& 136.85 & 0.47 & 0.50 & 0.00 \\
    \hline
    Event Count & 7186 & 6742  & 3039 & 1659 &1& 97.43& 1.30 & 0.26 & 0.01\\
    \hline
    Character Count & 49.97 & 45.21 & 27.60 & 16.62 &1& 103.09 & 1.75 & 0.19 & 0.01 \\
    \hline
    Protagonist Conc. & 0.24 & 0.23  & 0.09 & 0.09 &1&144.99& 0.87 & 0.35 & 0.00 \\
    \hline
    Written in 1st Person & 0.39 & 0.54 & 0.49 & 0.50 &1& 152.25& 4.21 & 0.04 & 0.02  \\
    \hline
    Token Count & 138125 & 124244 & 50615 & 32159 &1& 106.63&  4.34 & 0.04 & 0.03\\
    \hline
    Avg Word Length & 4.18 & 4.08  & 0.15 & 0.13 &1& 130.63& 19.47 & 0.00 & 0.10\\
    \hline
    Avg Sentence Length & 14.54 & 13.61 & 2.30  & 2.09 &1&139.53 & 7.69 & 0.01 & 0.04 \\
    \hline
    Tuldava Score & 3.44 & 3.25  & 0.34 & 0.28 &1&127.63& 15.72 & 0.00 & 0.08 \\
    \hline
  \end{tabular}
  \caption{Welch’s ANOVA statistics for inter-genre comparison for female authors, with romance novels excluded. Values are rounded to two decimal places, except for df1, event count, and token count, which are rounded to zero decimal places. $p<0.05$ indicates significance. }
  \label{tab:anova_genre_female_no_rom}
\end{table}

\begin{table}[H]
  \centering
  \small
  \renewcommand{\arraystretch}{1.2}
  \setlength{\tabcolsep}{5pt}
  \begin{tabular}{|>{\raggedright\arraybackslash}p{3.2cm}|c|c|c|c|c|c|c|c|c|c|}
    \hline
    \textbf{\shortstack[l]{Feature}} & \textbf{\shortstack{\\SF \\Mean}} & \textbf{\shortstack{\\MY \\Mean}}  & \textbf{\shortstack{\\SF\\ STD}} & \textbf{\shortstack{\\MY \\STD}} & \textbf{df1} & \textbf{df2} & $\boldsymbol{F}$ & $\boldsymbol{p}$ & $\boldsymbol{\eta^2_p}$ \\
    \hline
    Narrative Pace & 2.24 & 2.24  & 0.07 & 0.08  & 1 & 229.78 & 0.11 & 0.74 & 0.00 \\
    \hline
    Narrative Distance & 2.47 & 2.49  & 0.06 & 0.06  & 1 & 233.16 & 6.02 & 0.01 & 0.02 \\
    \hline
    Circuitousness & 0.23 & 0.25  & 0.05 & 0.04  & 1 & 249.87 & 7.99 & 0.01 & 0.03 \\
    \hline
    Topical Heterogeneity & 3.00 & 2.99  & 0.12 & 0.08  & 1 & 257.97 & 0.52 & 0.47 & 0.00 \\
    \hline
    Event Count & 7335 & 6595  & 3515 & 2030 &  1 & 245.86 & 4.57 & 0.03 & 0.01 \\
    \hline
    Character Count & 54.39 & 51.36  & 40.55 & 20.06  & 1 & 230.00 & 0.63 & 0.43 & 0.00 \\
    \hline
    Protagonist Conc. & 0.24 & 0.23  & 0.13 & 0.09  & 1 & 257.47 & 0.42 & 0.52 & 0.00 \\
    \hline
    Written in 1st Person & 0.36 & 0.49  & 0.48 & 0.50 & 1 & 229.27 & 4.46 & 0.04 & 0.02 \\
    \hline
    Token Count & 147223 & 125185  & 66641 & 40462  & 1 & 250.07 & 10.92 & 0.00 & 0.04 \\
    \hline
    Avg Word Length & 4.27 & 4.15 & 0.17 & 0.12  & 1 & 257.89 & 45.41 & 0.00 & 0.14 \\
    \hline
    Avg Sentence Length & 14.53 & 14.08  & 2.56 & 1.98  & 1 & 257.18 & 2.62 & 0.11 & 0.01 \\
    \hline
    Tuldava Score & 3.53 & 3.37  & 0.39 & 0.27  & 1 & 257.15 & 15.54 & 0.00 & 0.05 \\
    \hline
  \end{tabular}
  \caption{Welch’s ANOVA statistics for inter-genre comparison for male authors. Values are rounded to two decimal places, except for df1, event count, and token count, which are rounded to zero decimal places. $p<0.05$ indicates significance. }
  \label{tab:anova_genre_male}
\end{table}

\begin{table} [H]
  \centering
  \small
  \renewcommand{\arraystretch}{1.2}
  \setlength{\tabcolsep}{5pt}
  \begin{tabular}{|>{\raggedright\arraybackslash}p{3.2cm}|c|c|c|c|c|c|c|c|c|}
    \hline
    \textbf{\shortstack[l]{Feature}} & \textbf{\shortstack{\\Genre \\Mean}} & \textbf{\shortstack{\\Lit \\Mean}}  & \textbf{\shortstack{\\Genre\\ STD}} & \textbf{\shortstack{\\Lit \\STD}} & \textbf{df1} & \textbf{df2} & $\boldsymbol{F}$ & $\boldsymbol{p}$ & $\boldsymbol{\eta^2_p}$ \\
    \hline
    Narrative Pace & 2.28 & 2.29 & 0.09 & 0.08 & 1 & 310.31 & 1.32 & 0.25 & 0 \\
    \hline
    Narrative Distance & 2.51 & 2.49 & 0.07 & 0.08 & 1 & 274.38 & 15.71 & 0.00 & 0.02 \\
    \hline
    Circuitousness & 0.23 & 0.20 & 0.05 & 0.07 & 1 & 242.14 & 38.25 & 0.00 & 0.06 \\
    \hline
    Topical Heterogeneity & 3.02 & 2.98 & 0.09 & 0.12 & 1 & 253.63 & 12.71 & 0.00 & 0.02 \\
    \hline
    Event Count & 6844 & 5657 & 2760 & 3175 & 1 & 278.22 & 21.90 & 0.00 & 0.03 \\
    \hline
    Character Count & 43.43 & 49.25 & 27.27 & 40.35 & 1 & 242.26 & 3.51 & 0.06 & 0.01 \\
    \hline
    Protagonist Conc. & 0.26 & 0.21 & 0.11 & 0.11 & 1 & 303.73 & 25.64 & 0.00 & 0.03 \\
    \hline
    Written in 1st Person & 0.49 & 0.57 & 0.50 & 0.49 & 1 & 313.99 & 3.63 & 0.06 & 0 \\
    \hline
    Token Count & 124604 & 120939 & 50722 & 64443 & 1 & 261.7 & 0.52 & 0.47 & 0 \\
    \hline
    Avg Word Length & 4.12 & 4.17 & 0.18 & 0.14 & 1 & 393.99 & 15.99 & 0.00 & 0.01 \\
    \hline
    Avg Sentence Length & 13.92 & 16.72 & 2.25 & 3.42 & 1 & 239.31 & 113.77 & 0.00 & 0.17 \\
    \hline
    Tuldava Score & 3.33 & 3.66 & 0.35 & 0.37 & 1 & 291.39 & 123.95 & 0.00 & 0.14 \\
    \hline
  \end{tabular}
  \caption{Welch’s ANOVA statistics for literary vs genre comparison with combined author gender. Values are rounded to two decimal places, except for df1, event count, and token count, which are rounded to zero decimal places. $p<0.05$ indicates significance. }
  \label{tab:anova_lit_all}
\end{table}

\begin{table}[H]
  \centering
  \small
  \renewcommand{\arraystretch}{1.2}
  \setlength{\tabcolsep}{5pt}
  \begin{tabular}{|>{\raggedright\arraybackslash}p{3.2cm}|c|c|c|c|c|c|c|c|c|}
    \hline
    \textbf{\shortstack[l]{Feature}} & \textbf{\shortstack{\\Genre \\Mean}} & \textbf{\shortstack{\\Lit \\Mean}}  & \textbf{\shortstack{\\Genre\\ STD}} & \textbf{\shortstack{\\Lit \\STD}} & \textbf{df1} & \textbf{df2} & $\boldsymbol{F}$ & $\boldsymbol{p}$ & $\boldsymbol{\eta^2_p}$ \\
    \hline
    Narrative Pace & 2.31 & 2.30 & 0.08 & 0.09 & 1 & 106.98 & 0.18 & 0.67 & 0.00 \\
    \hline
    Narrative Distance & 2.54 & 2.50 & 0.06 & 0.09 & 1 & 94.58 & 13.28 & 0.00 & 0.04 \\
    \hline
    Circuitousness & 0.23 & 0.19 & 0.05 & 0.09 & 1 & 89.47 & 12.90 & 0.00 & 0.05 \\
    \hline
    Topical Heterogeneity & 3.03 & 2.99 & 0.08 & 0.13 & 1 & 92.20 & 8.06 & 0.01 & 0.03 \\
    \hline
    Event Count &  6740& 5596 &  2605& 3185 & 1 & 101.54 & 9.10 & 0.00 & 0.02 \\
    \hline
    Character Count & 37.00 & 45.65  & 20.10 & 34.83 & 1 & 89.97 & 4.62 & 0.03 & 0.02 \\
    \hline
    Protagonist Conc. & 0.27 & 0.22 & 0.10 & 0.11 & 1 & 109.14 & 17.83 & 0.00 & 0.04 \\
    \hline
    Written in 1st Person & 0.54 & 0.51 & 0.50 & 0.50 & 1 & 113.40 & 0.32 & 0.57 & 0.00 \\
    \hline
    Token Count & 116005 & 117351  & 43601 & 58018& 1 & 97.90 & 0.04 & 0.84 & 0.00 \\
    \hline
    Avg Word Length & 4.05 & 4.15 & 0.16 & 0.13 & 1 & 135.95 & 37.72 & 0.00 & 0.06 \\
    \hline
    Avg Sentence Length  & 13.66 & 16.61& 2.16 & 3.03 & 1 & 95.95 & 68.80 & 0.00 & 0.18 \\
    \hline
    Tuldava Score & 3.24 & 3.64 & 0.31 & 0.33 & 1 & 108.95 & 100.36 & 0.00 & 0.19 \\
    \hline
  \end{tabular}
  \caption{Welch’s ANOVA statistics for genre vs literary fiction comparison for female authors. Values are rounded to two decimal places, except for df1, event count, and token count, which are rounded to zero decimal places. $p<0.05$ indicates significance. }
  \label{tab:anova_literary_female}
\end{table}

\begin{table}[H]
  \centering
  \small
  \renewcommand{\arraystretch}{1.2}
  \setlength{\tabcolsep}{5pt}
  \begin{tabular}{|>{\raggedright\arraybackslash}p{3.2cm}|c|c|c|c|c|c|c|c|c|}
    \hline
    \textbf{\shortstack[l]{Feature}} & \textbf{\shortstack{\\Genre \\Mean}} & \textbf{\shortstack{\\Lit \\Mean}}  & \textbf{\shortstack{\\Genre\\ STD}} & \textbf{\shortstack{\\Lit \\STD}} & \textbf{df1} & \textbf{df2} & $\boldsymbol{F}$ & $\boldsymbol{p}$ & $\boldsymbol{\eta^2_p}$ \\
    \hline
    Narrative Pace & 2.39 & 2.30 & 0.08 & 0.09 & 1 & 133.19 & 2.7 & 0.10 & 0.01 \\
    \hline
    Narrative Distance & 2.51 & 2.50 & 0.06 & 0.09 & 1 & 103.79 & 1.9 & 0.17 & 0.01 \\
    \hline
    Circuitousness & 0.23 & 0.19 & 0.05 & 0.09 & 1 & 102.09 & 11.22 & 0.00 & 0.06 \\
    \hline
    Topical Heterogeneity & 3.02 & 2.99 & 0.07 & 0.13 & 1 & 100.71 & 5.54 & 0.02 & 0.03 \\
    \hline
    Event Count &  6911.11& 5596 &  2288& 3185 & 1 & 114.99 & 11.20 & 0.00 & 0.05 \\
    \hline
    Character Count & 47.03 & 45.65  & 21.53 & 34.83 & 1 & 105.47 & 0.11 & 0.74 & 0.00 \\
    \hline
    Protagonist Conc. & 0.23 & 0.22 & 0.09 & 0.11 & 1 & 123.88 & 1.90 & 0.17 & 0.01 \\
    \hline
    Written in 1st Person & 0.48 & 0.51 & 0.50 & 0.50 & 1 & 149.34 & 0.16 & 0.69 & 0.00 \\
    \hline
    Token Count & 129532 & 117351  & 40636 & 58018 & 1 & 113.16 & 2.92 & 0.09 & 0.01 \\
    \hline
    Avg Word Length & 4.12 & 4.15 & 0.15 & 0.13 & 1 & 170.18 & 2.95 & 0.09 & 0.01 \\
    \hline
    Avg Sentence Length  & 13.96 & 16.61& 2.21 & 3.03 & 1 & 116.2 & 49.77 & 0.00 & 0.19 \\
    \hline
    Tuldava Score & 3.32 & 3.64 & 0.31 & 0.33 & 1 & 141.79 & 53.16 & 0.00 & 0.17 \\
    \hline
  \end{tabular}
  \caption{Welch’s ANOVA statistics for genre vs literary fiction comparison for female authors, excluding romance. Values are rounded to two decimal places, except for df1, event count, and token count, which are rounded to zero decimal places. $p<0.05$ indicates significance. }
  \label{tab:anova_literary_female_no_rom}
\end{table}

\begin{table}[H]
  \centering
  \small
  \renewcommand{\arraystretch}{1.2}
  \setlength{\tabcolsep}{5pt}
  \begin{tabular}{|>{\raggedright\arraybackslash}p{3.2cm}|c|c|c|c|c|c|c|c|c|}
    \hline
    \textbf{\shortstack[l]{Feature}} & \textbf{\shortstack{\\Genre \\Mean}} & \textbf{\shortstack{\\Lit \\Mean}}  & \textbf{\shortstack{\\Genre\\ STD}} & \textbf{\shortstack{\\Lit \\STD}} & \textbf{df1} & \textbf{df2} & $\boldsymbol{F}$ & $\boldsymbol{p}$ & $\boldsymbol{\eta^2_p}$ \\
    \hline
    Narrative Pace & 2.24 & 2.28 & 0.08 & 0.08 & 1 & 196.49 & 19.23 & 0.00 & 0.05 \\
    \hline
    Narrative Distance & 2.48 & 2.48 & 0.06 & 0.07 & 1 & 184.41 & 0.22 & 0.64 & 0.00 \\
    \hline
    Circuitousness & 0.24 & 0.20 & 0.05 & 0.06 & 1 & 171.27 & 31.53 & 0.00 & 0.09 \\
    \hline
    Topical Heterogeneity & 3.00 & 2.98 & 0.10 & 0.12 & 1 & 183.10 & 1.85 & 0.18 & 0.01 \\
    \hline
    Event Count & 7022 & 5702 & 2996 & 3181 & 1 & 197.08 & 13.87 & 0.00 & 0.04 \\
    \hline
    Character Count & 53.11 & 51.85 & 33.43 & 43.88 & 1 & 166.92 & 0.07 & 0.79 & 0.00 \\
    \hline
    Protagonist Conc. & 0.24 & 0.21 & 0.12 & 0.12 & 1 & 213.29 & 4.64 & 0.03 & 0.01 \\
    \hline
    Written in 1st Person & 0.42 & 0.61 & 0.49 & 0.48 & 1 & 213.53 & 13.20 & 0.00 & 0.03 \\
    \hline
    Token Count & 137899 & 123525 & 57990 & 68844 & 1 & 179.73 & 3.71 & 0.06 & 0.01 \\
    \hline
    Avg Word Length & 4.22 & 4.18 & 0.16 & 0.15 & 1 & 226.86 & 5.99 & 0.02 & 0.01 \\
    \hline
    Avg Sentence Length & 14.34 & 16.80 & 2.34 & 3.69 & 1 & 148.98 & 42.09 & 0.00 & 0.14 \\
    \hline
    Tuldava Score & 3.47 & 3.67 & 0.35 & 0.40 & 1 & 185.84 & 22.97 & 0.00 & 0.06 \\
    \hline
  \end{tabular}
  \caption{Welch’s ANOVA statistics for genre vs literary fiction comparison for male authors. Values are rounded to two decimal places, except for df1, event count, and token count, which are rounded to zero decimal places. $p<0.05$ indicates significance. }
  \label{tab:anova_literary_male}
\end{table}

\section{Feature Histograms}
\label{appdx:histograms}
\begin{figure}[H]
  \centering
  \includegraphics[width=0.45\linewidth]{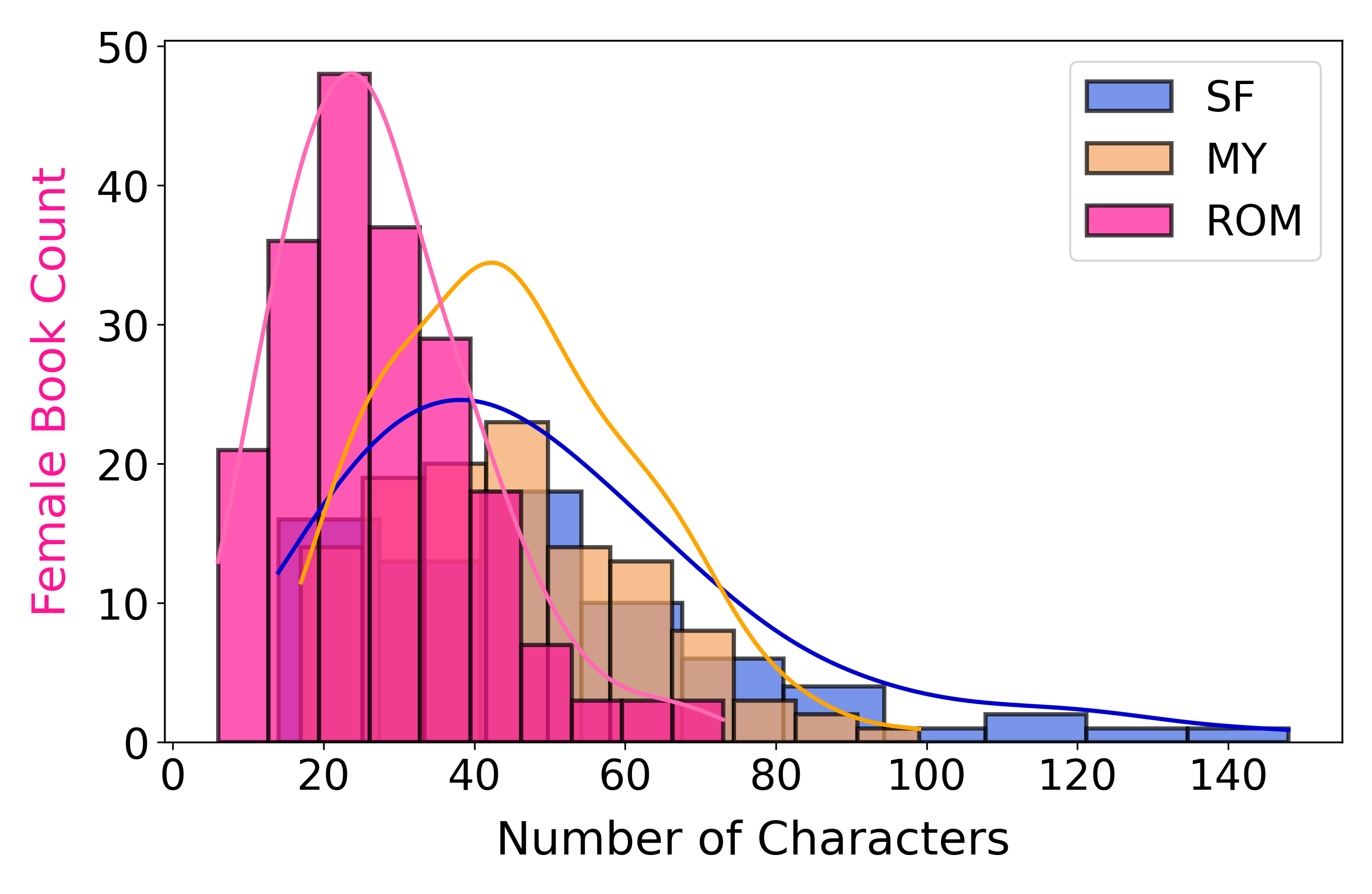}
  \includegraphics[width=0.45\linewidth]{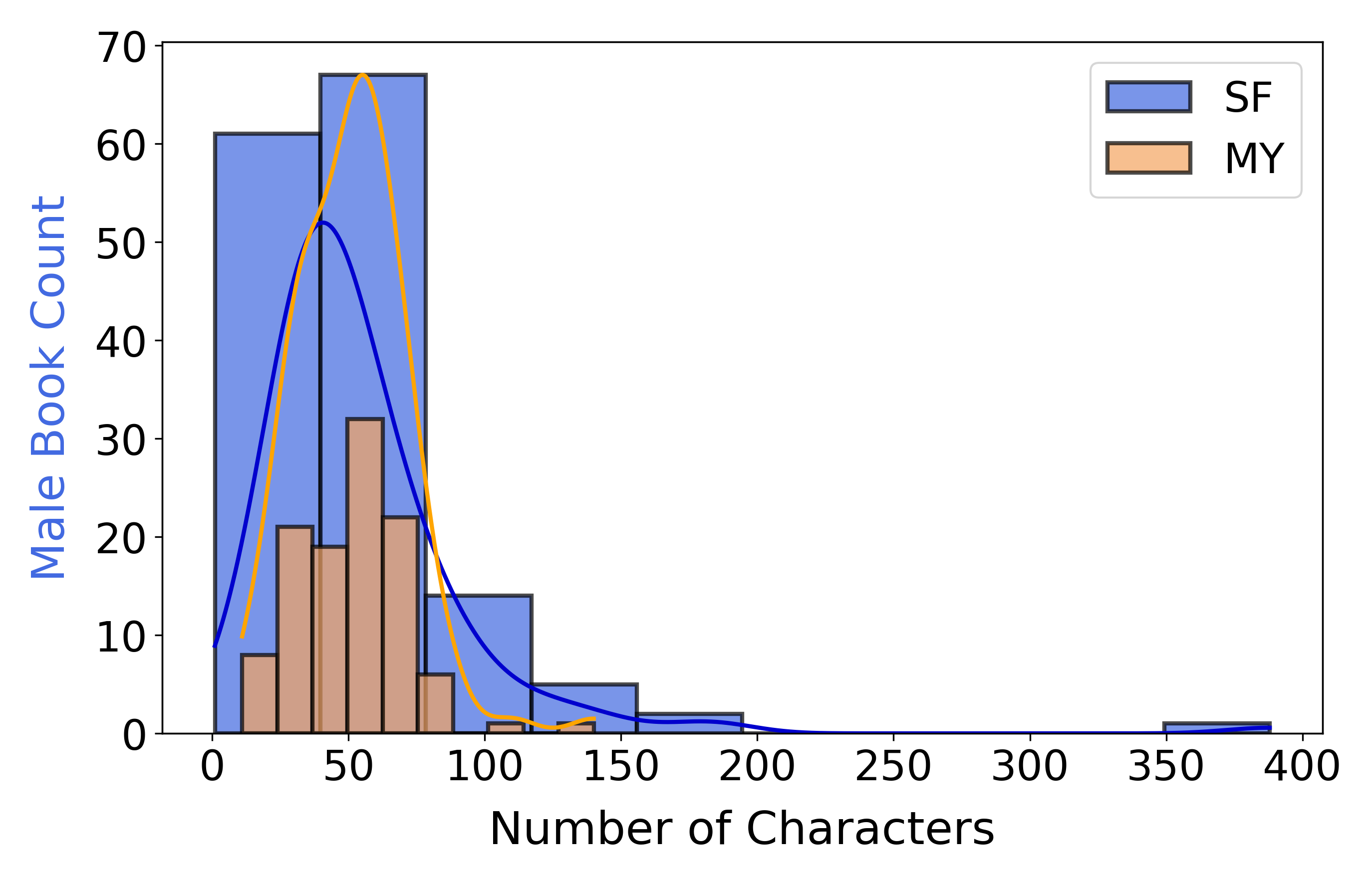}
\centering
  \includegraphics[width=0.45\linewidth]{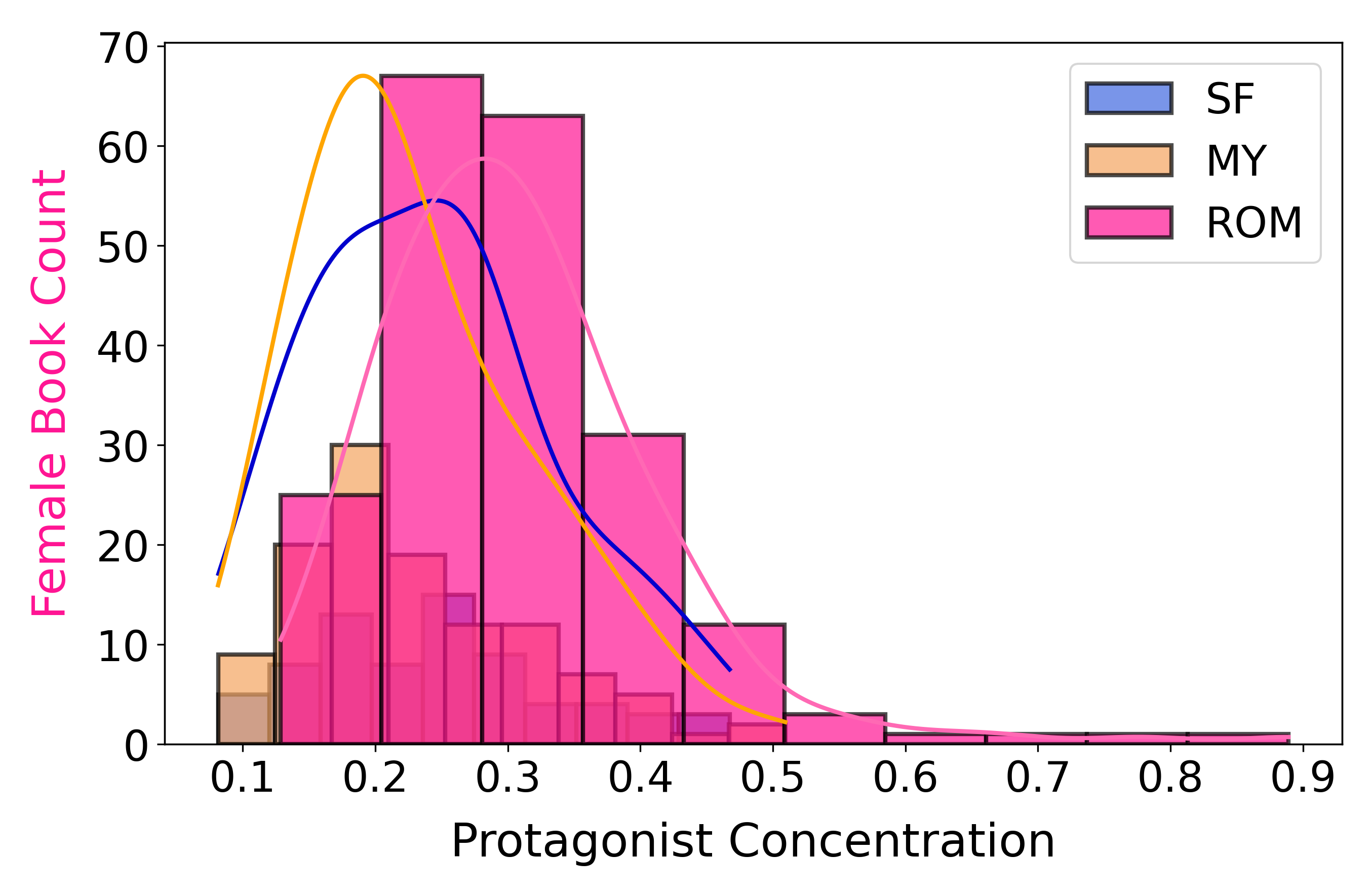}
  \includegraphics[width=0.45\linewidth]{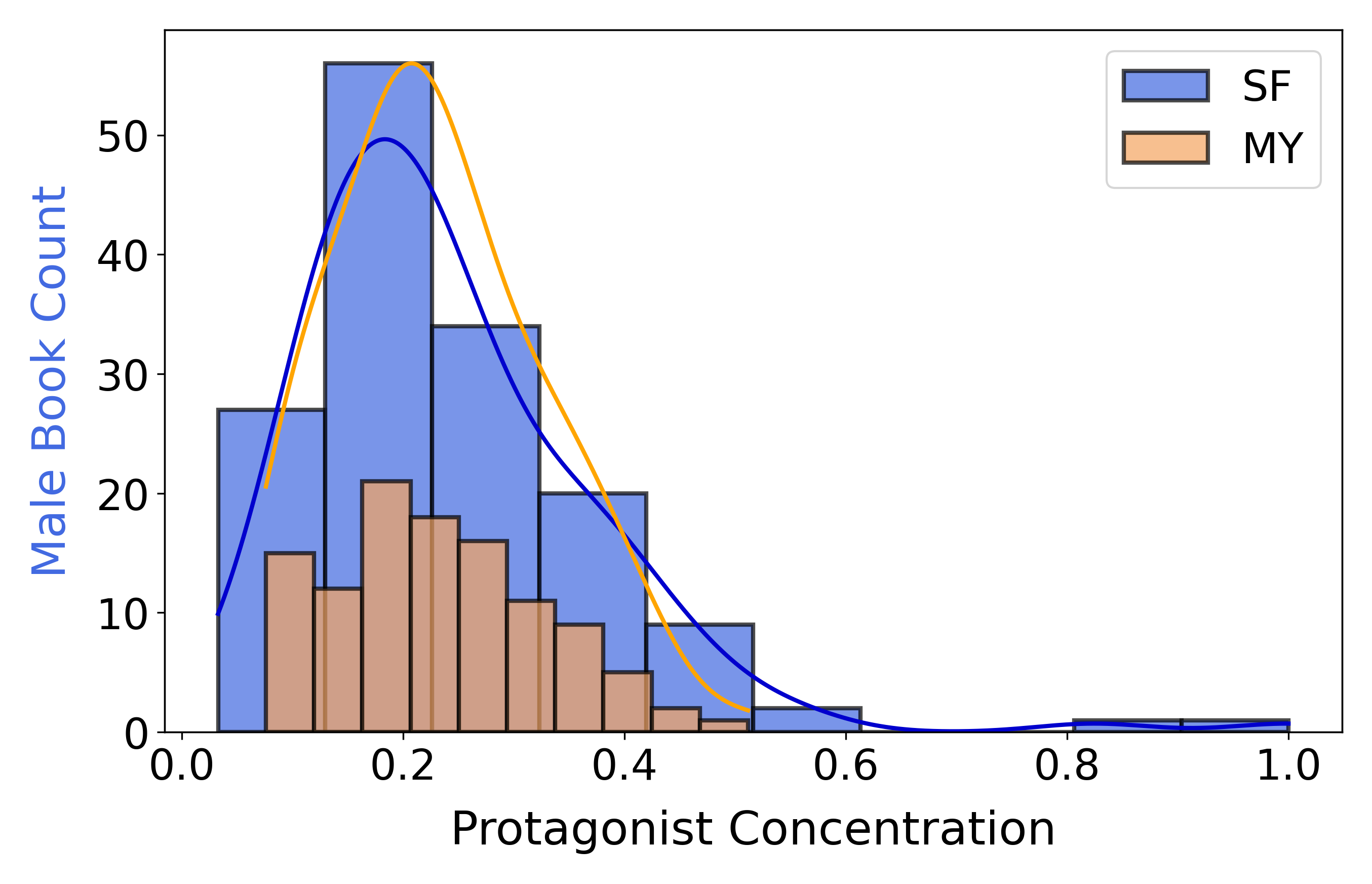}
    \caption{Character histograms for genre fiction by author gender}
  \label{fig:hist1}
\end{figure}

\begin{figure}[H]
  \centering
  \includegraphics[width=0.45\linewidth]{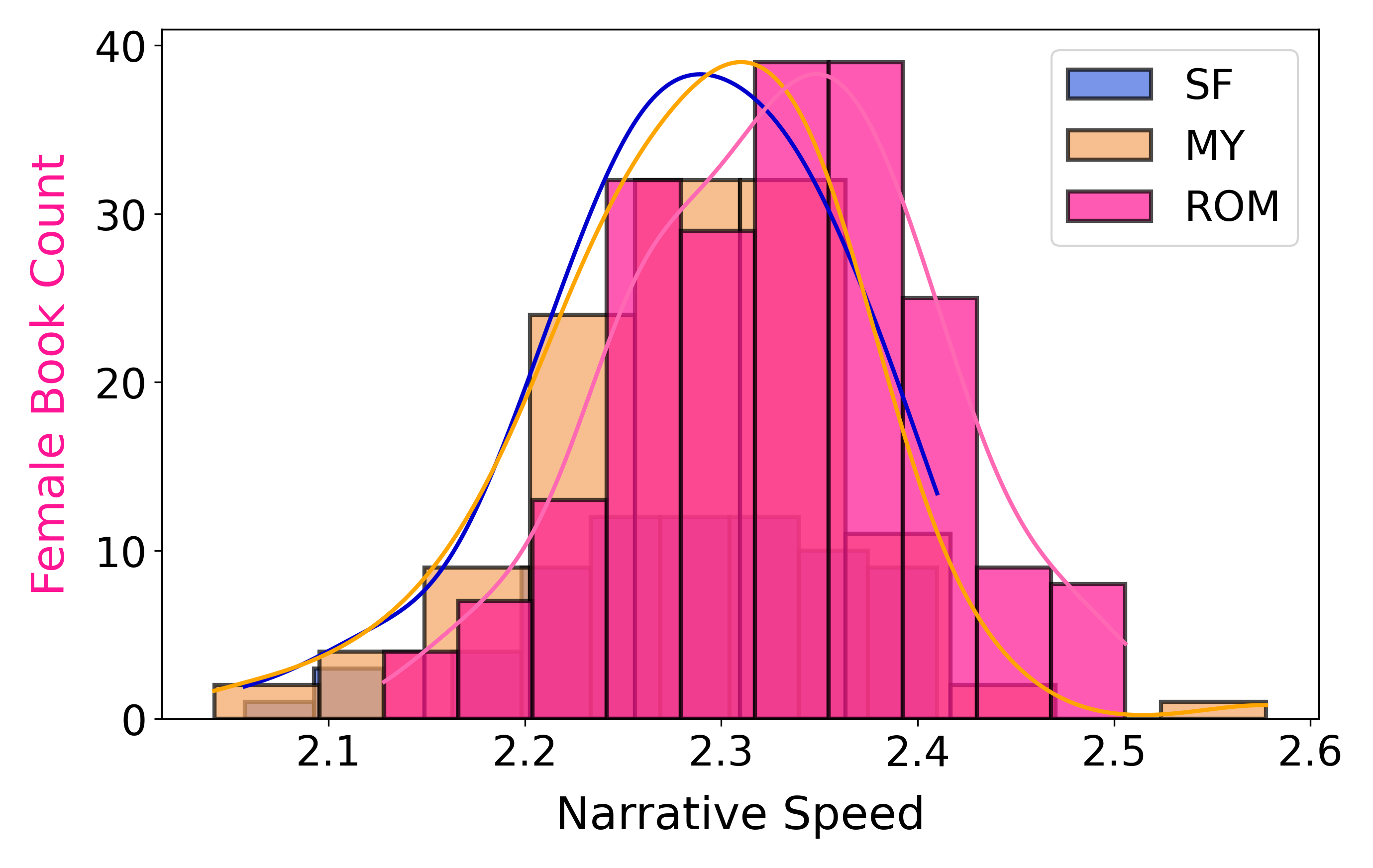}
  \includegraphics[width=0.45\linewidth]{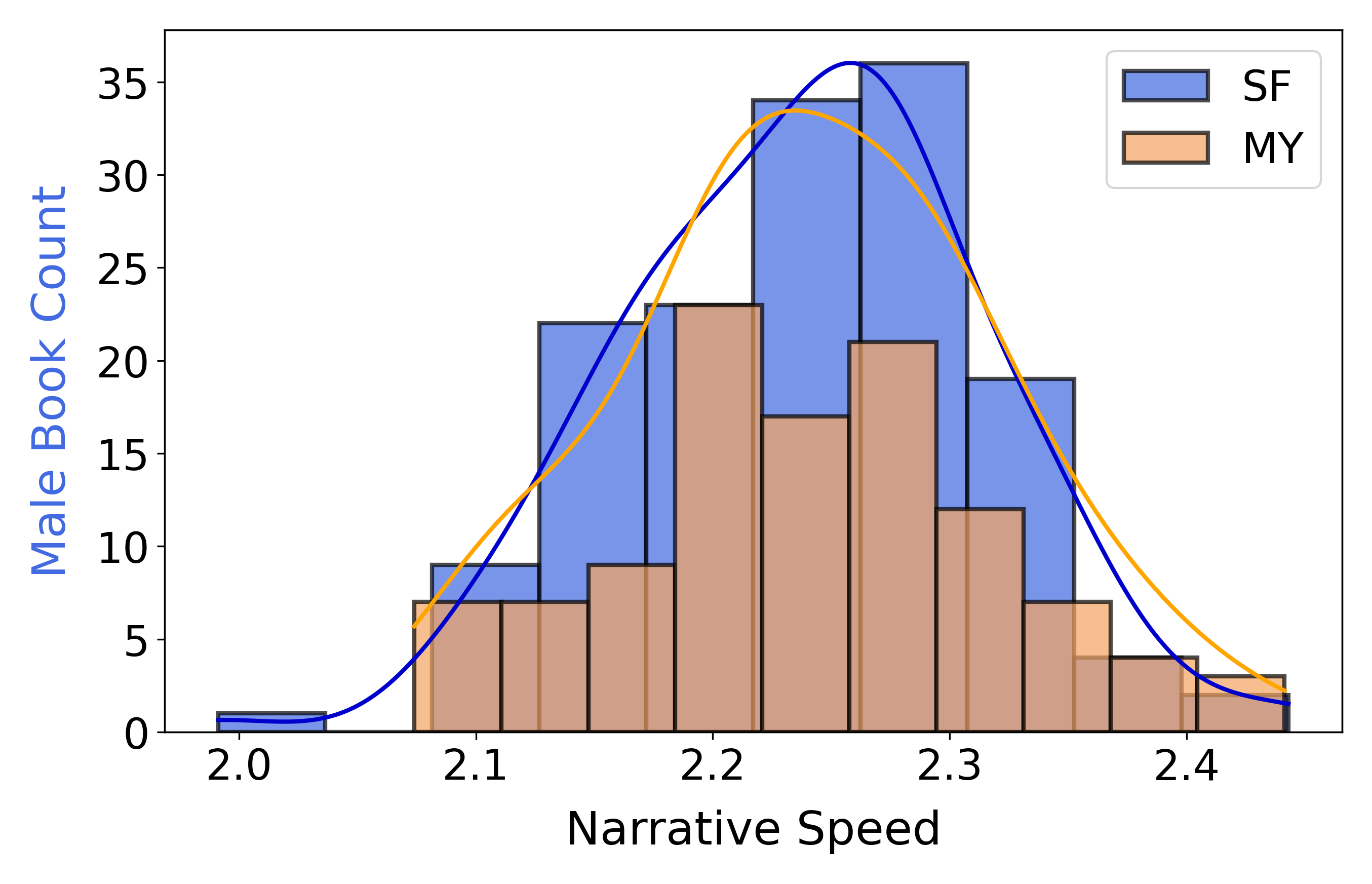}
  \centering
  \includegraphics[width=0.45\linewidth]{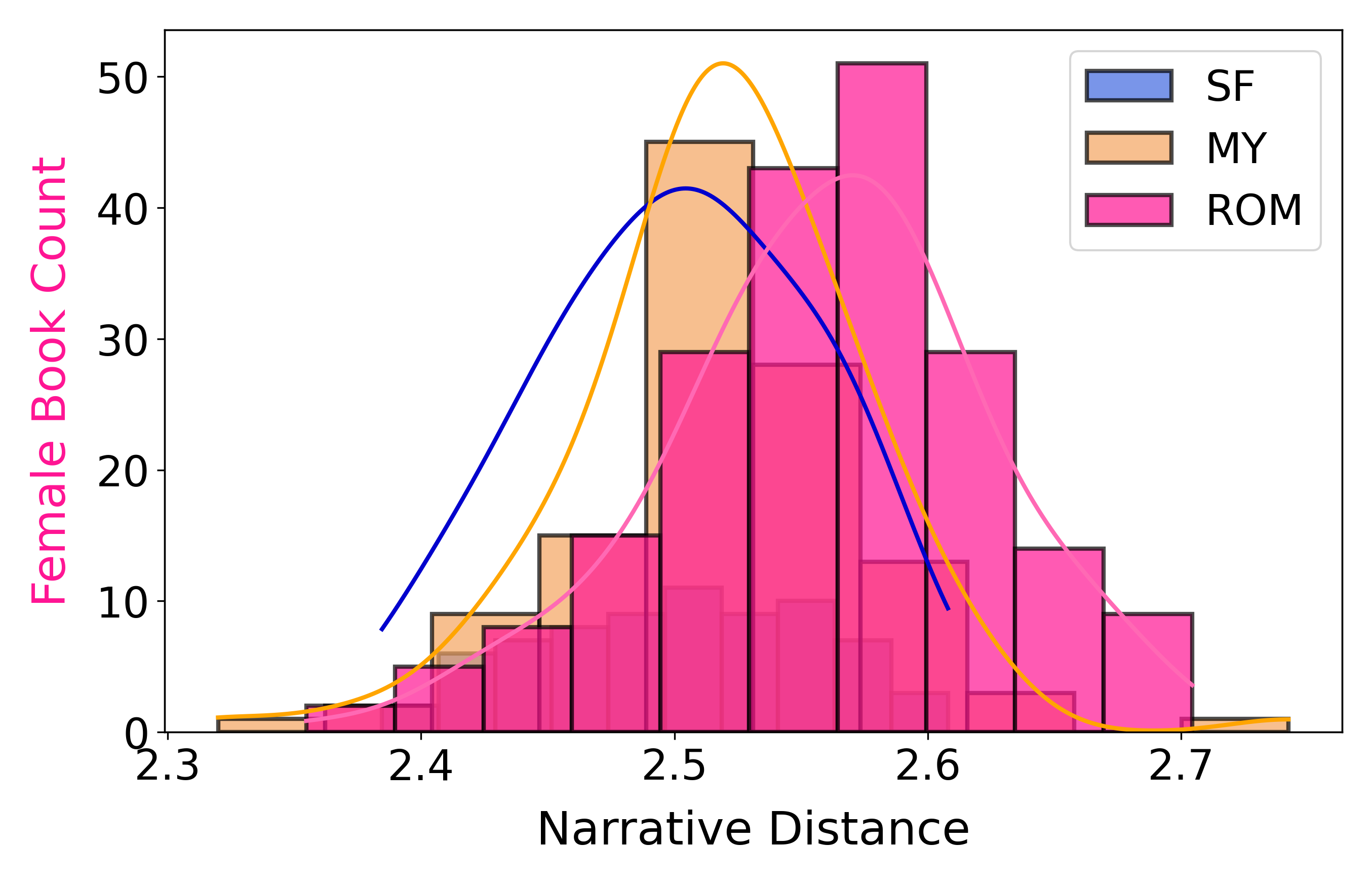}
  \includegraphics[width=0.45\linewidth]{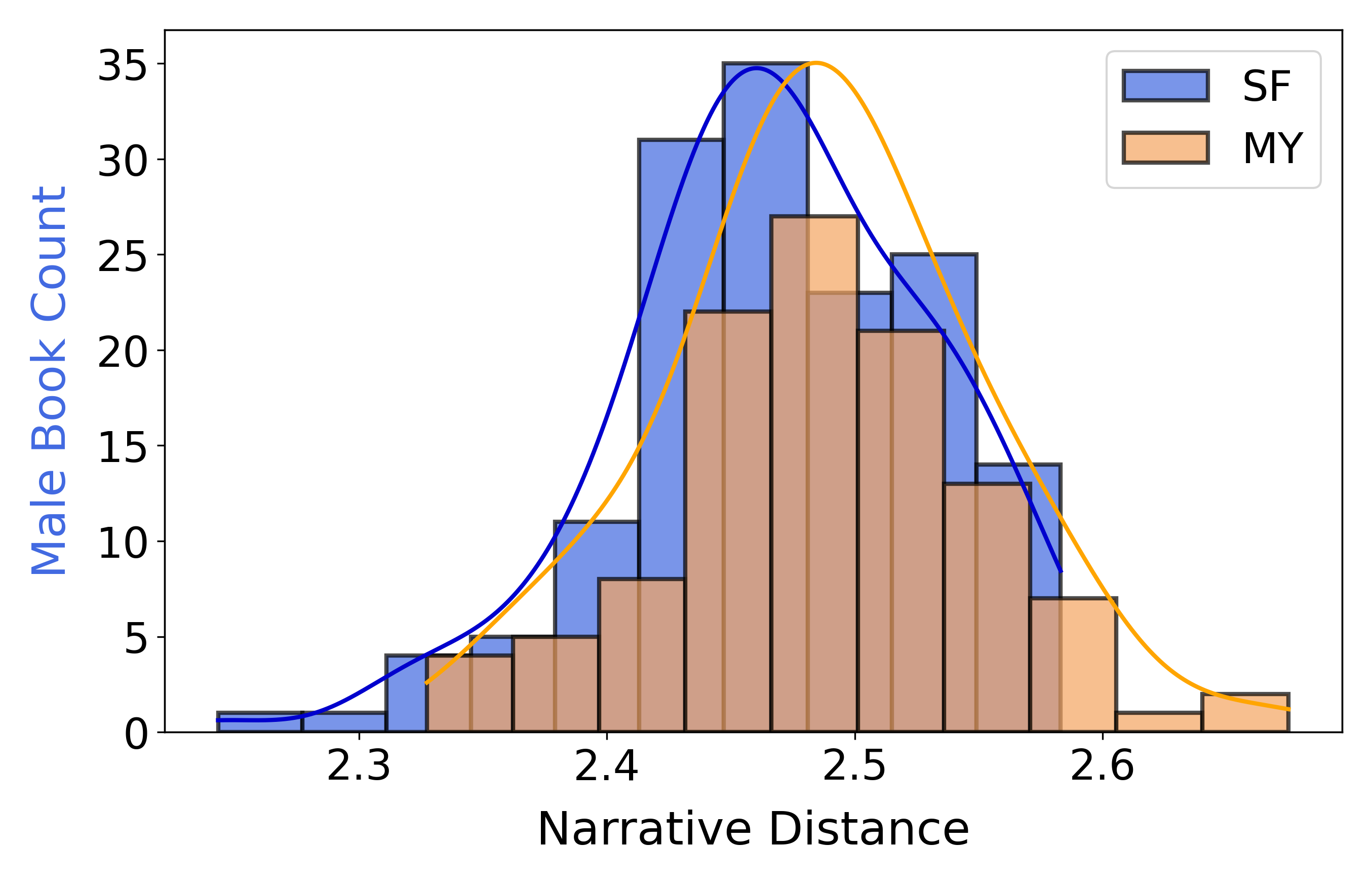}
  \centering
  \includegraphics[width=0.45\linewidth]{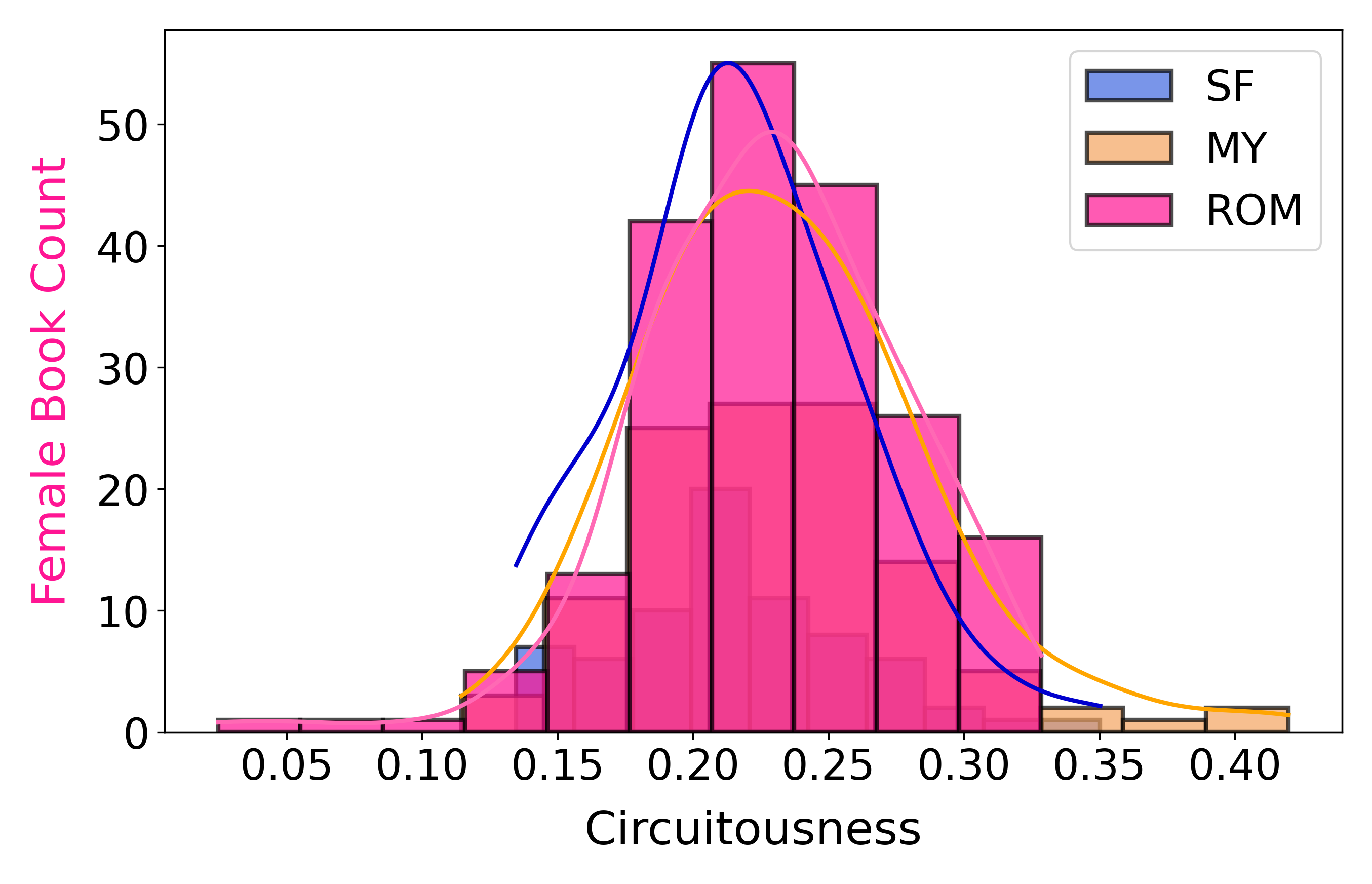}
  \includegraphics[width=0.45\linewidth]{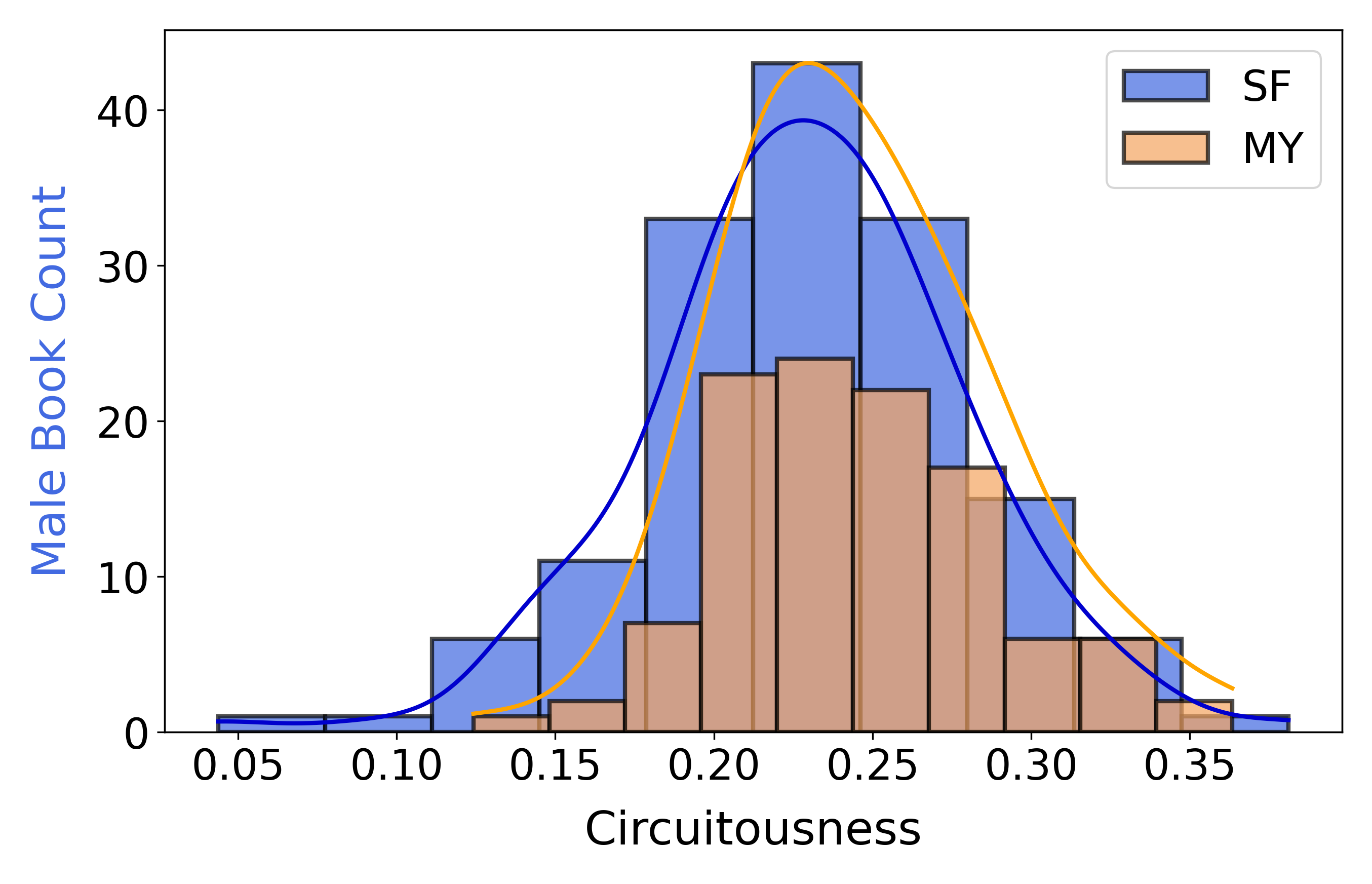}
  \centering
  \includegraphics[width=0.45\linewidth]{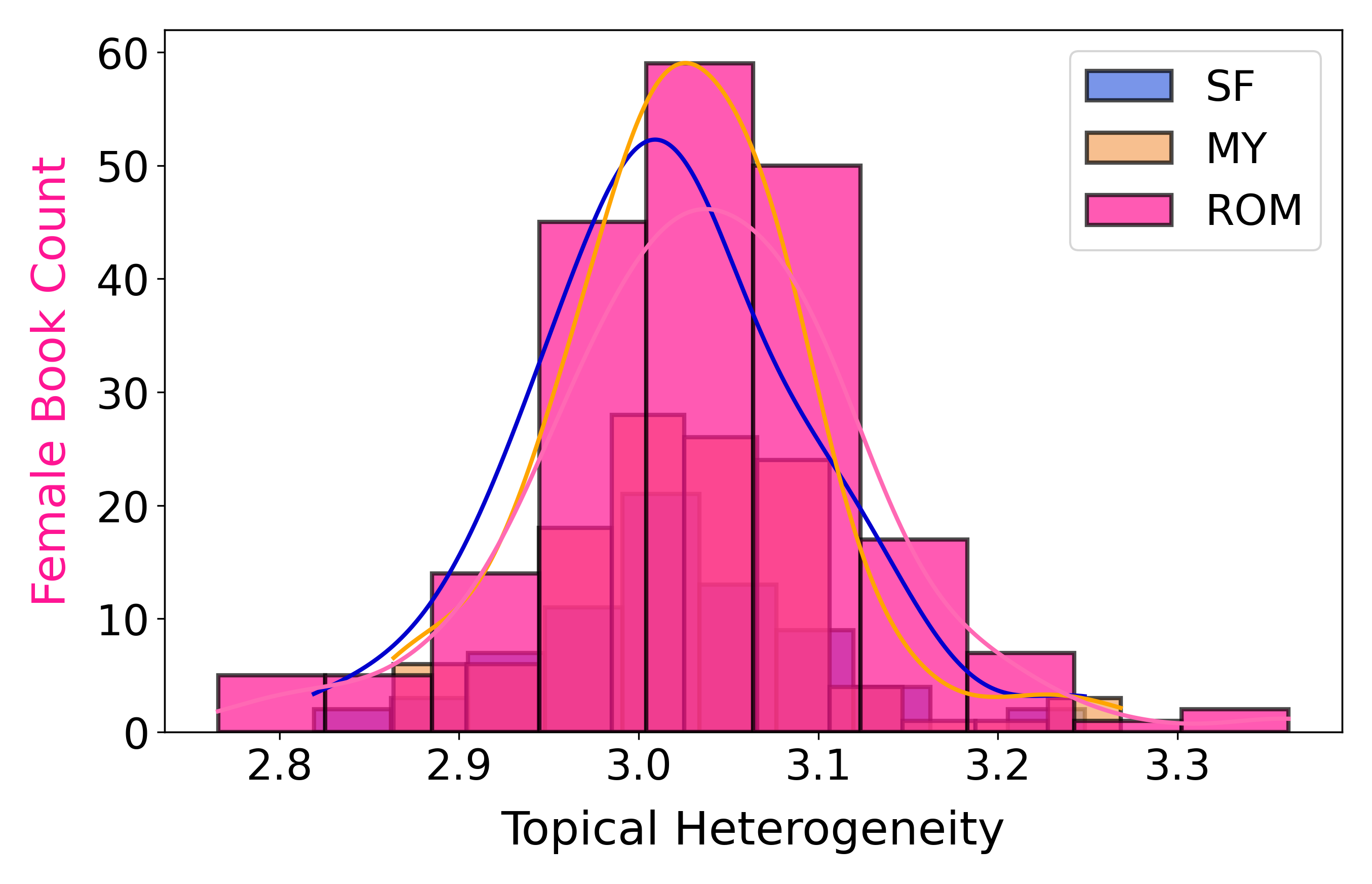}
  \includegraphics[width=0.45\linewidth]{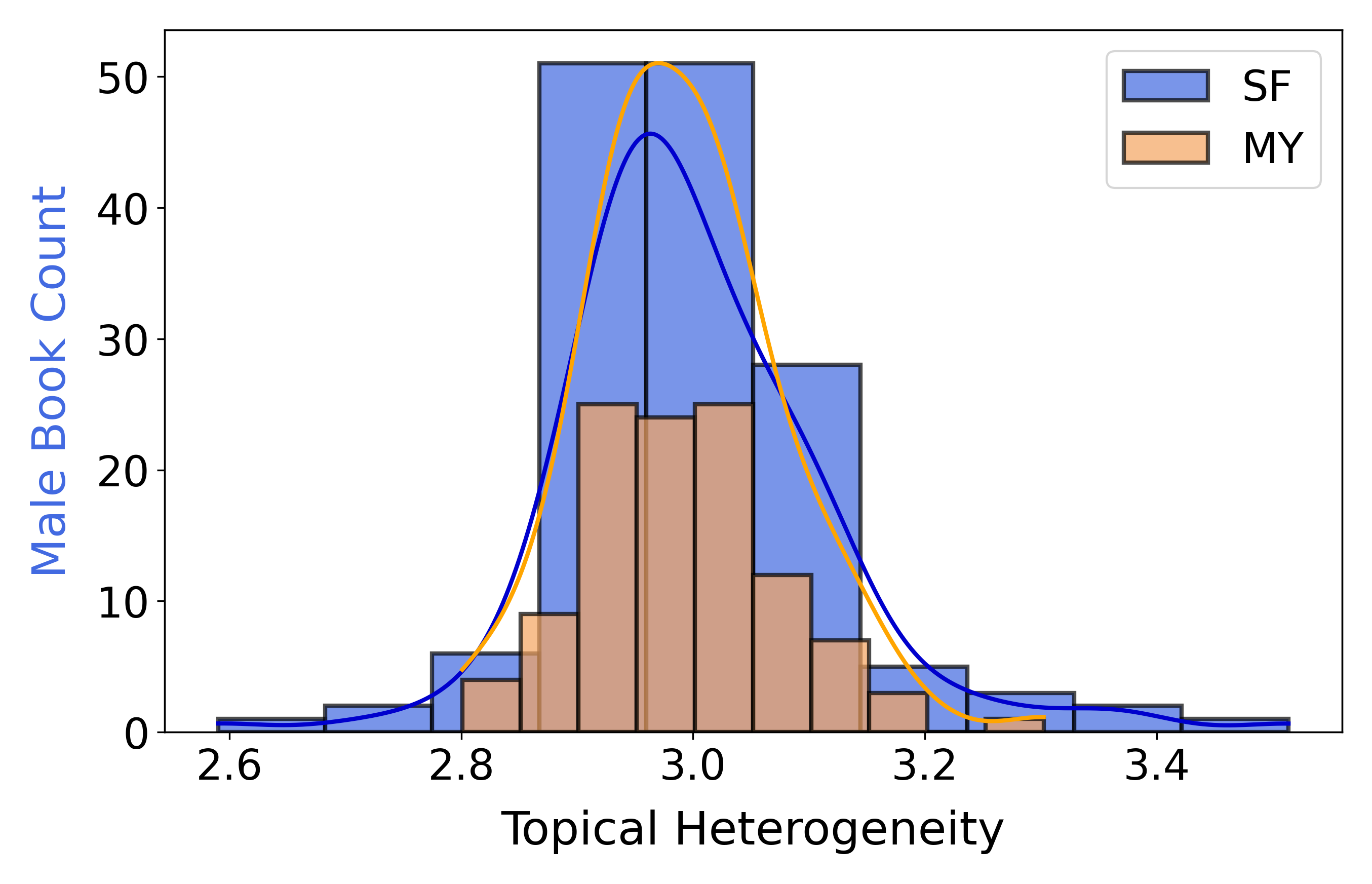}
  \centering
  \includegraphics[width=0.45\linewidth]{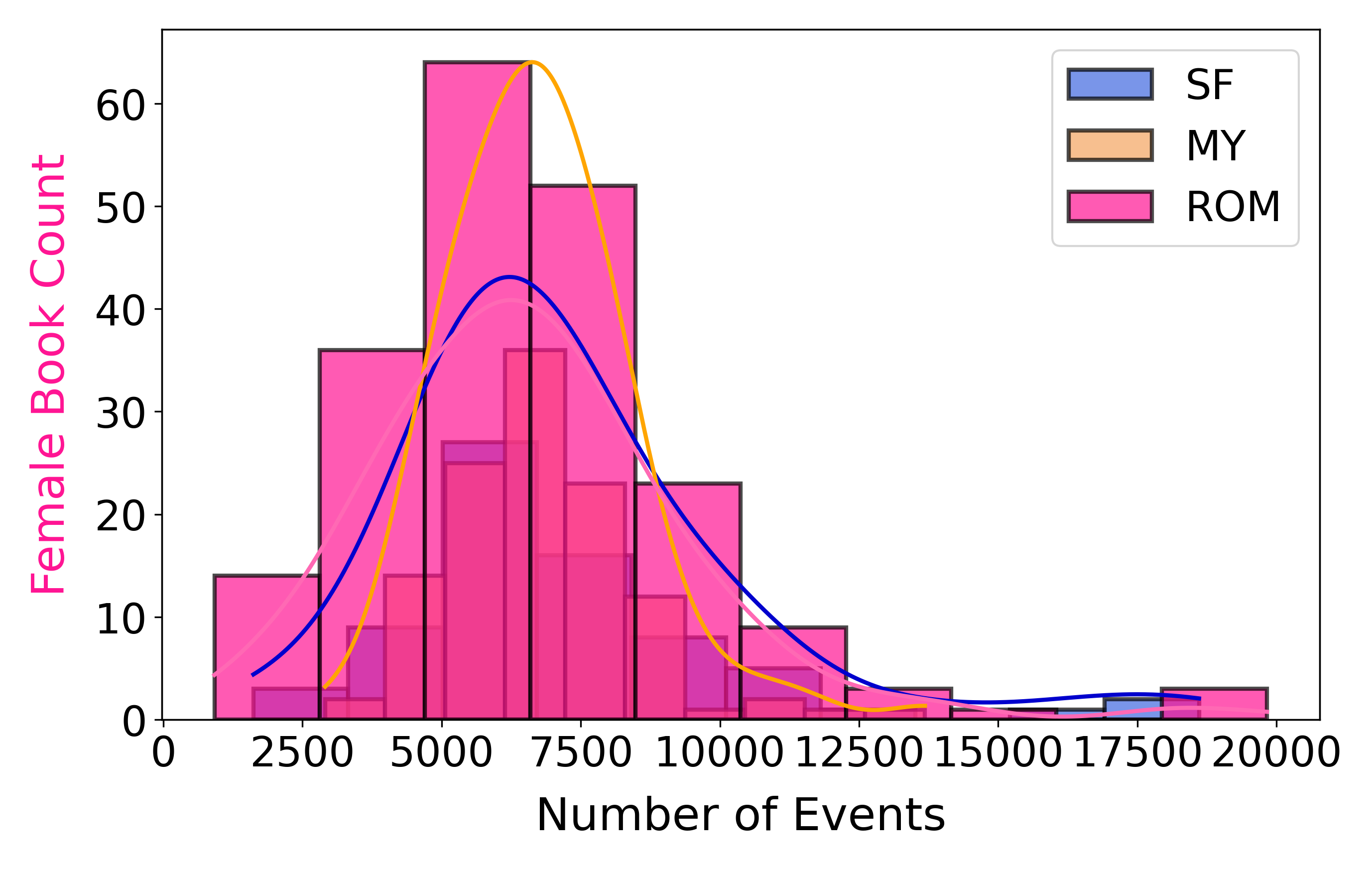}
  \includegraphics[width=0.45\linewidth]{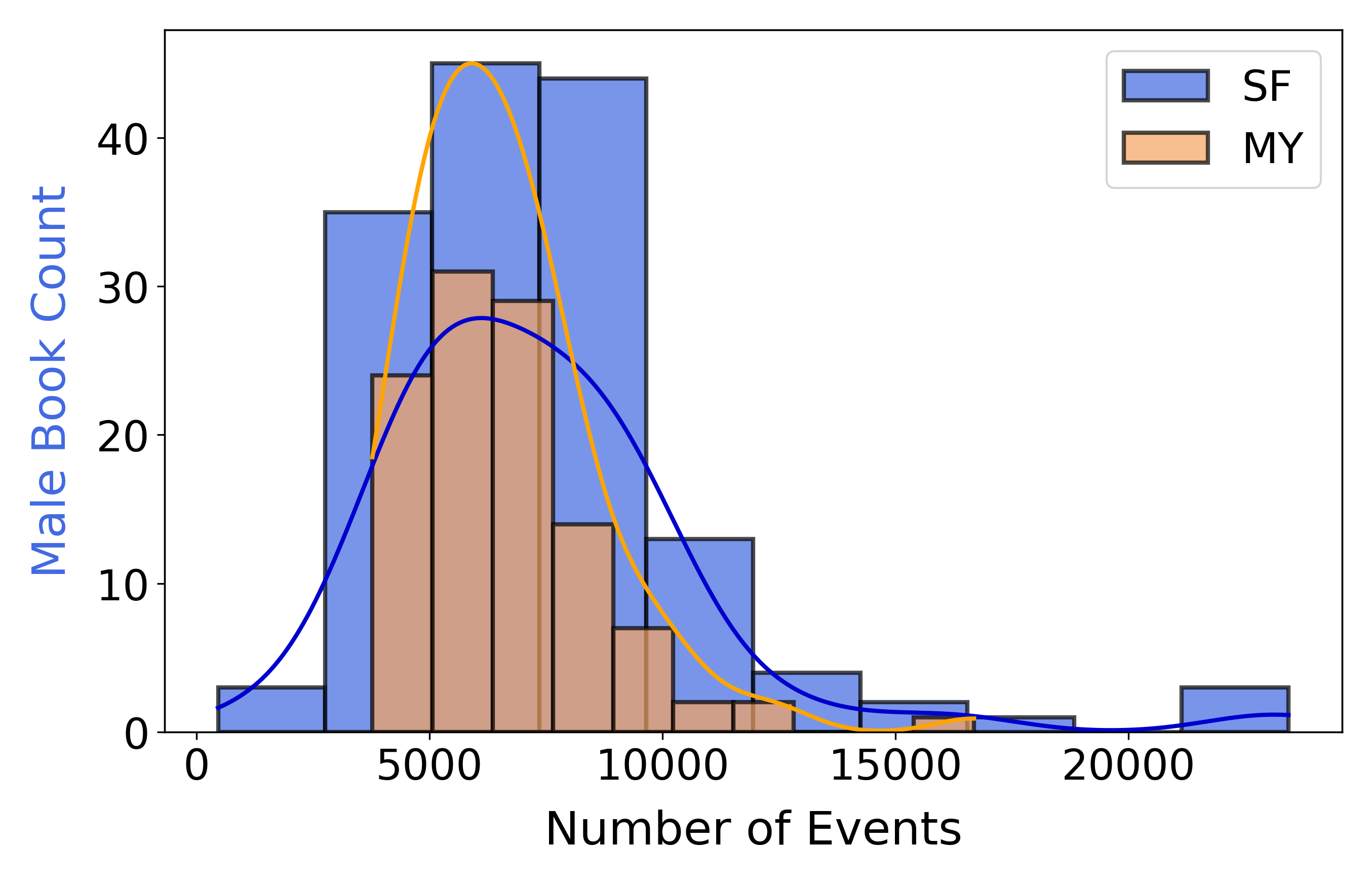}
    \caption{Narrative structure histograms for genre fiction by author gender}
  \label{fig:hist2}
\end{figure}

\begin{figure}[H]
  \centering
  \includegraphics[width=0.45\linewidth]{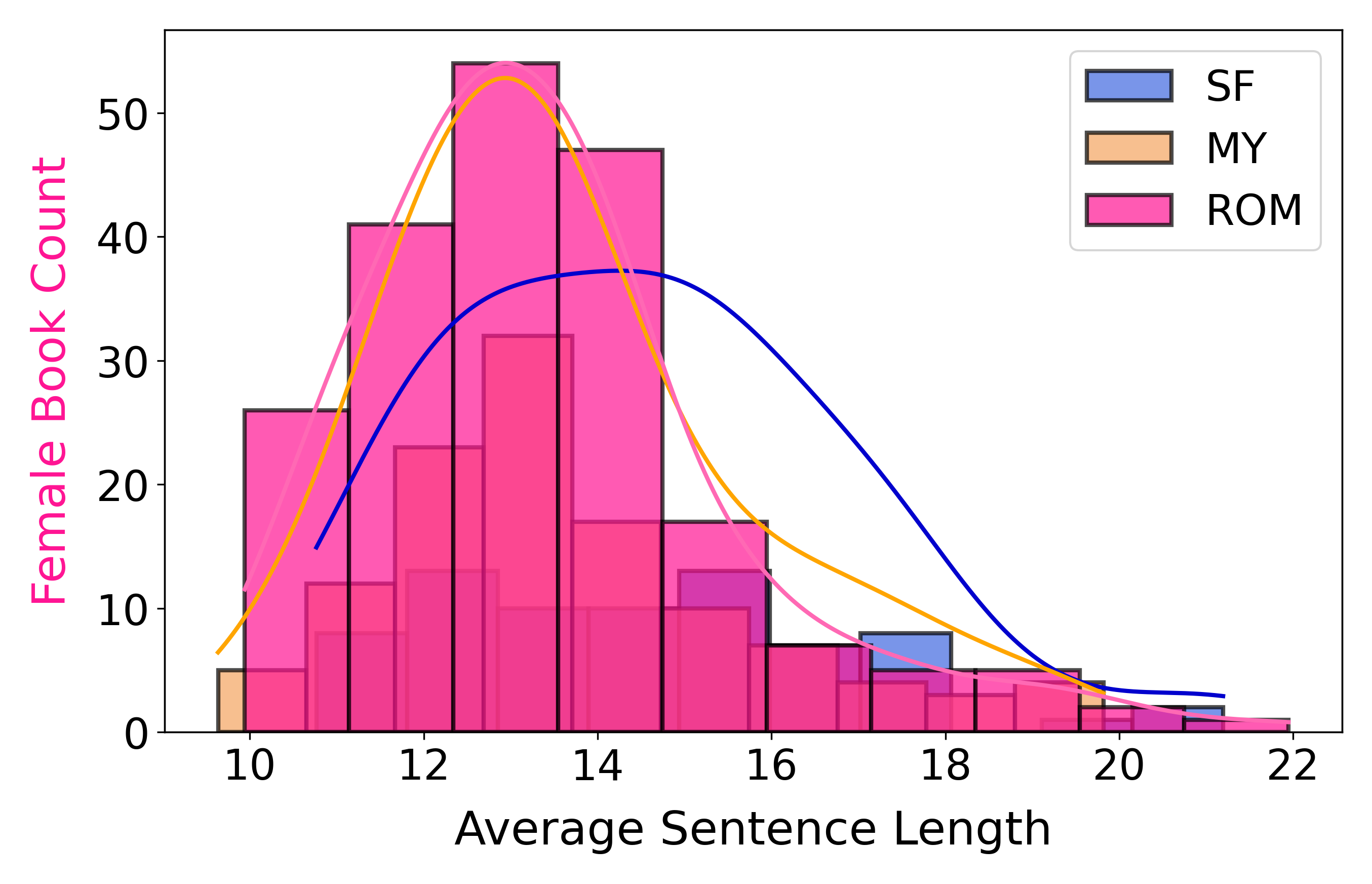}
  \includegraphics[width=0.45\linewidth]{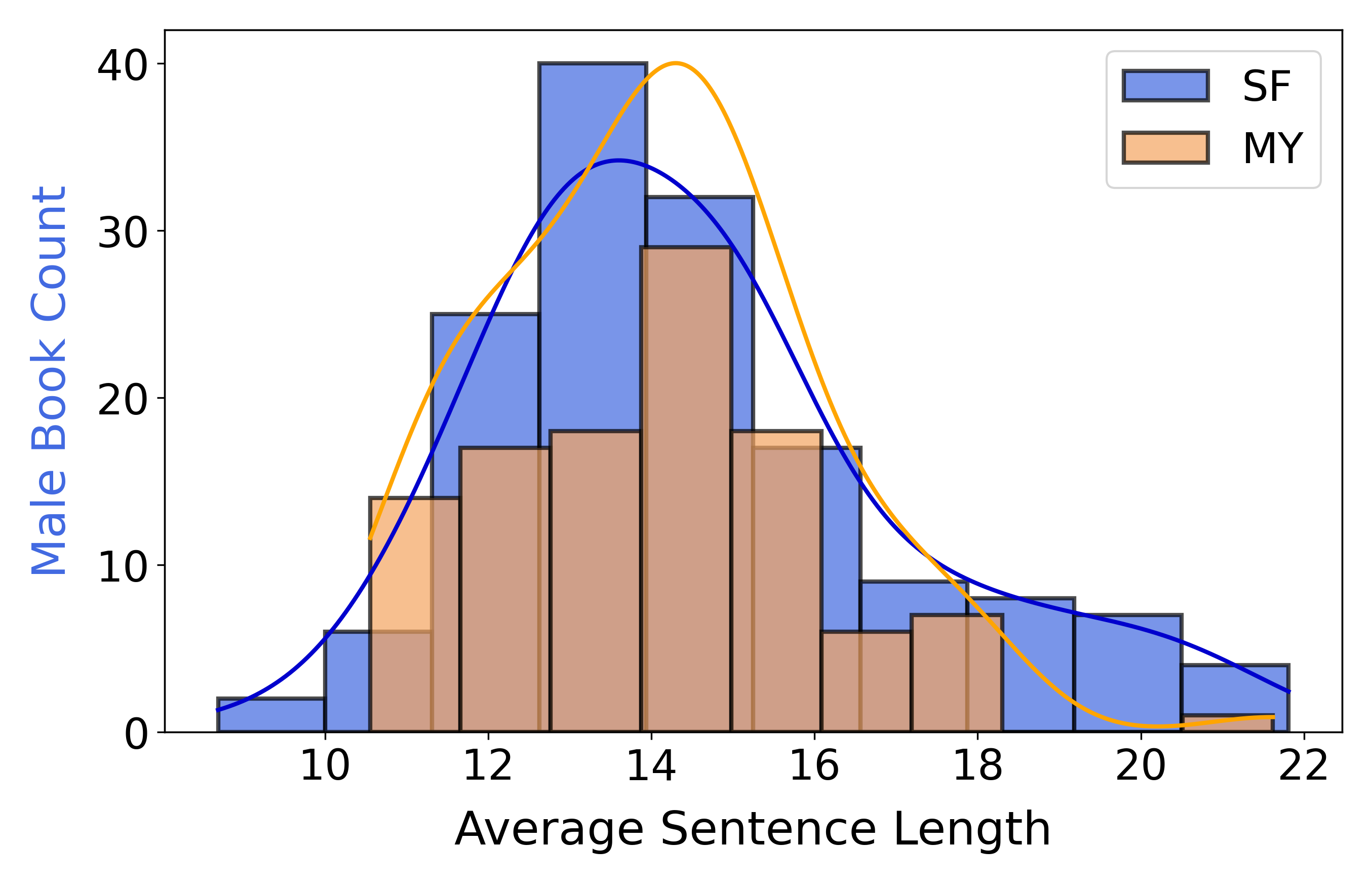}
  \centering
  \includegraphics[width=0.45\linewidth]{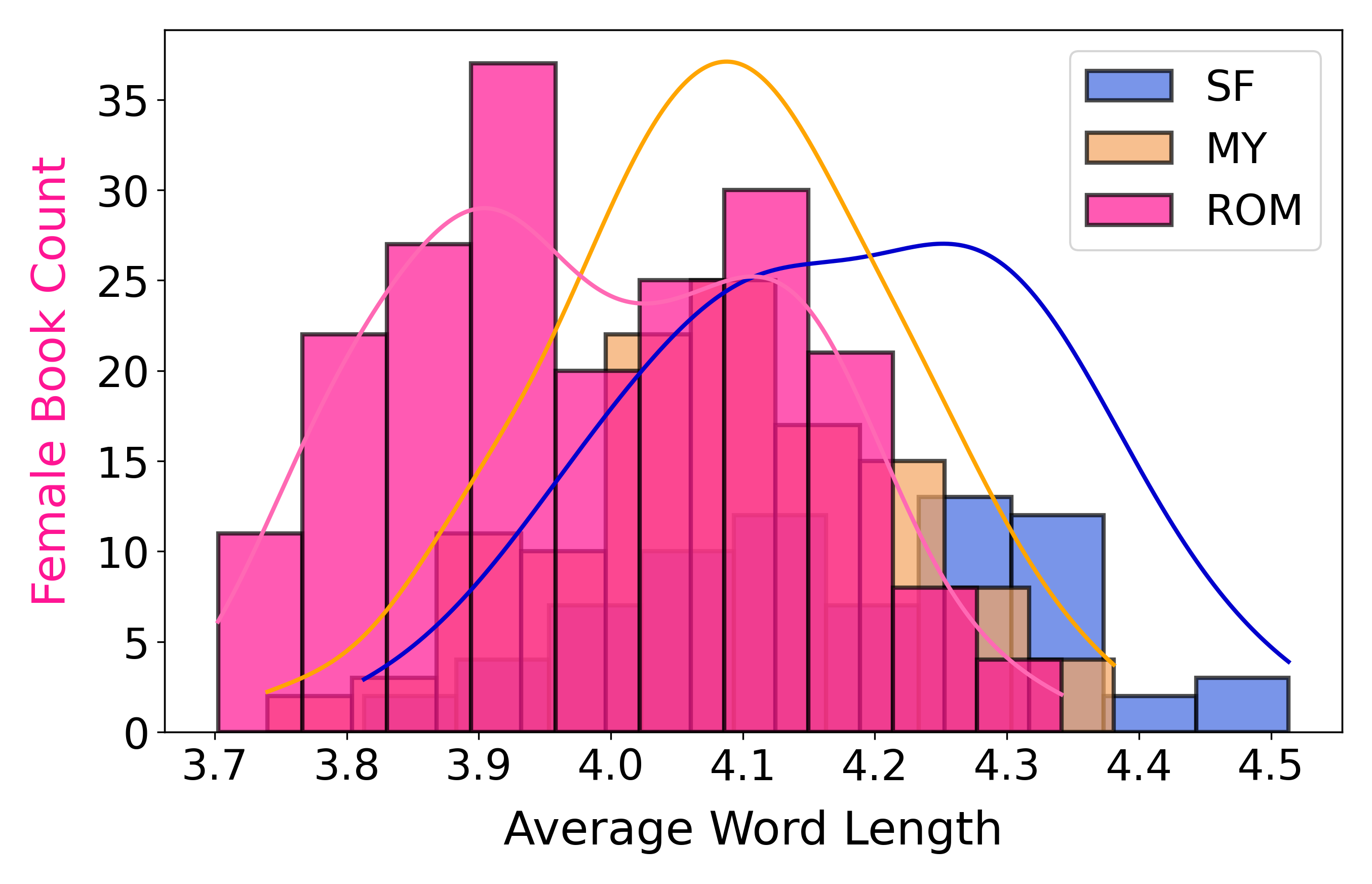}
  \includegraphics[width=0.45\linewidth]{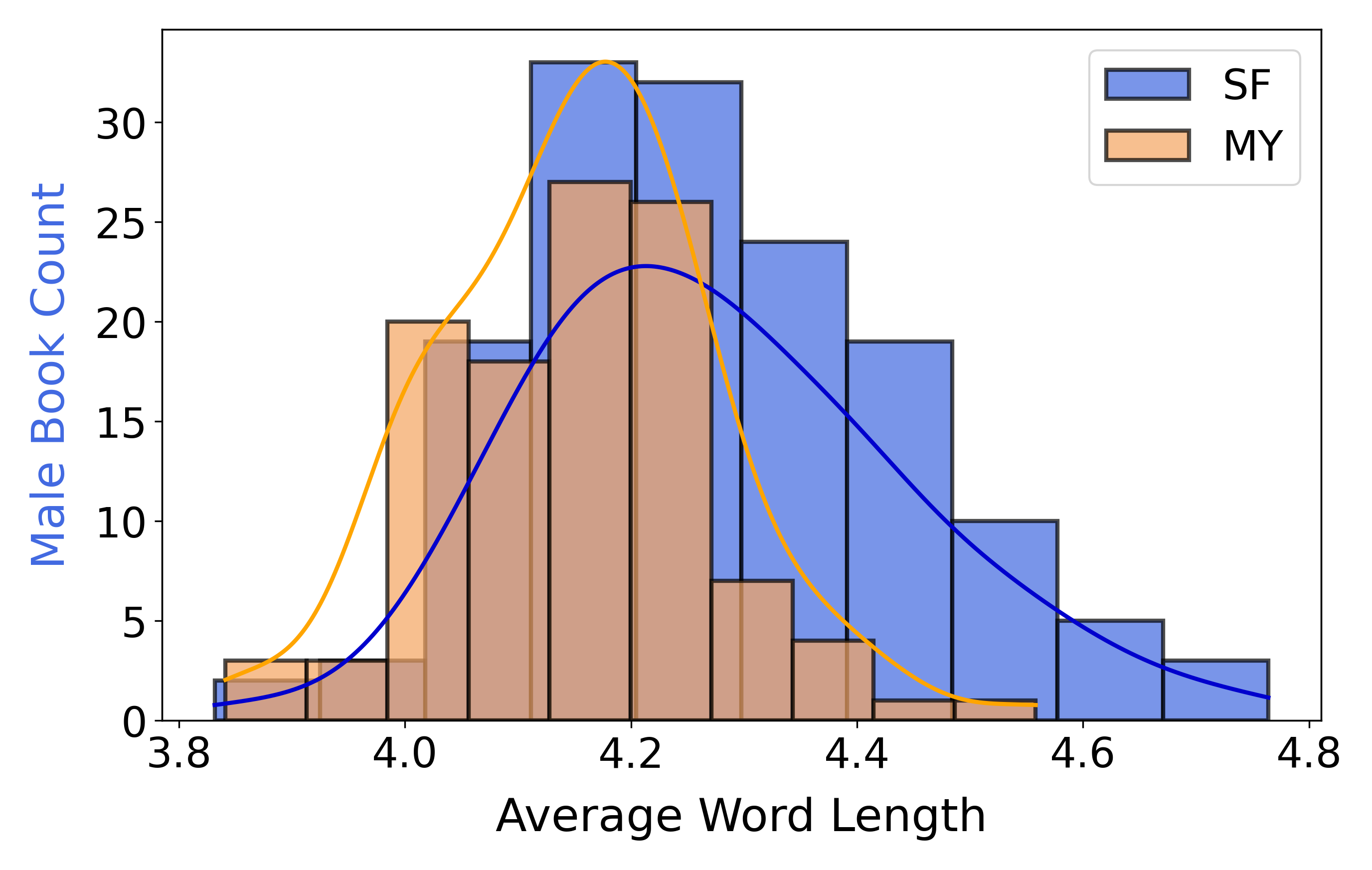}
  \centering
  \includegraphics[width=0.45\linewidth]{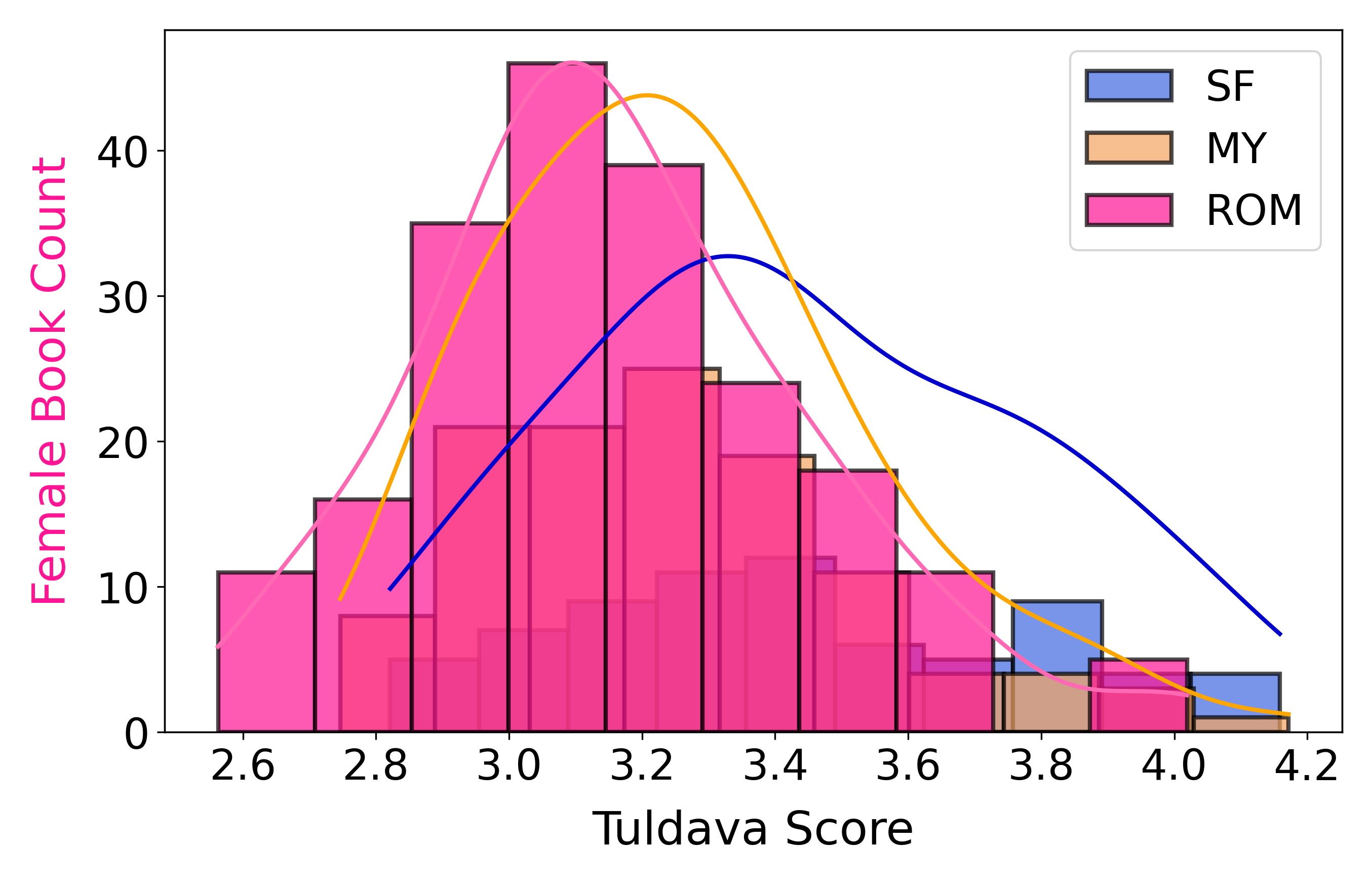}
  \includegraphics[width=0.45\linewidth]{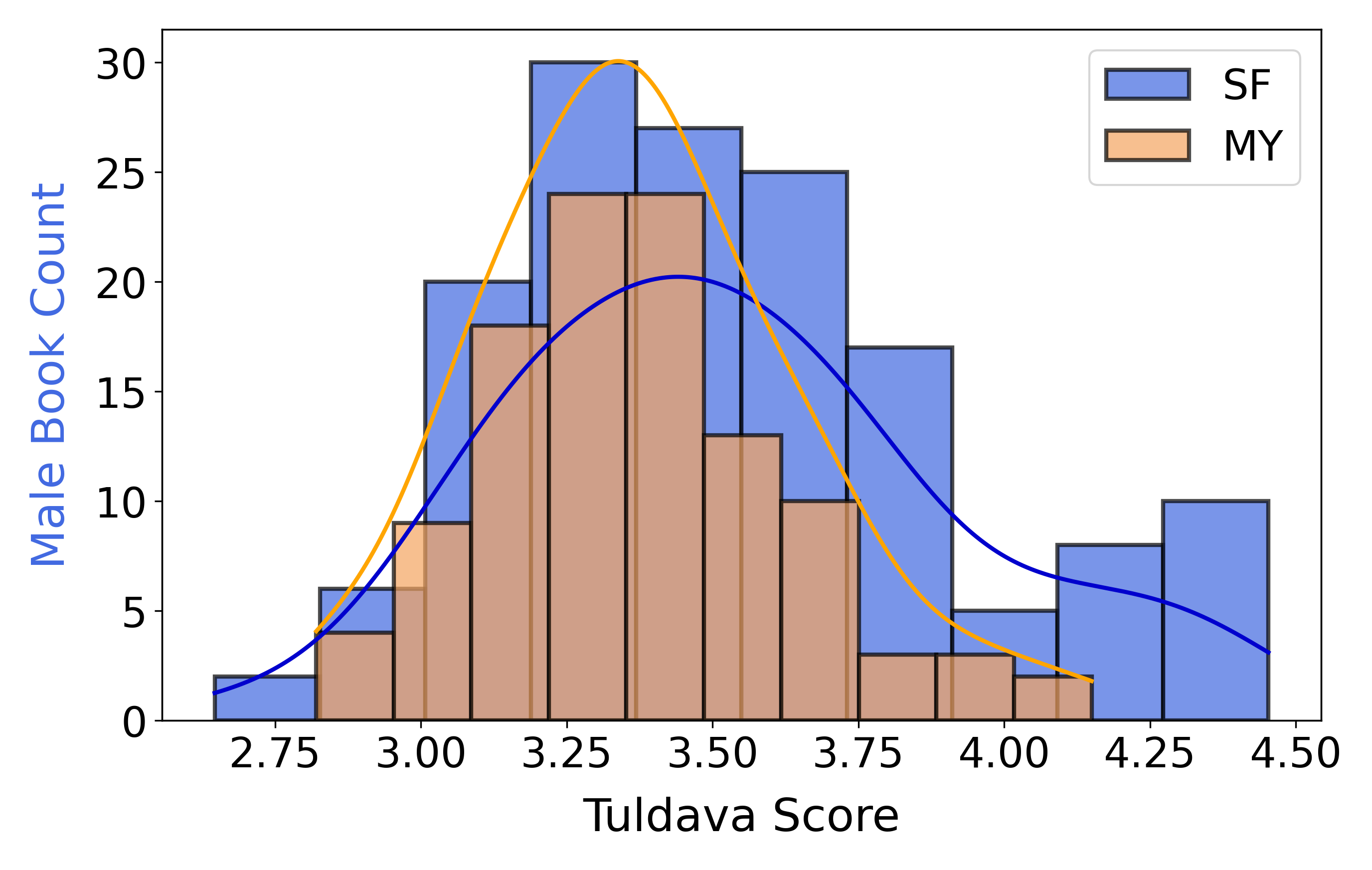}
  \centering
  \includegraphics[width=0.45\linewidth]{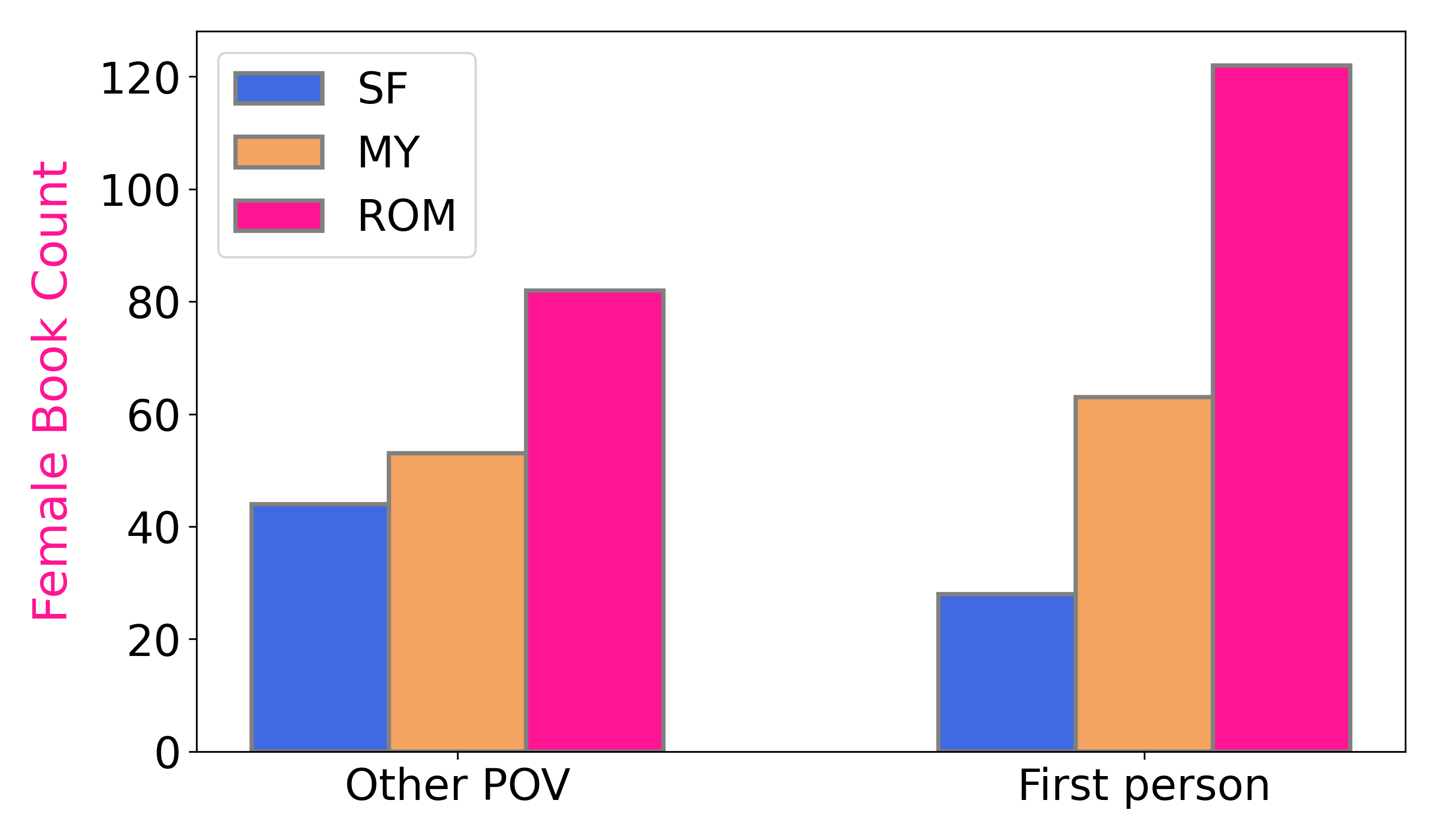}
  \includegraphics[width=0.45\linewidth]{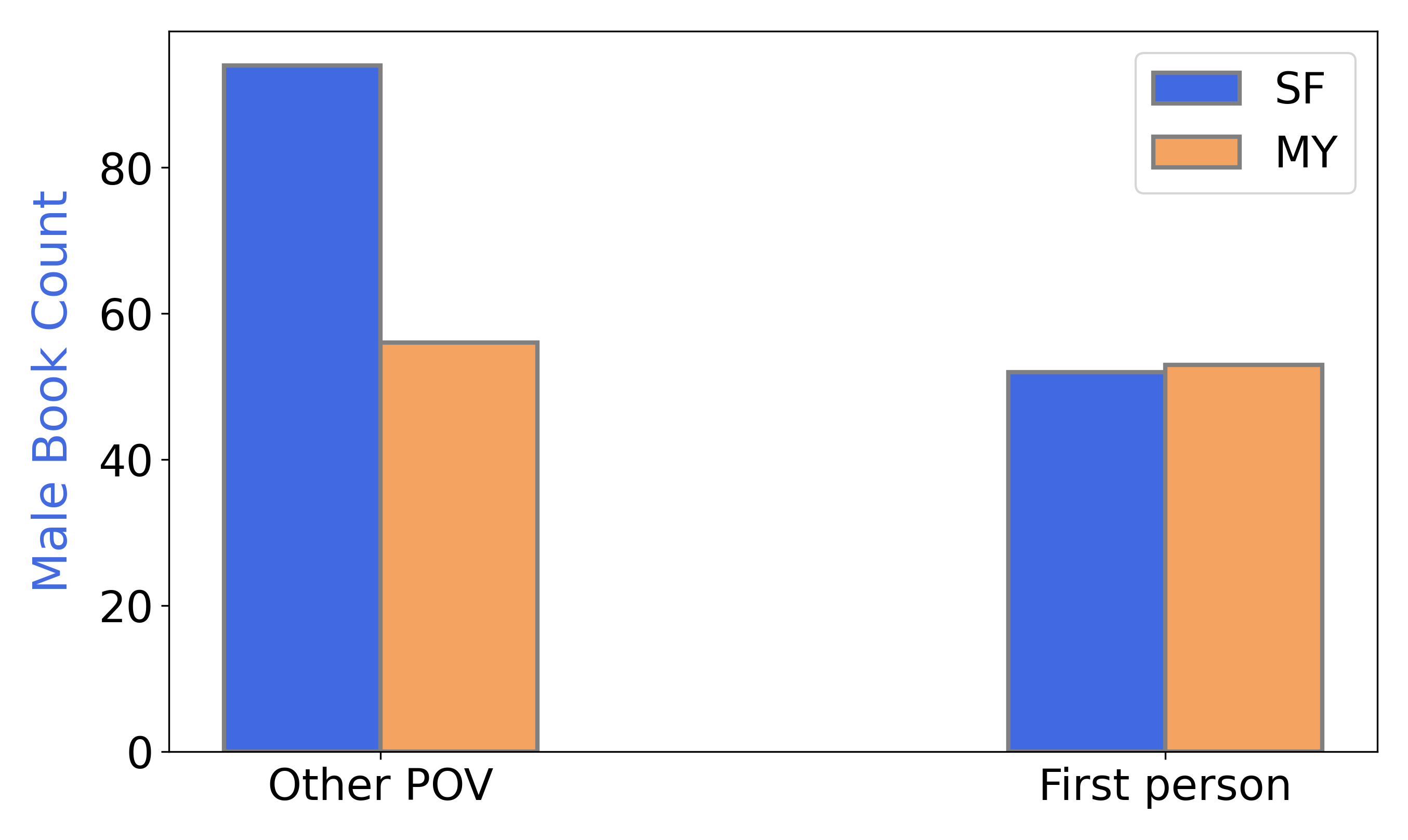}
    \caption{Language histograms for genre fiction by author gender}
  \label{fig:hist3}
\end{figure}

\begin{figure}[H]
  \centering
  \includegraphics[width=0.45\linewidth]{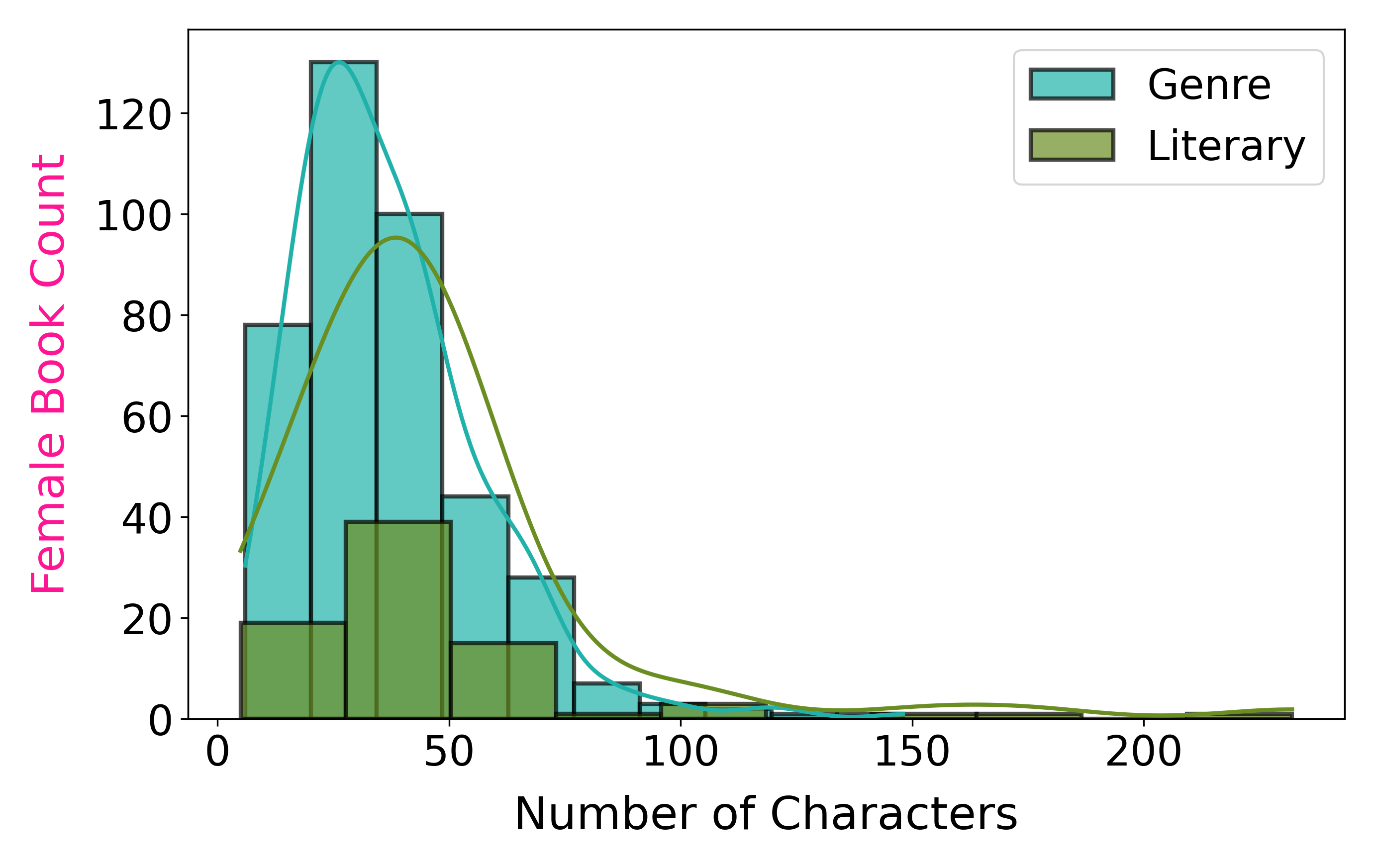}
  \includegraphics[width=0.45\linewidth]{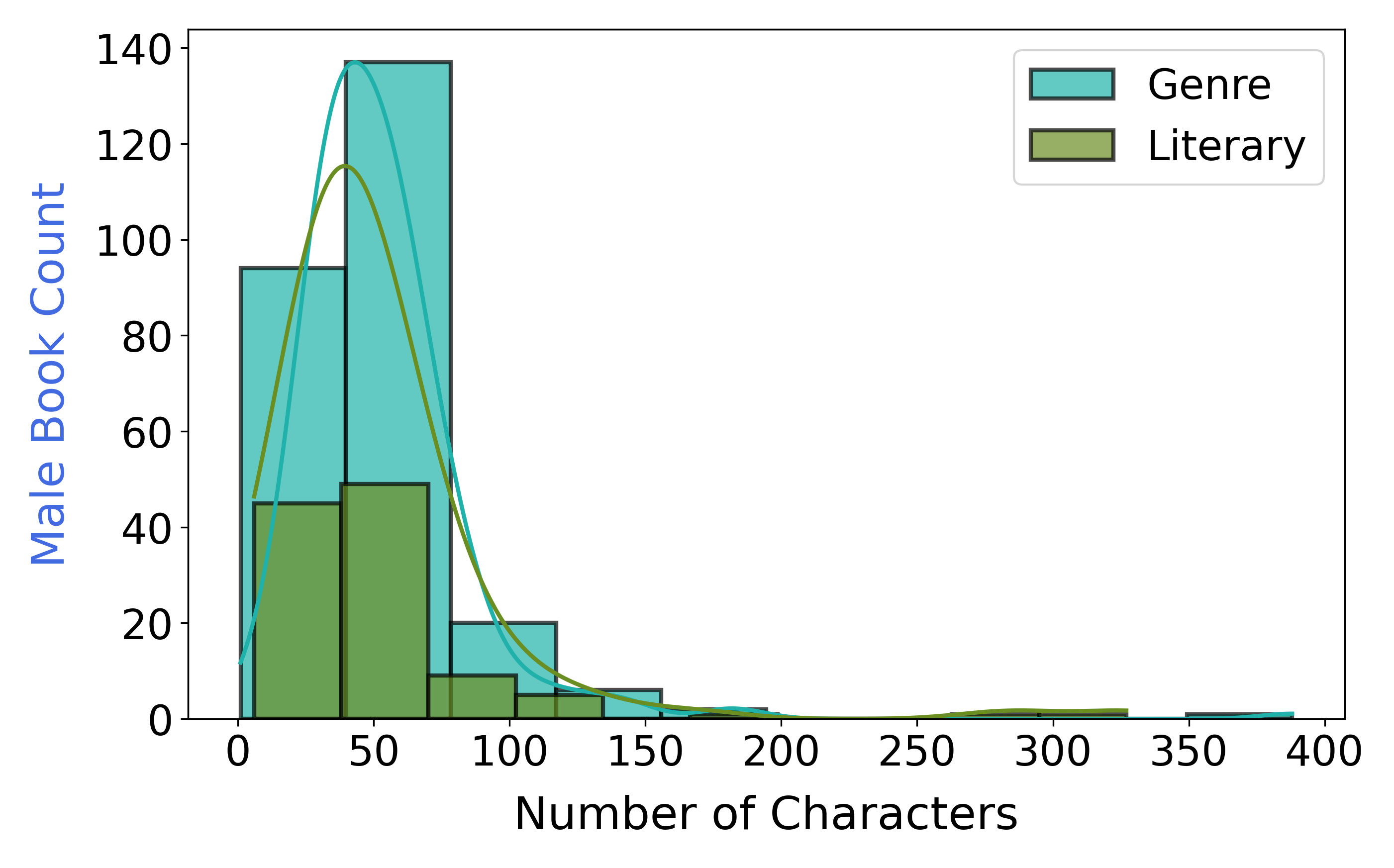}
\centering
  \includegraphics[width=0.45\linewidth]{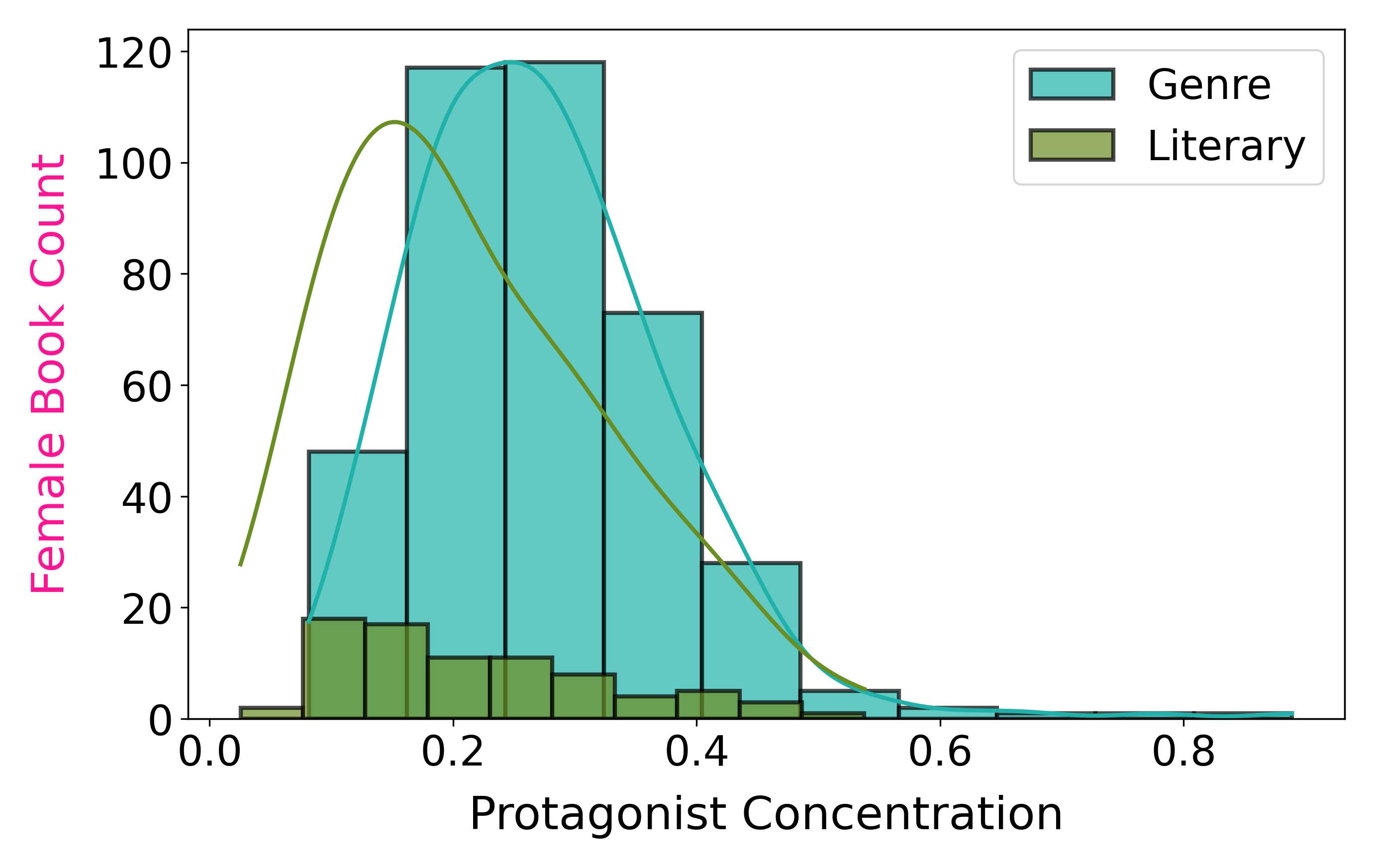}
  \includegraphics[width=0.45\linewidth]{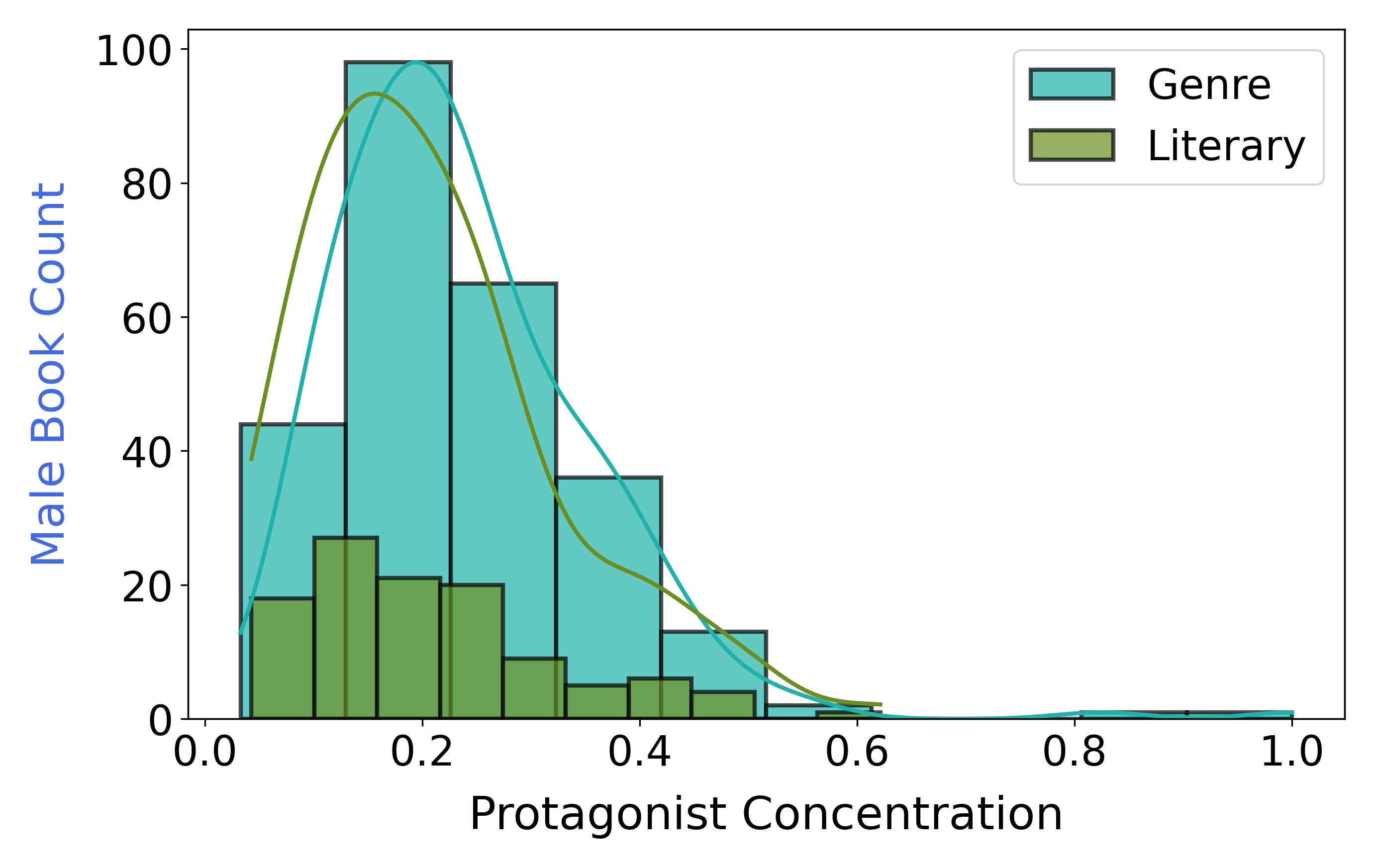}
    \caption{Character histograms for literary vs genre fiction by author gender}
  \label{fig:hist4}
\end{figure}

\begin{figure}[H]
  \centering
  \includegraphics[width=0.45\linewidth]{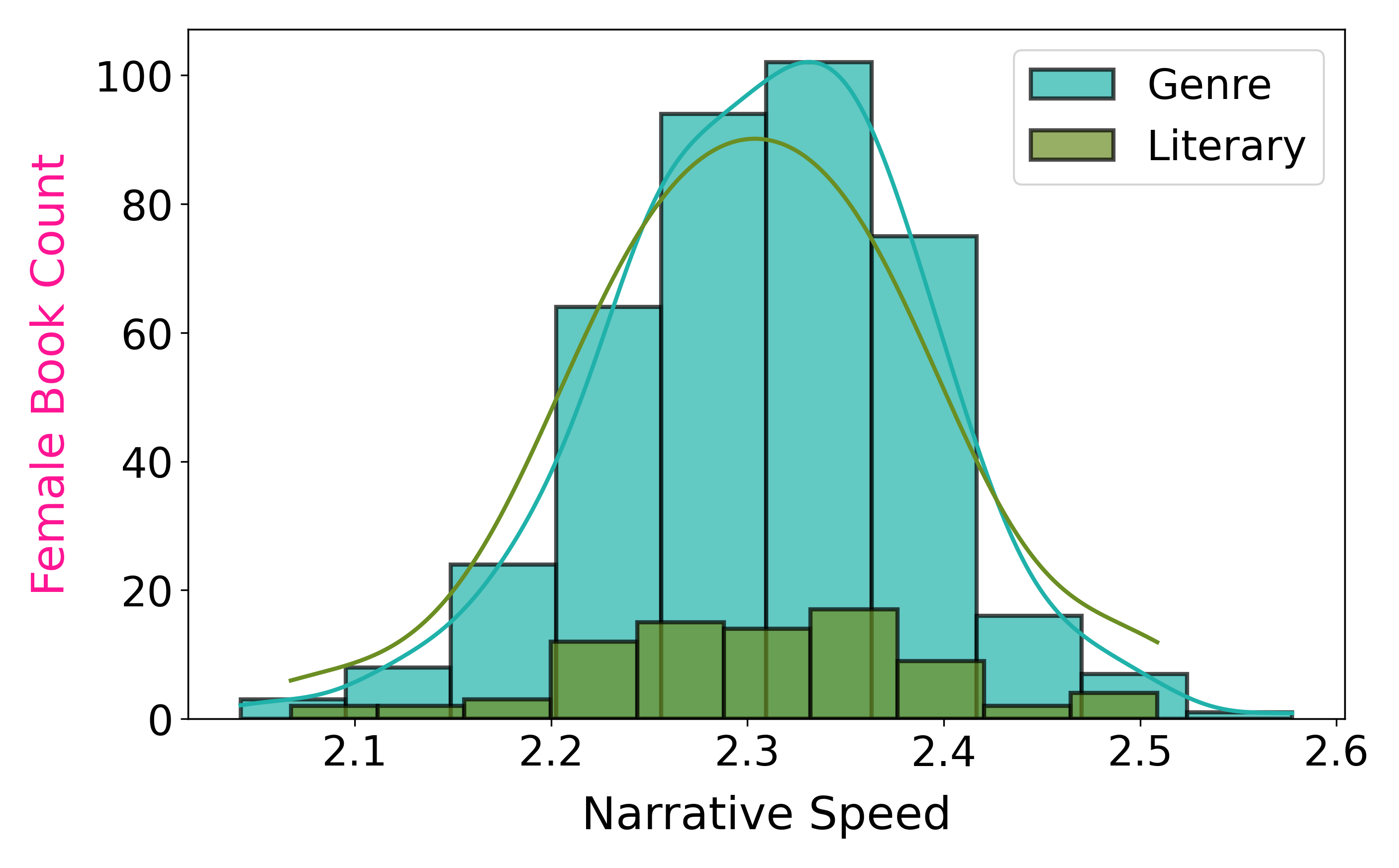}
  \includegraphics[width=0.45\linewidth]{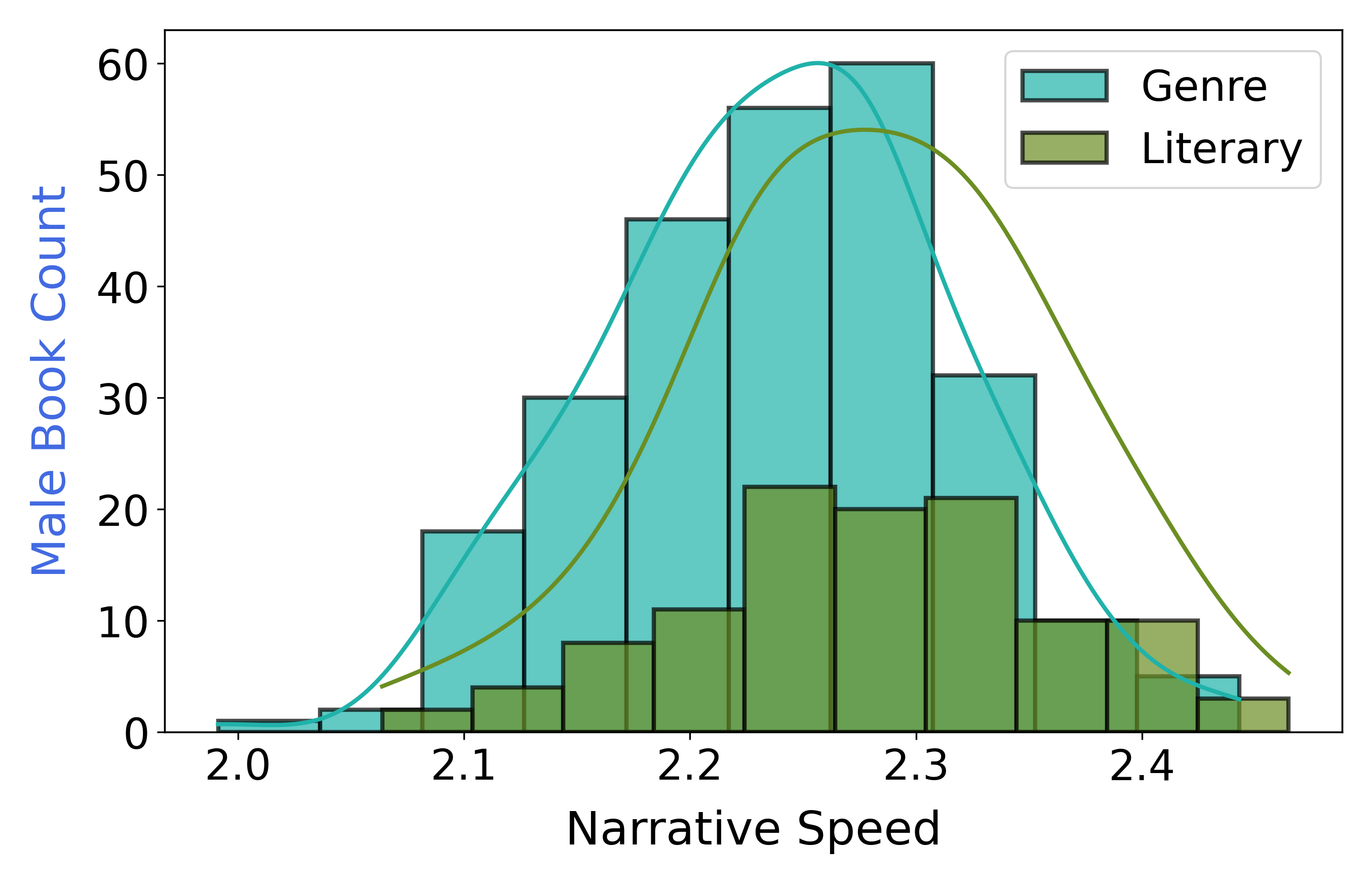}
  \centering
  \includegraphics[width=0.45\linewidth]{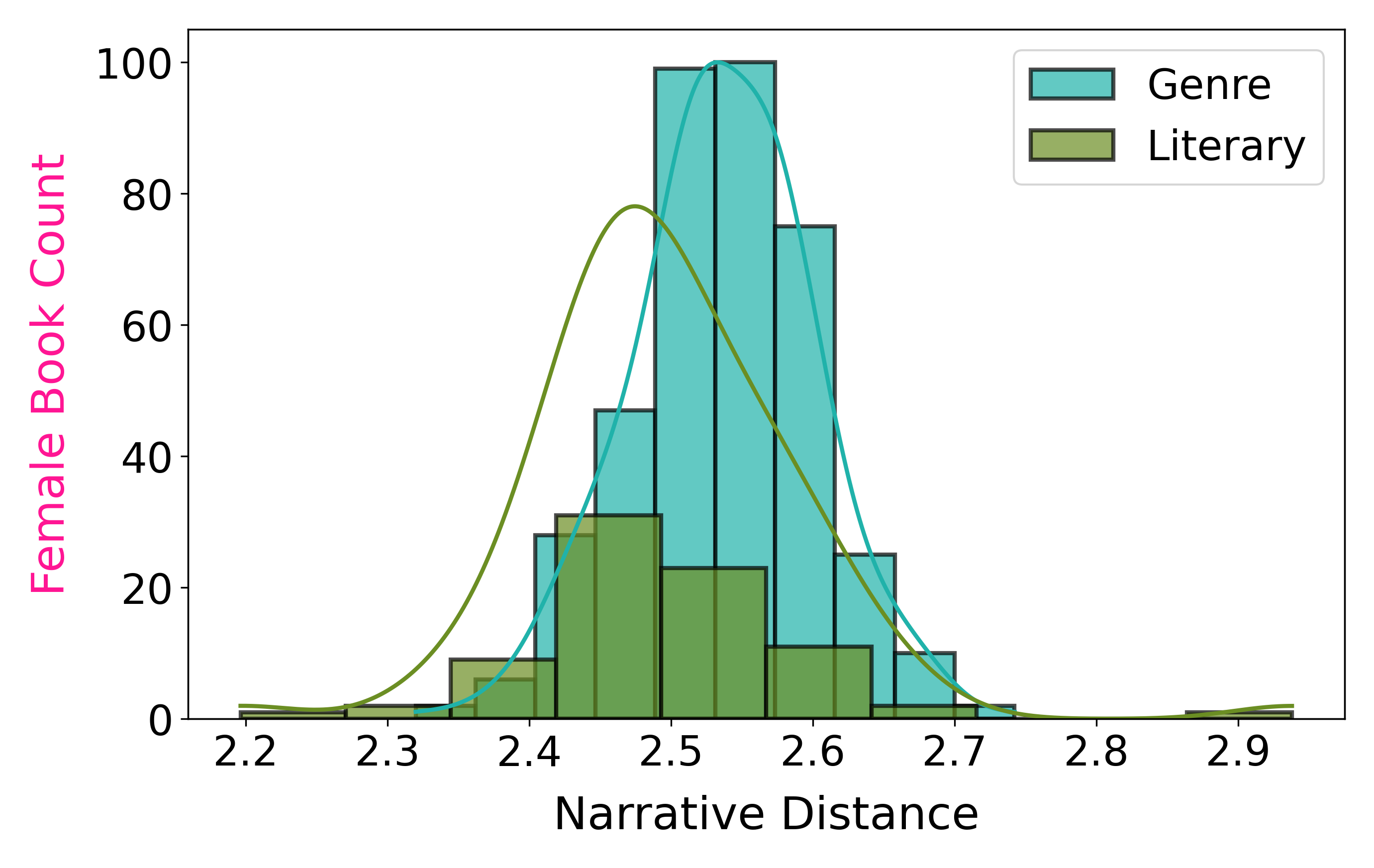}
  \includegraphics[width=0.45\linewidth]{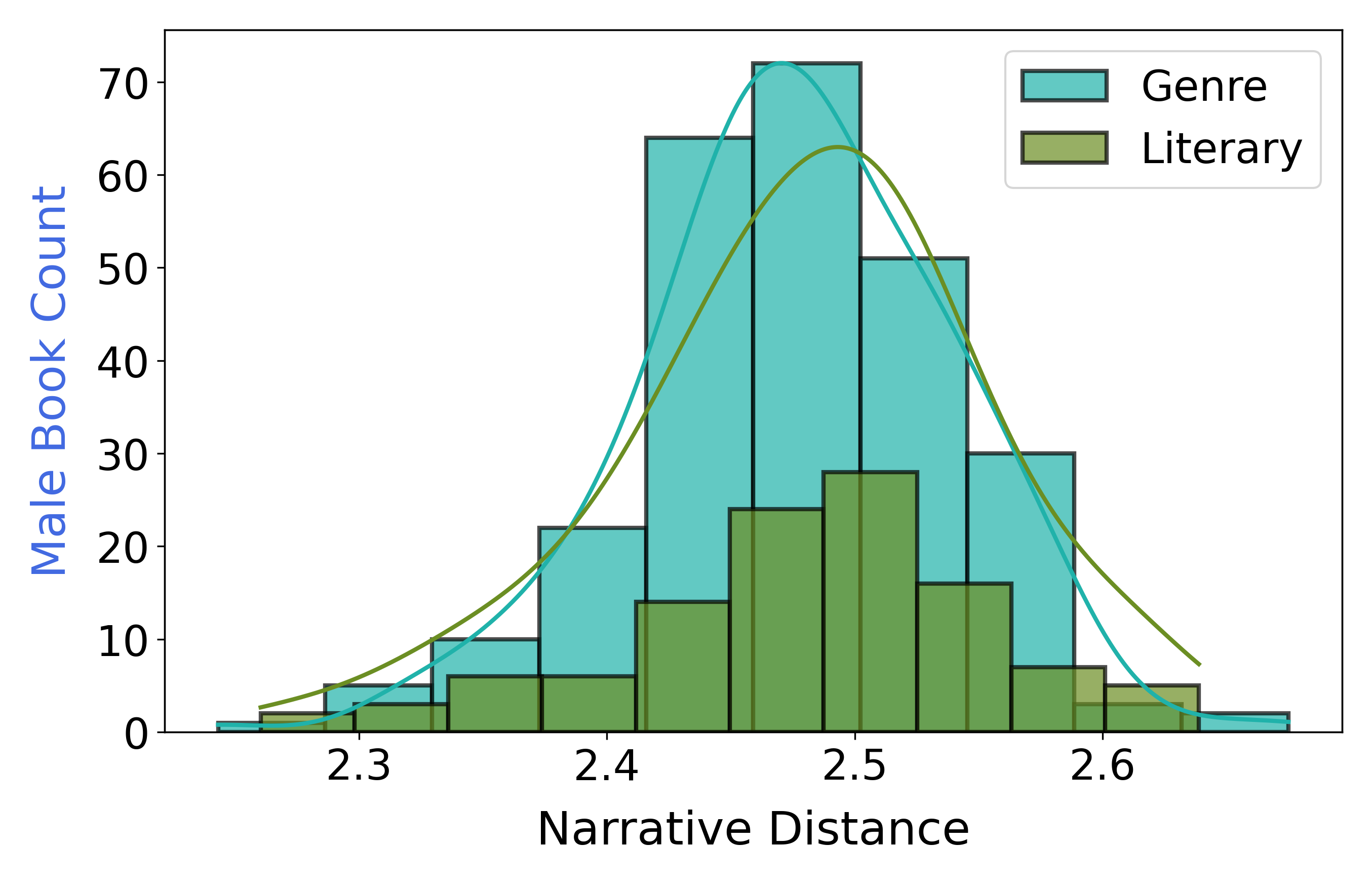}
  \centering
  \includegraphics[width=0.45\linewidth]{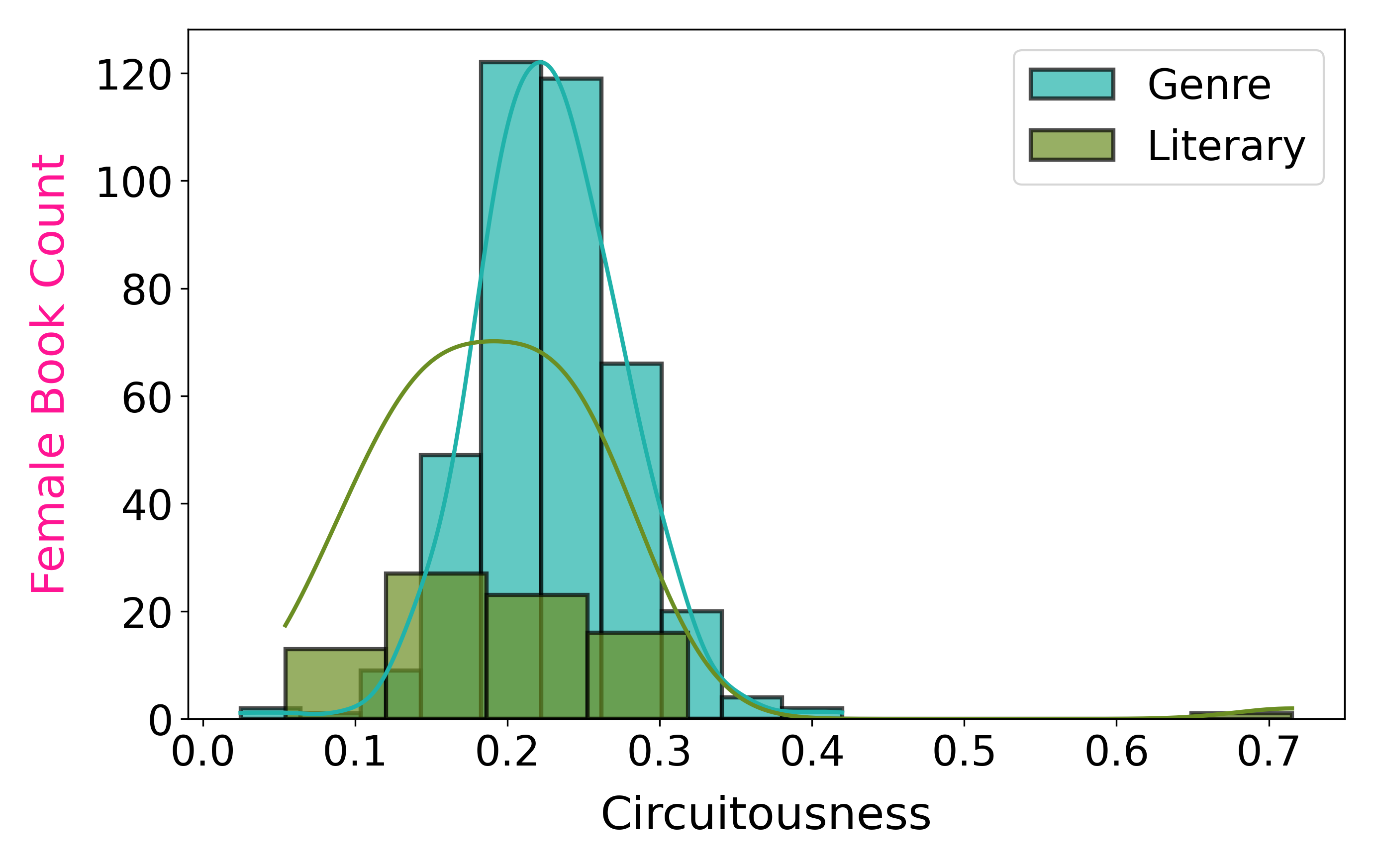}
  \includegraphics[width=0.45\linewidth]{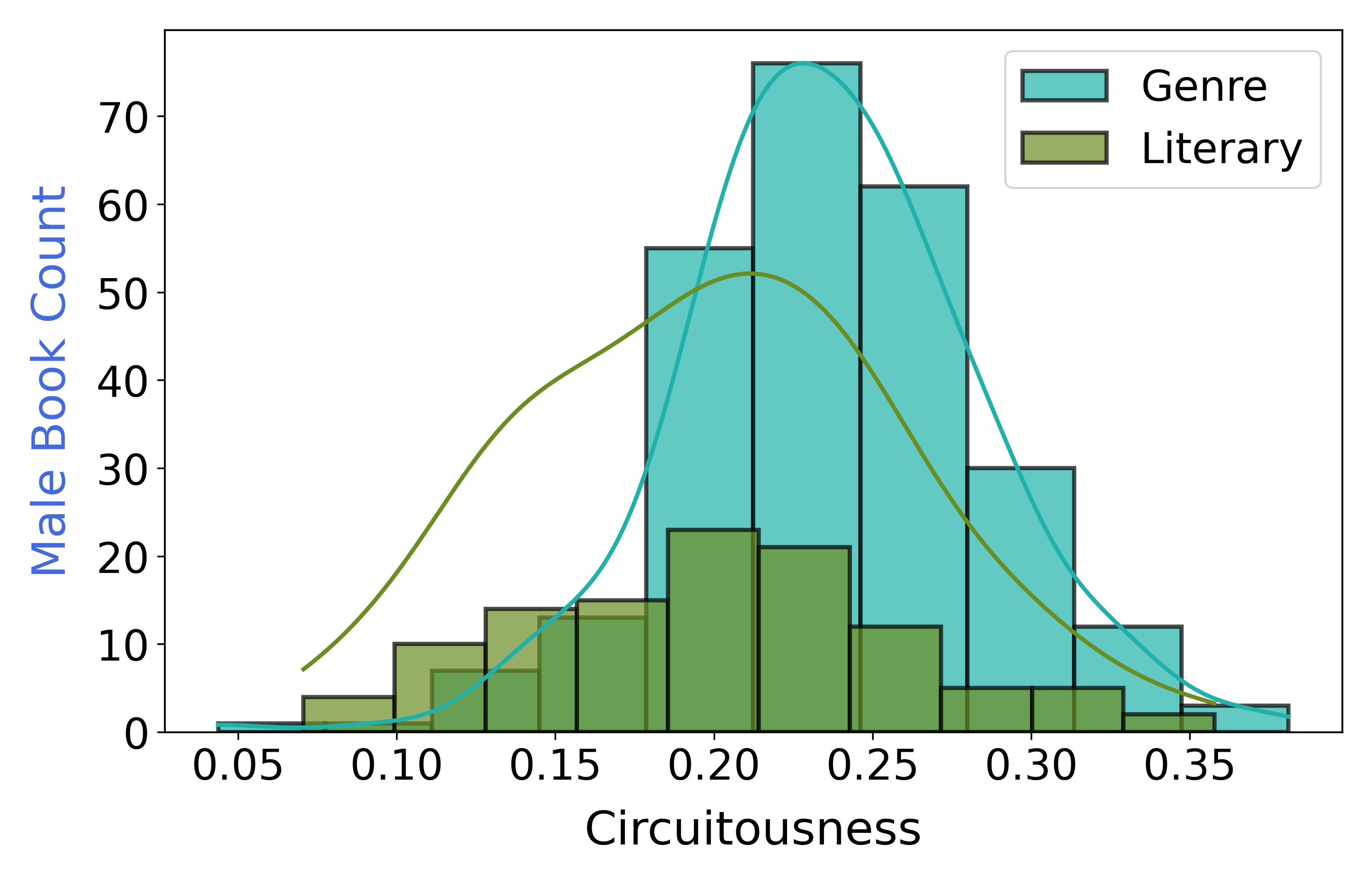}
  \centering
  \includegraphics[width=0.45\linewidth]{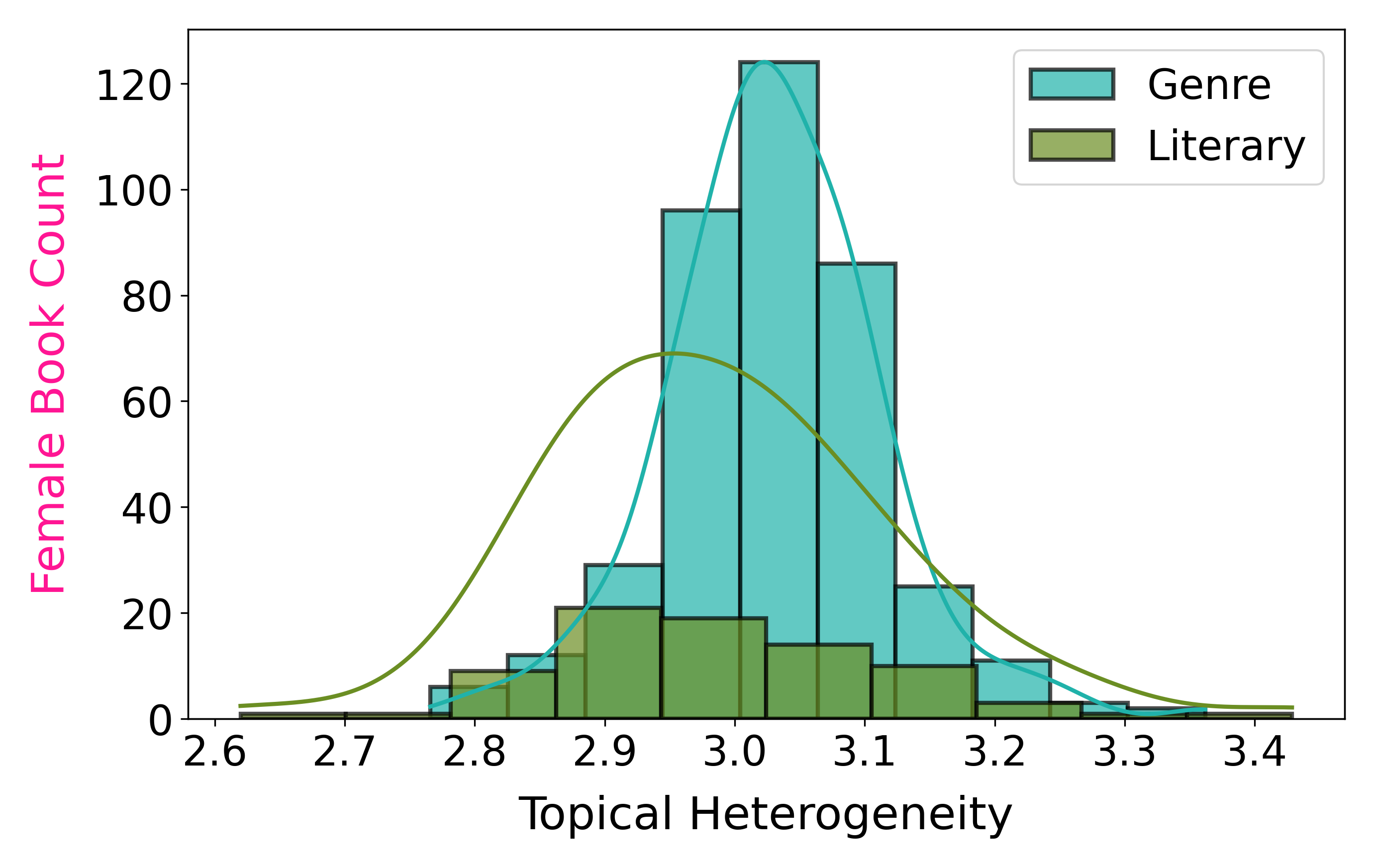}
  \includegraphics[width=0.45\linewidth]{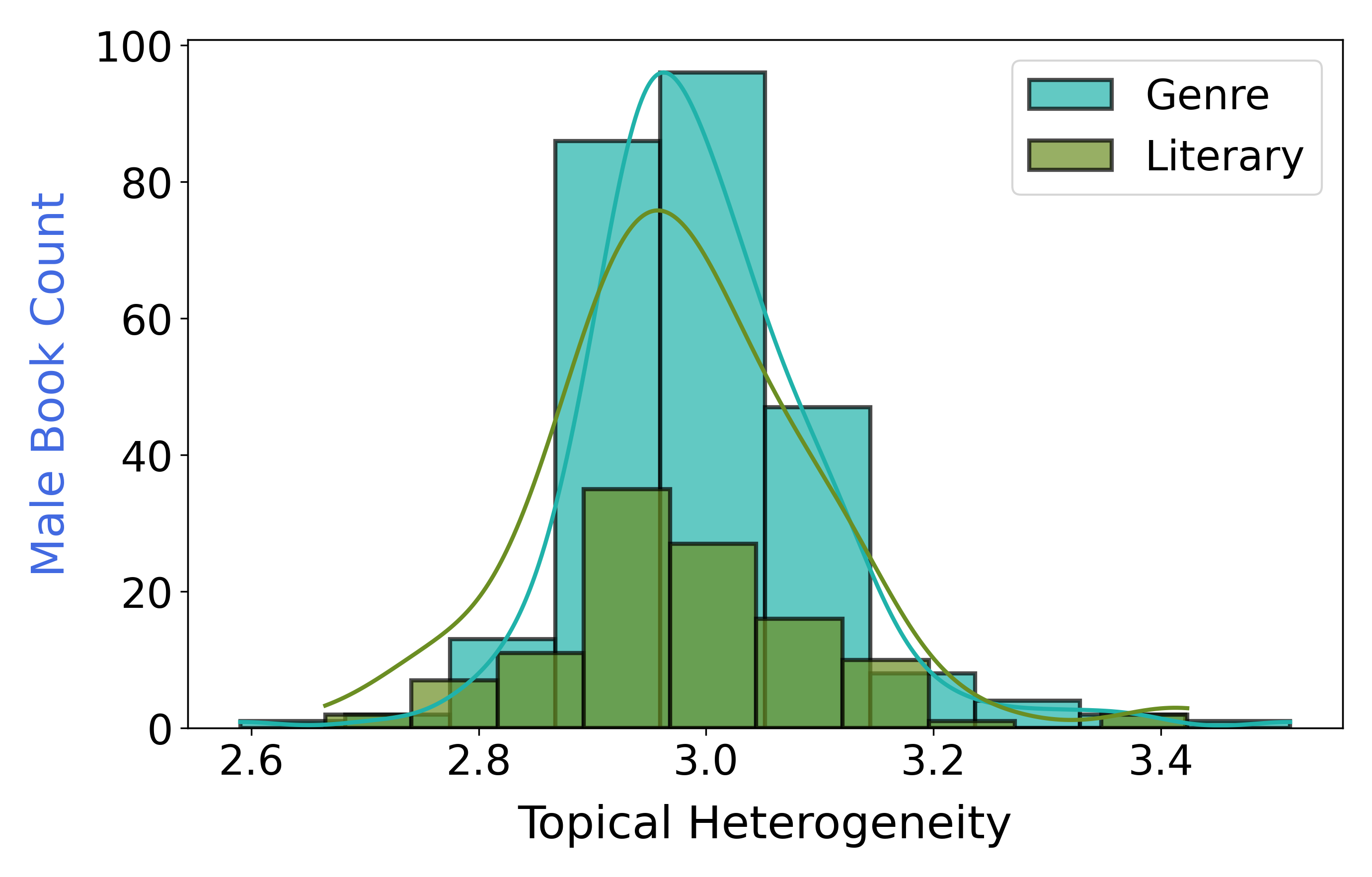}
  \centering
  \includegraphics[width=0.45\linewidth]{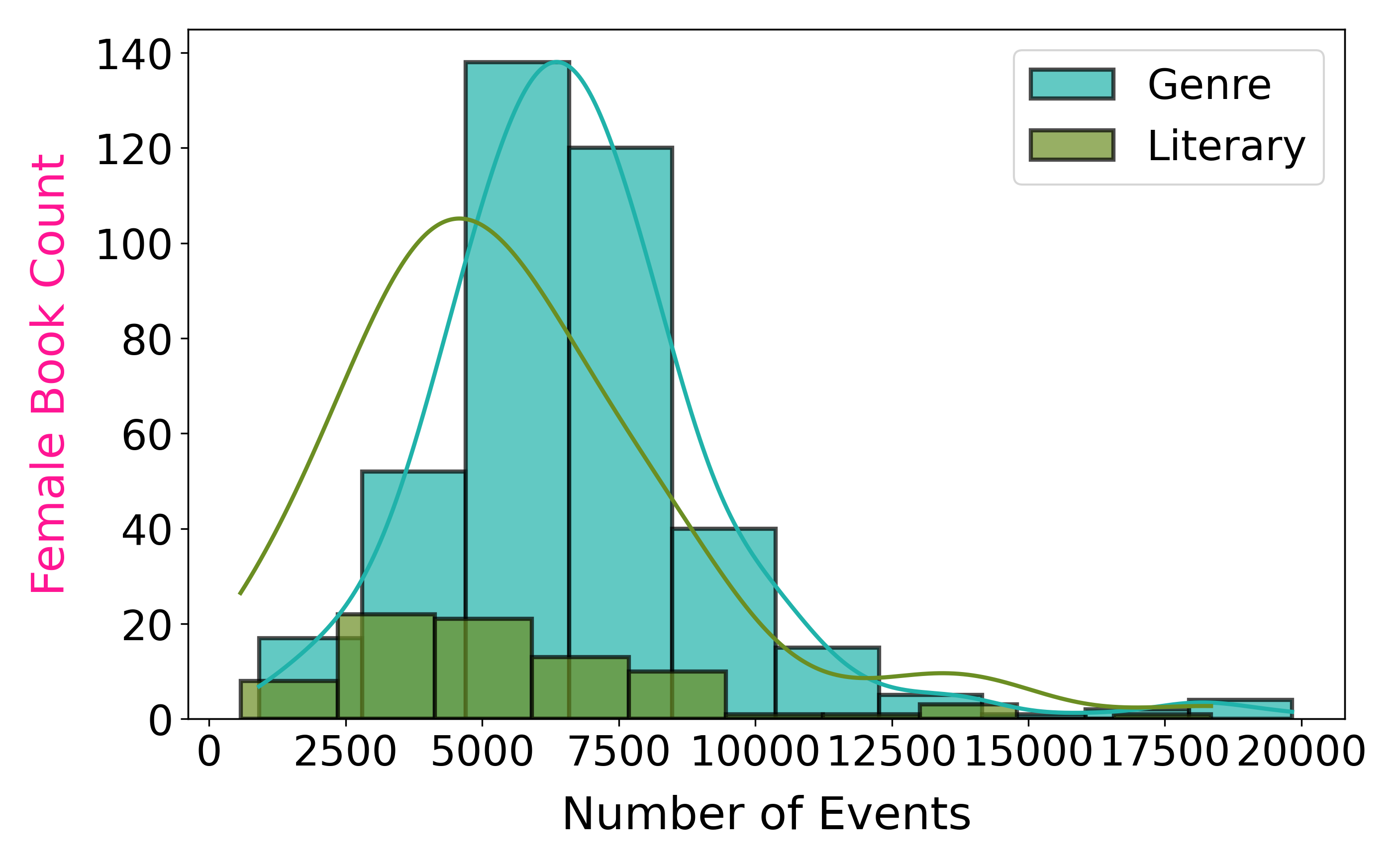}
  \includegraphics[width=0.45\linewidth]{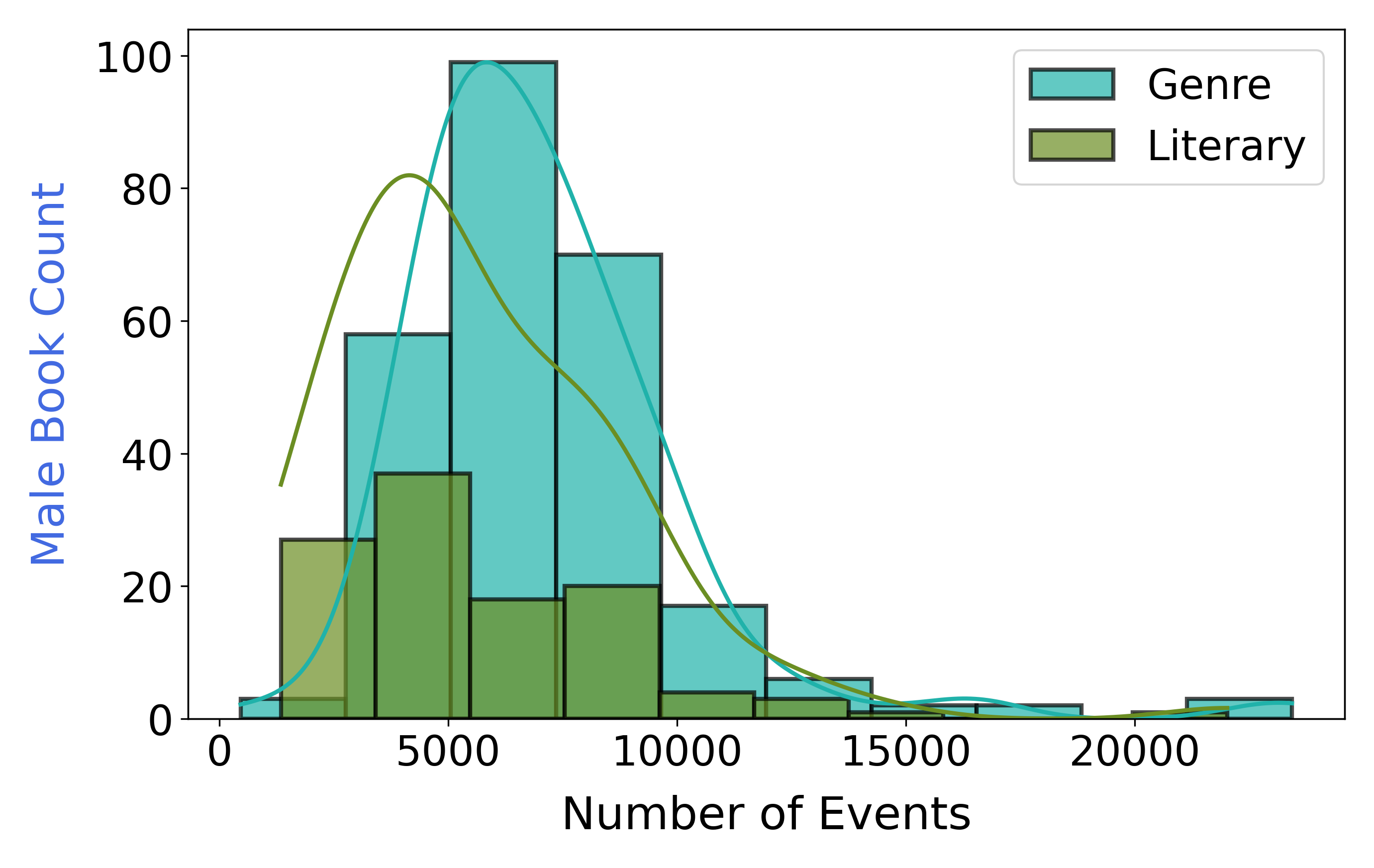}
    \caption{Narrative structure histograms for literary vs genre fiction by author gender}
  \label{fig:hist5}
\end{figure}

\begin{figure}[H]
  \centering
  \includegraphics[width=0.45\linewidth]{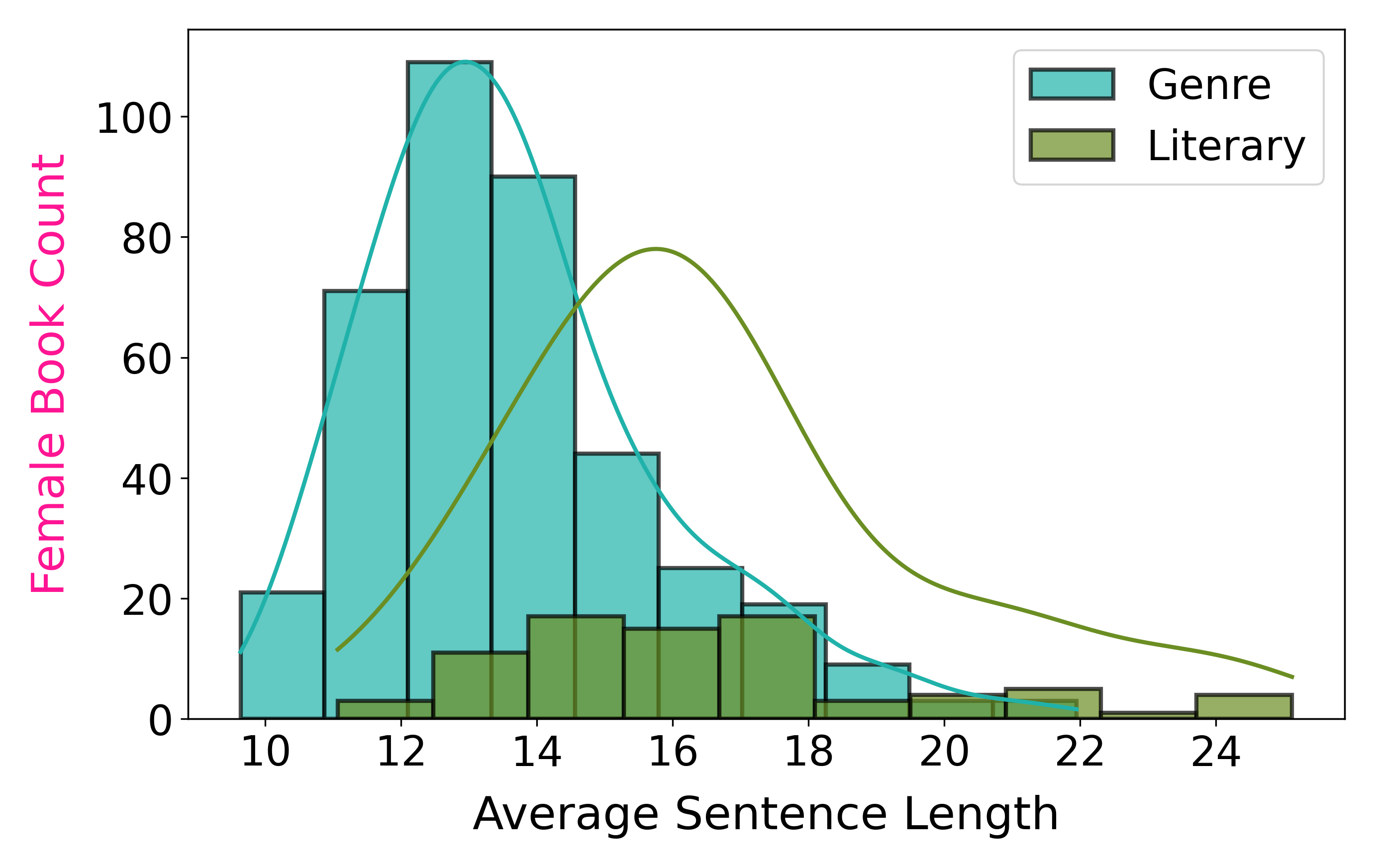}
  \includegraphics[width=0.45\linewidth]{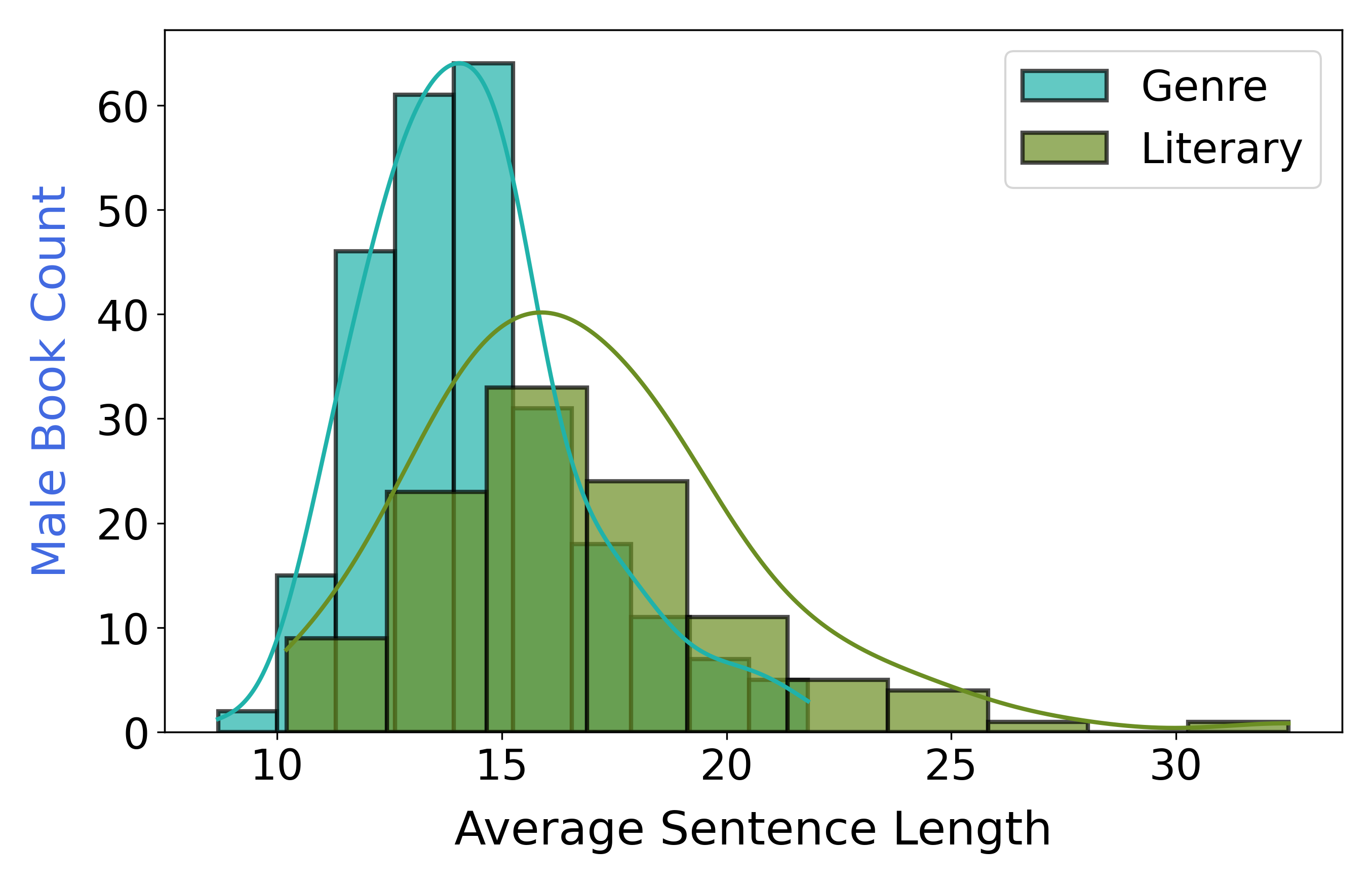}
  \centering
  \includegraphics[width=0.45\linewidth]{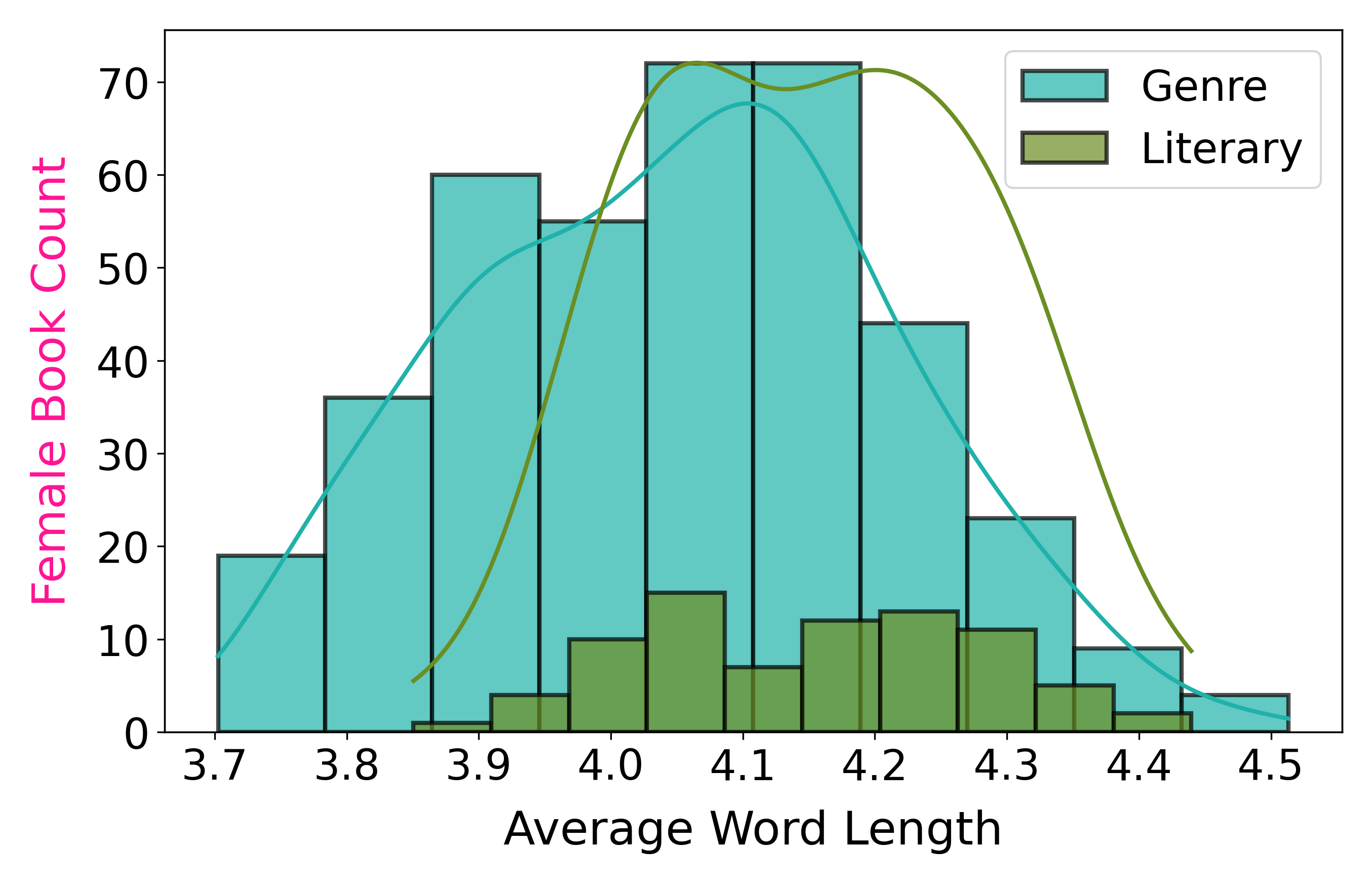}
  \includegraphics[width=0.45\linewidth]{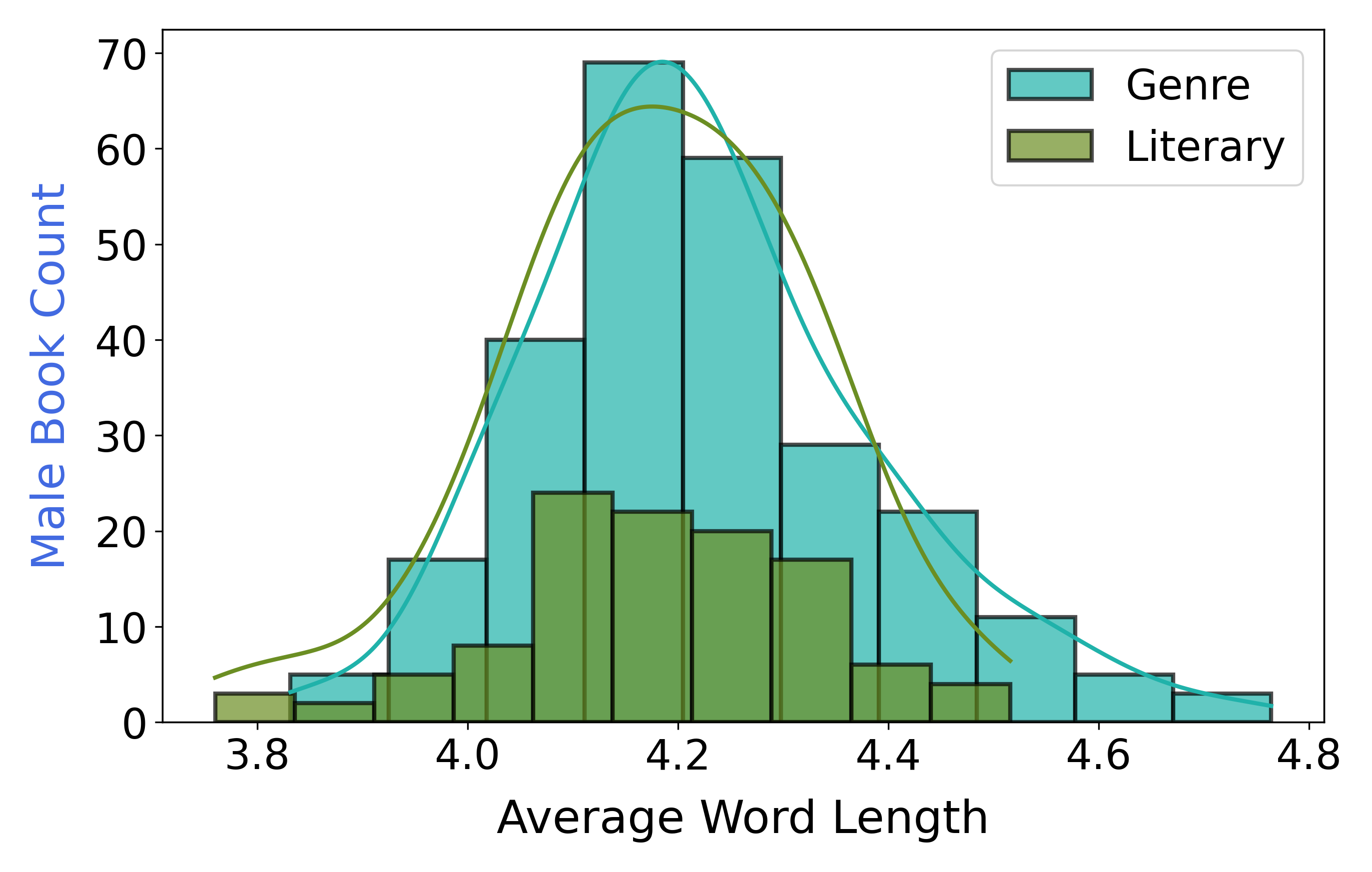}
  \centering
  \includegraphics[width=0.45\linewidth]{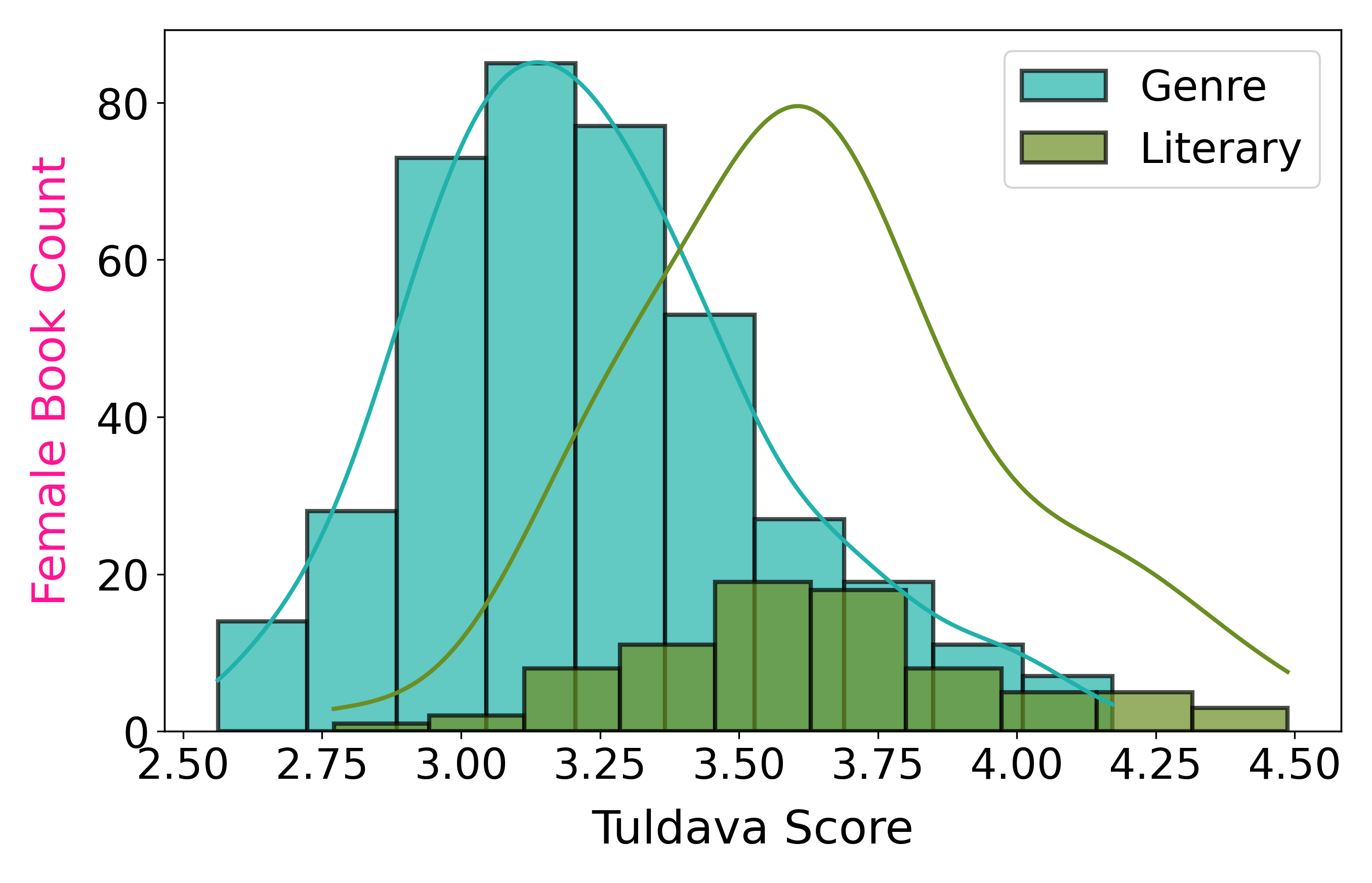}
  \includegraphics[width=0.45\linewidth]{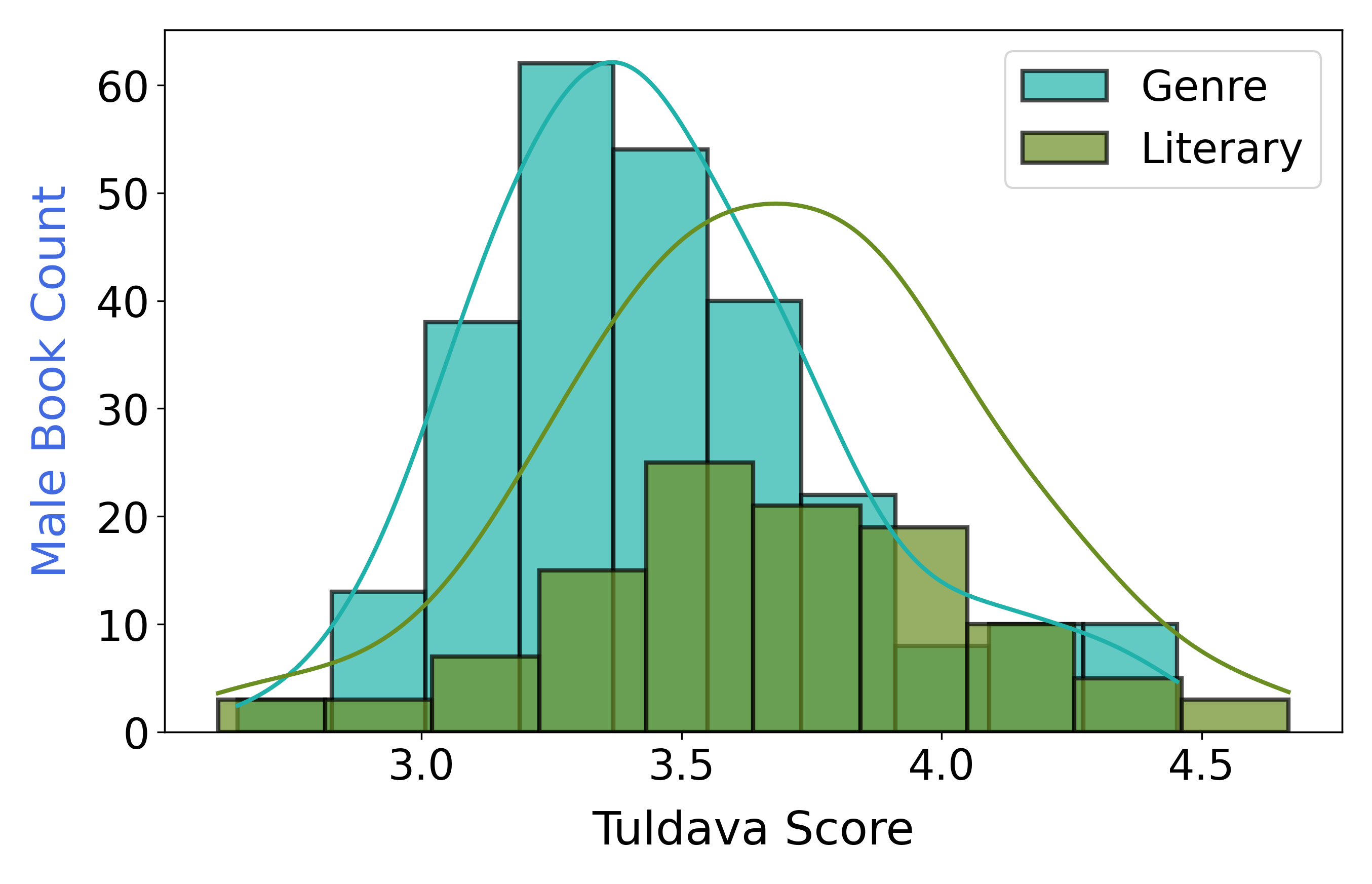}
  \centering
  \includegraphics[width=0.45\linewidth]{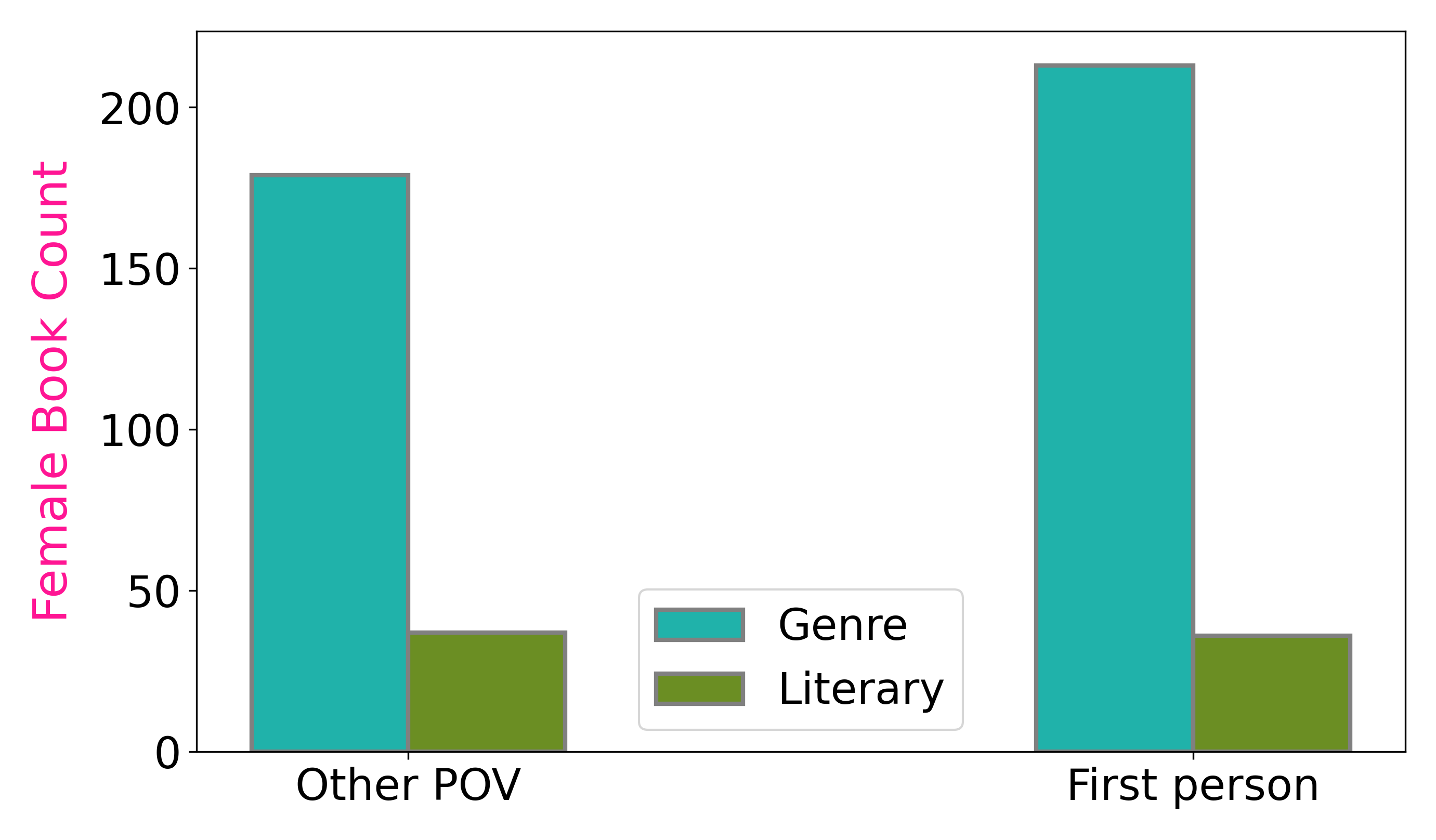}
  \includegraphics[width=0.45\linewidth]{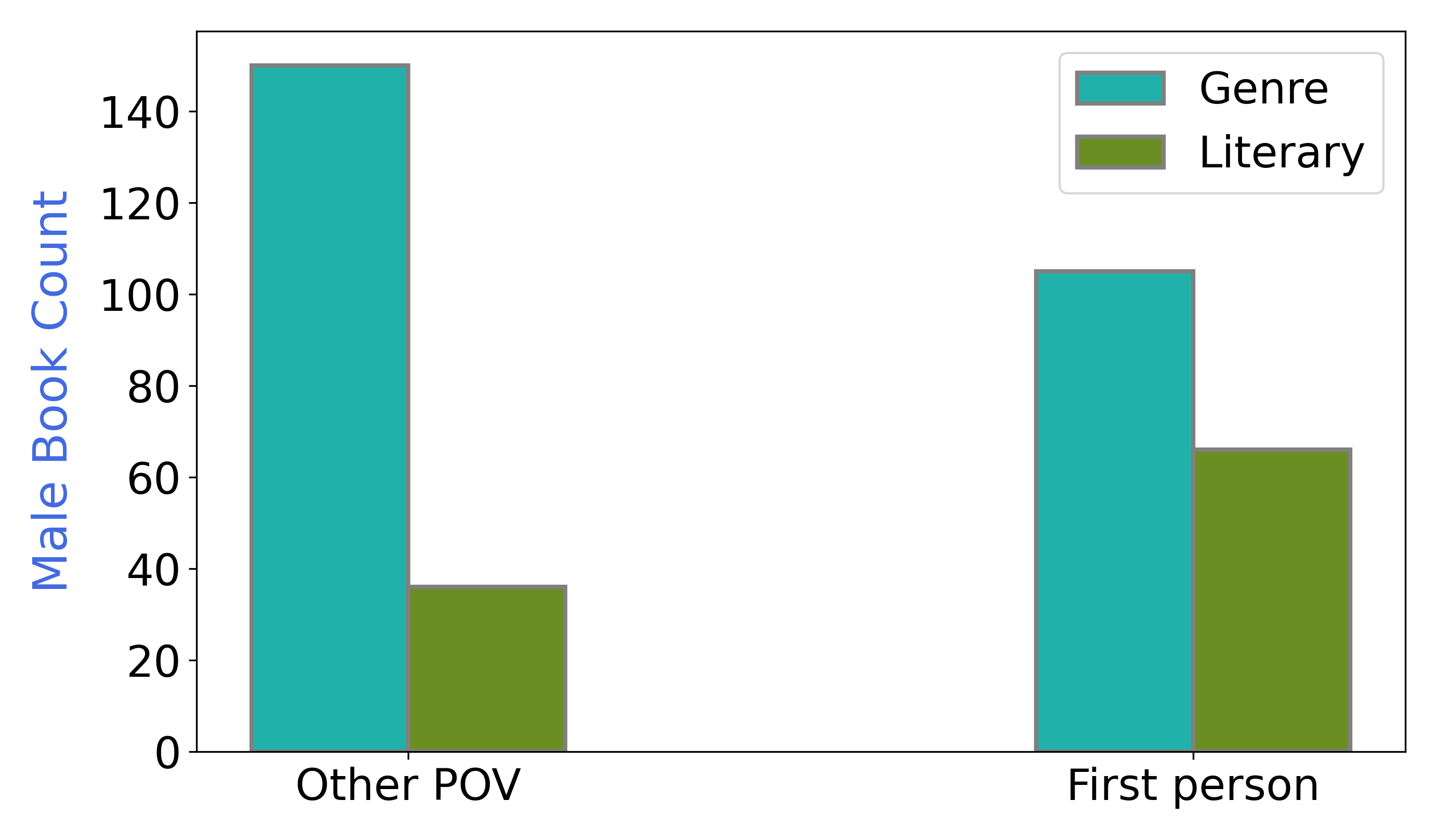}
    \caption{Language histograms for literary vs genre fiction by author gender}
  \label{fig:hist6}
\end{figure}

\begin{figure}[H]
  \centering
  \includegraphics[width=0.45\linewidth]{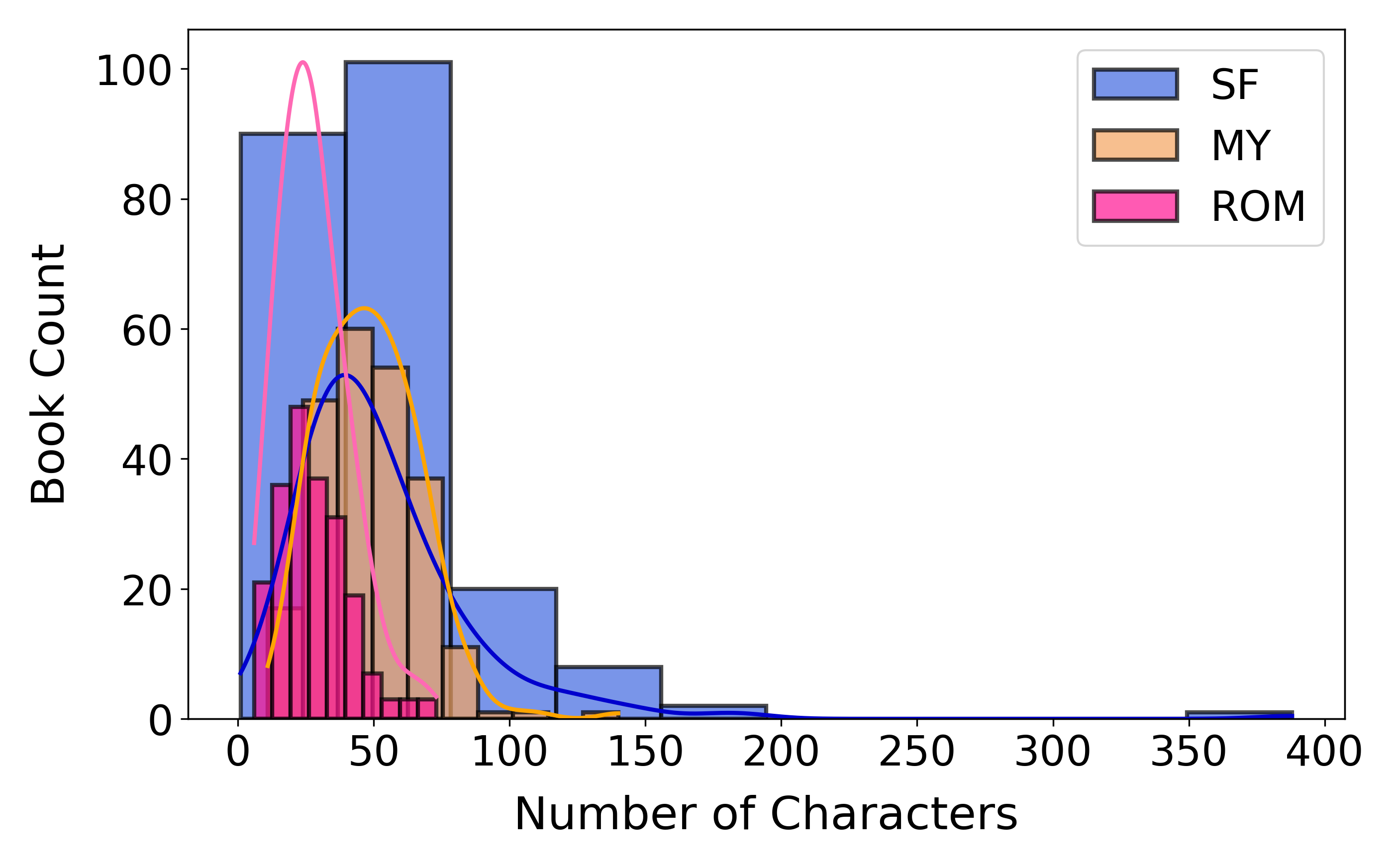}
  \includegraphics[width=0.45\linewidth]{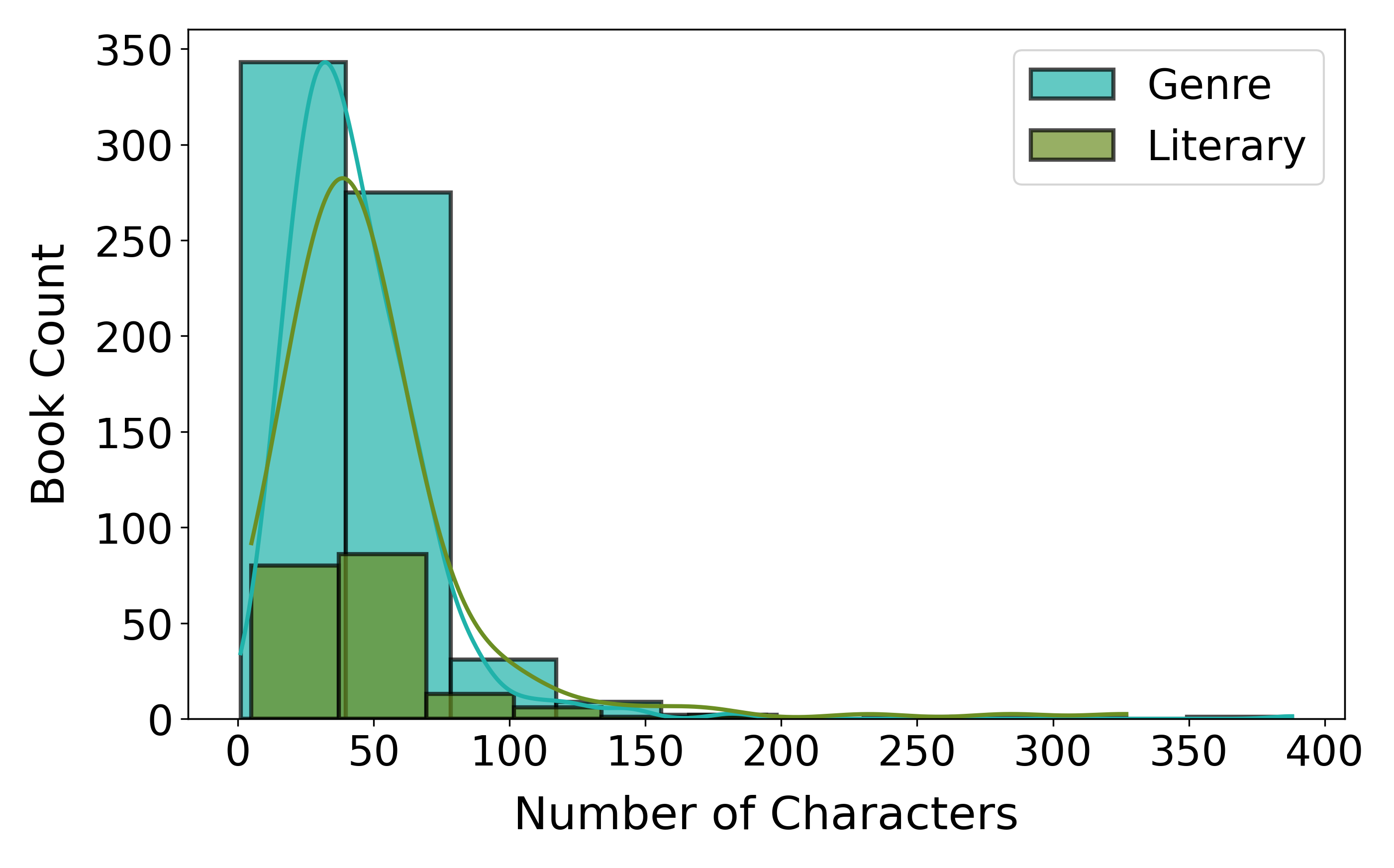}
\centering
  \includegraphics[width=0.45\linewidth]{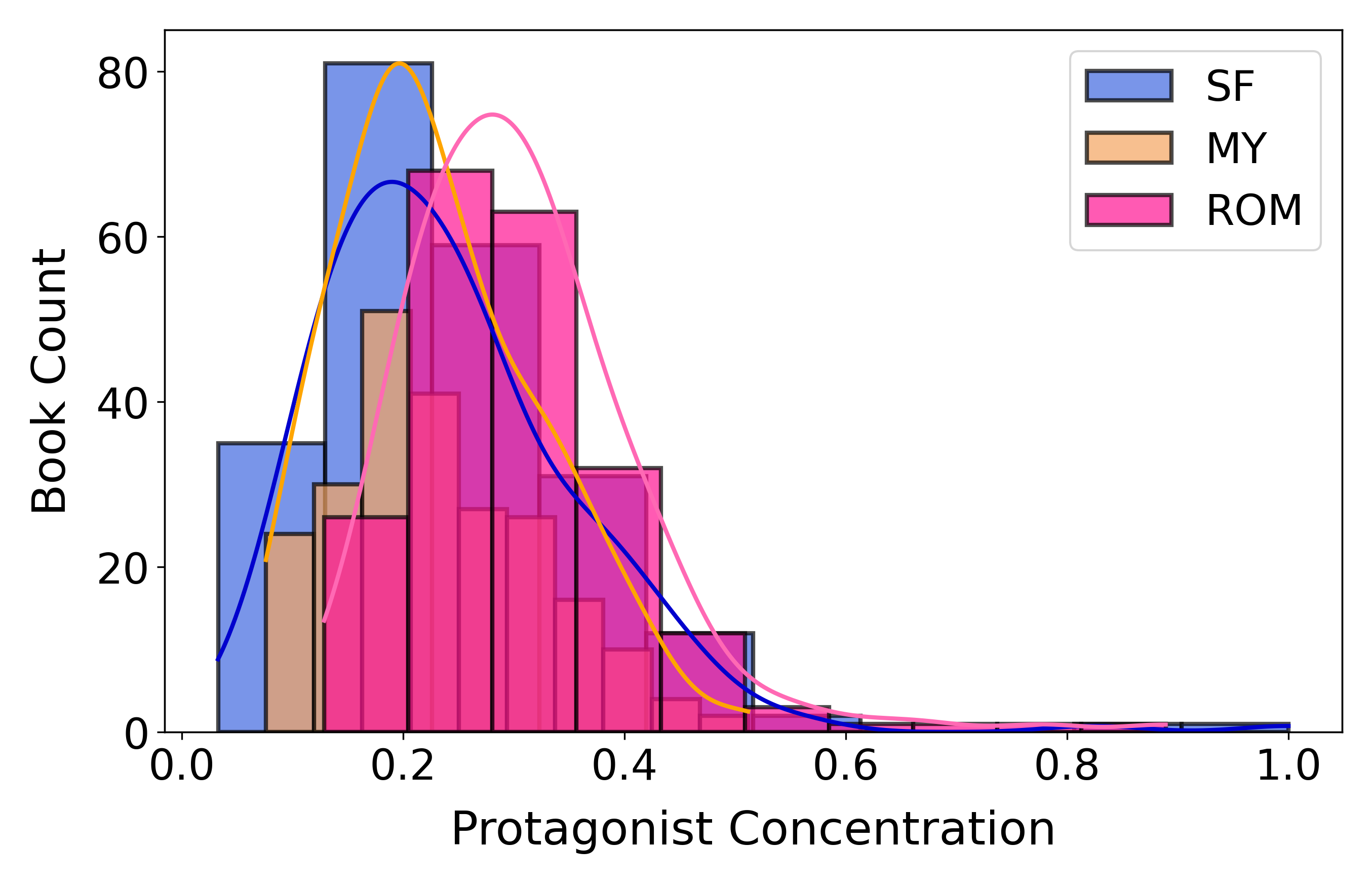}
  \includegraphics[width=0.45\linewidth]{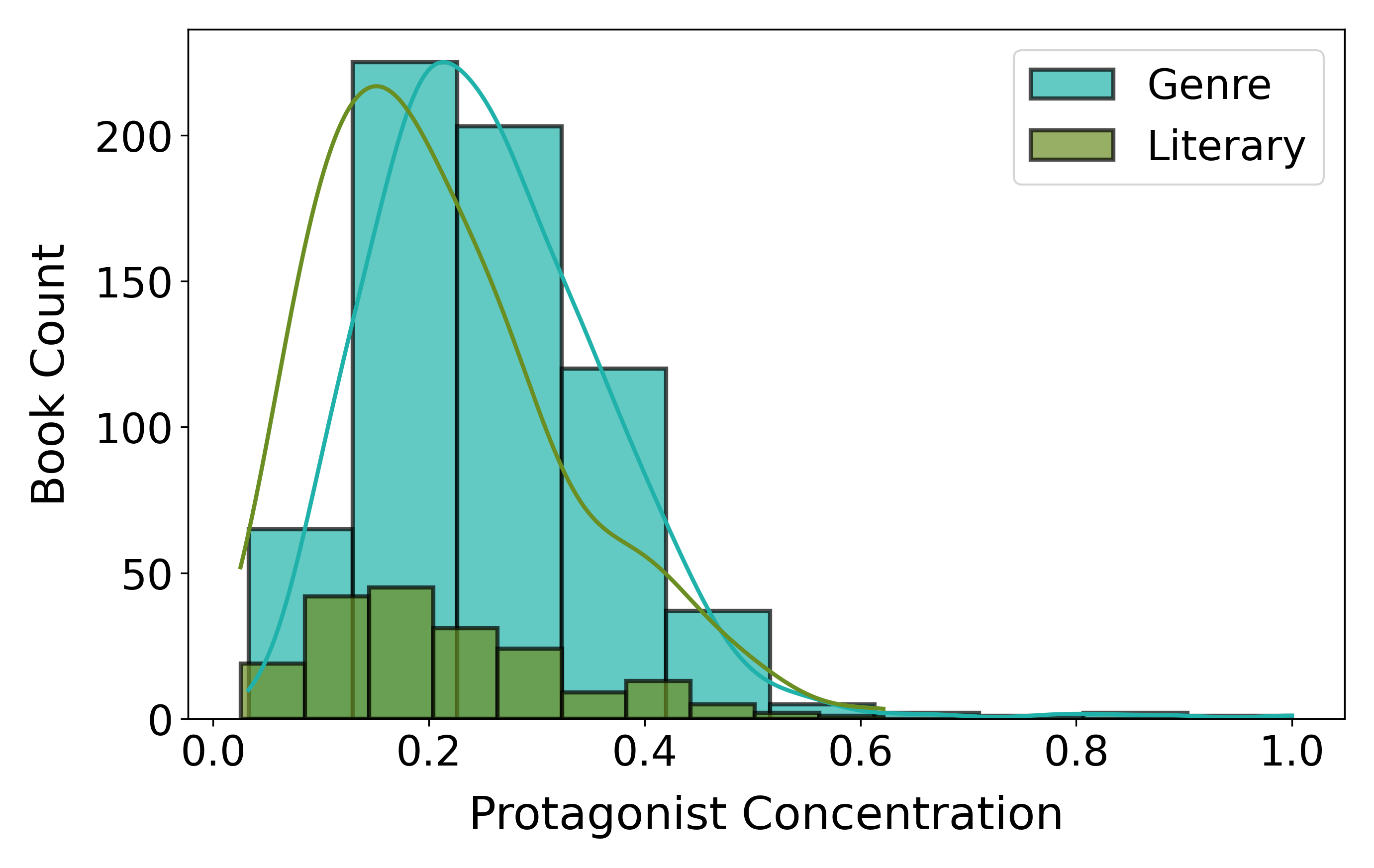}
    \caption{Character histograms}
  \label{fig:charhist1}
\end{figure}

\begin{figure}[H]
  \centering
  \includegraphics[width=0.45\linewidth]{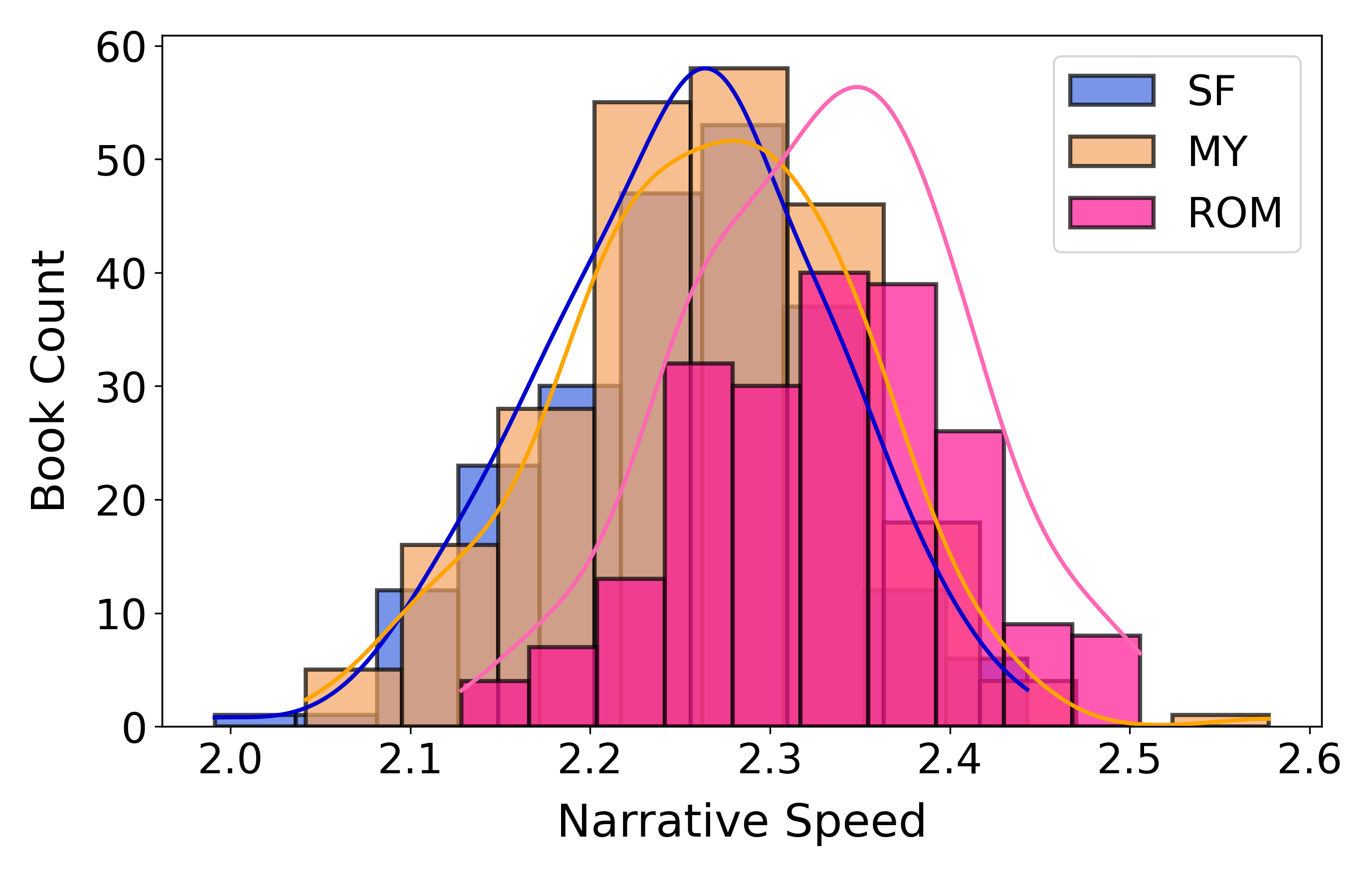}
  \includegraphics[width=0.45\linewidth]{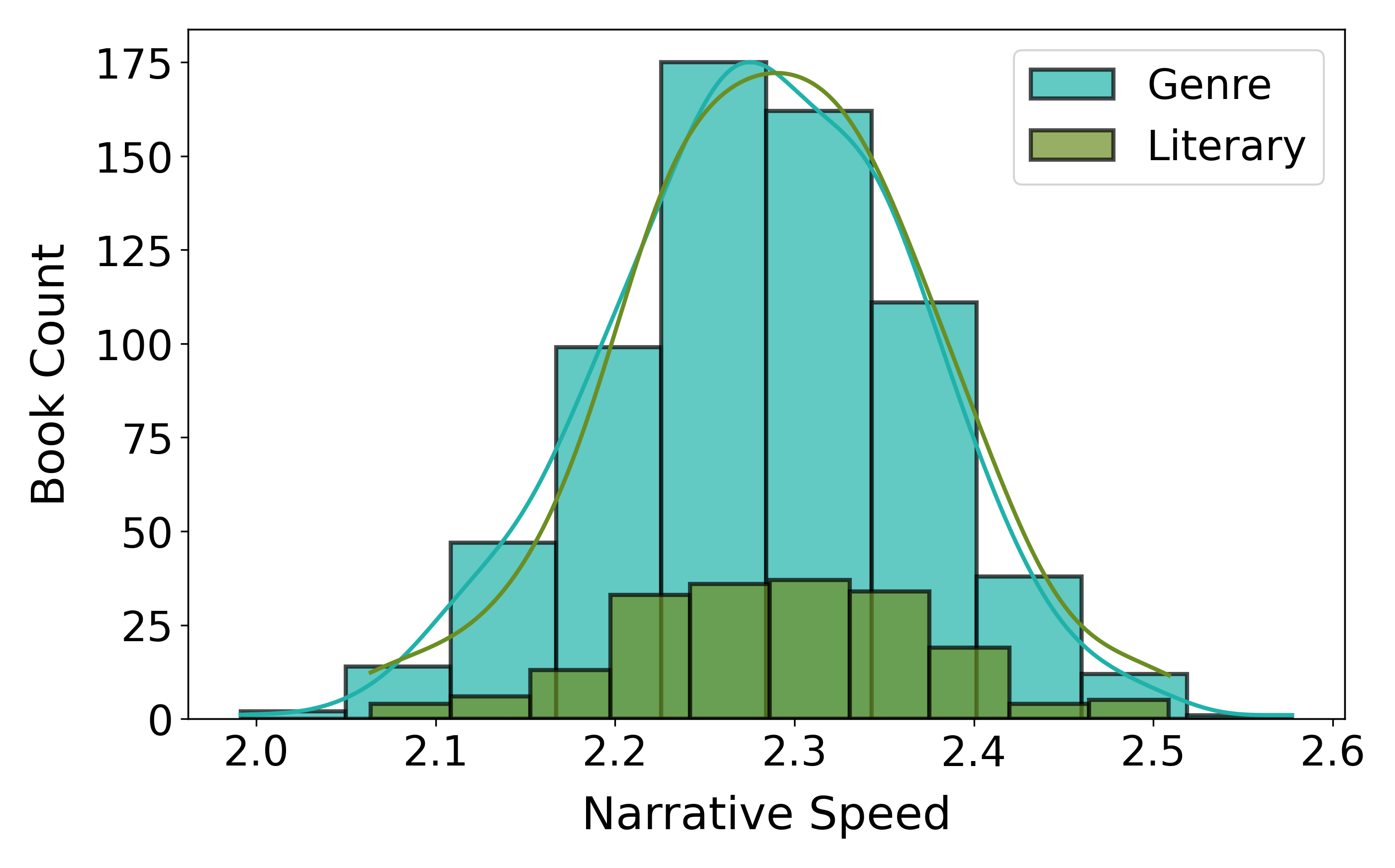}
  \centering
  \includegraphics[width=0.45\linewidth]{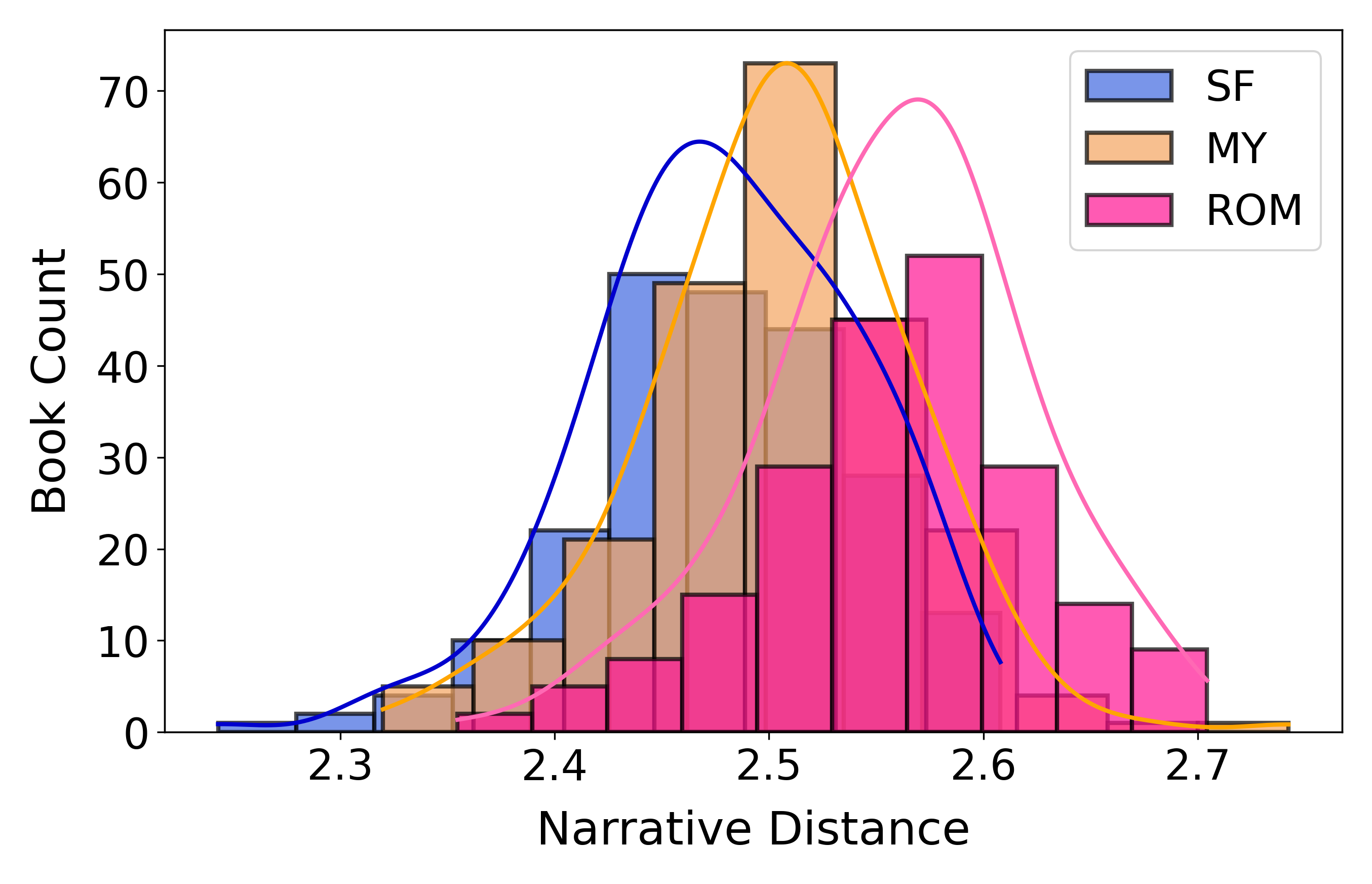}
  \includegraphics[width=0.45\linewidth]{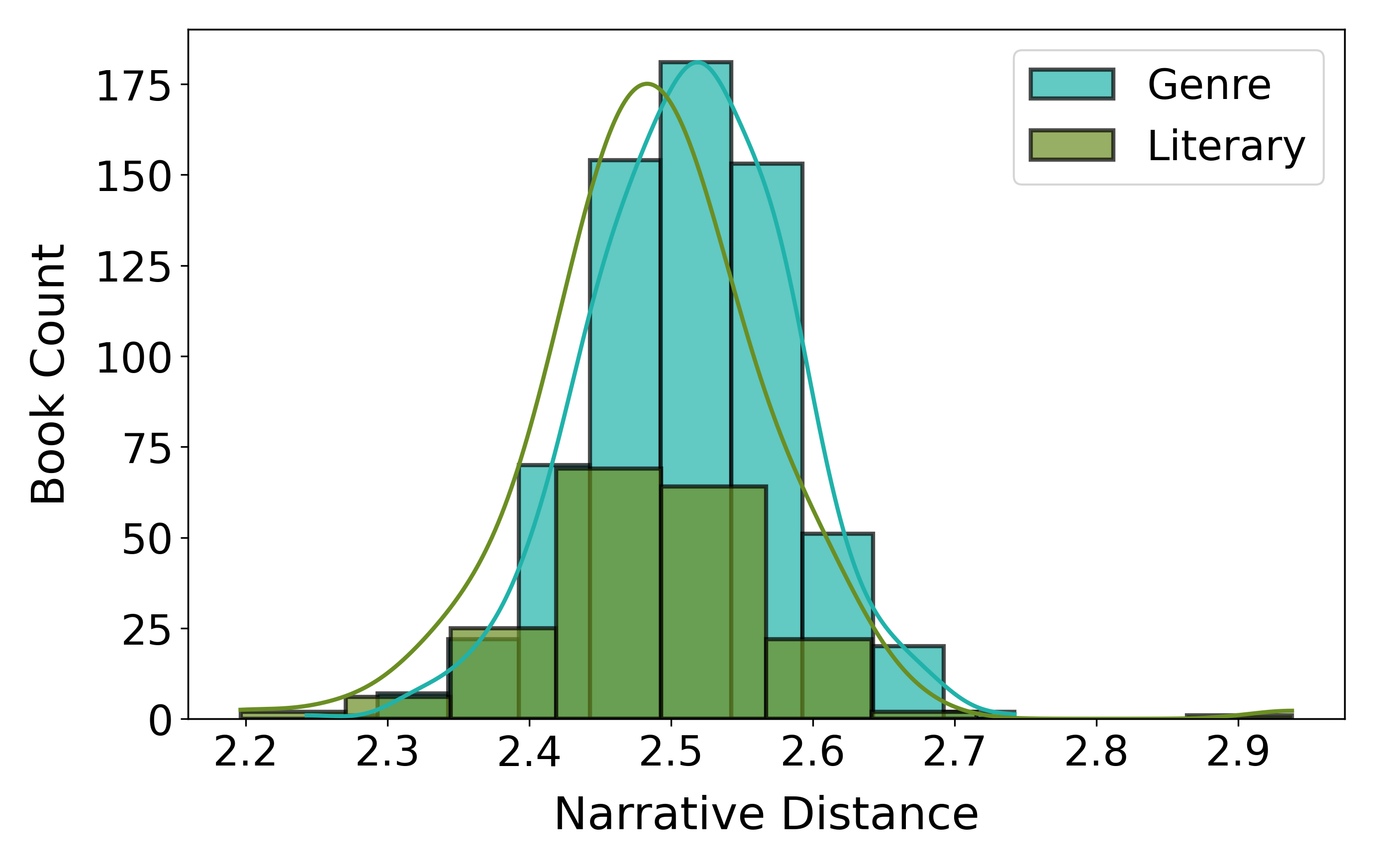}
  \centering
  \includegraphics[width=0.45\linewidth]{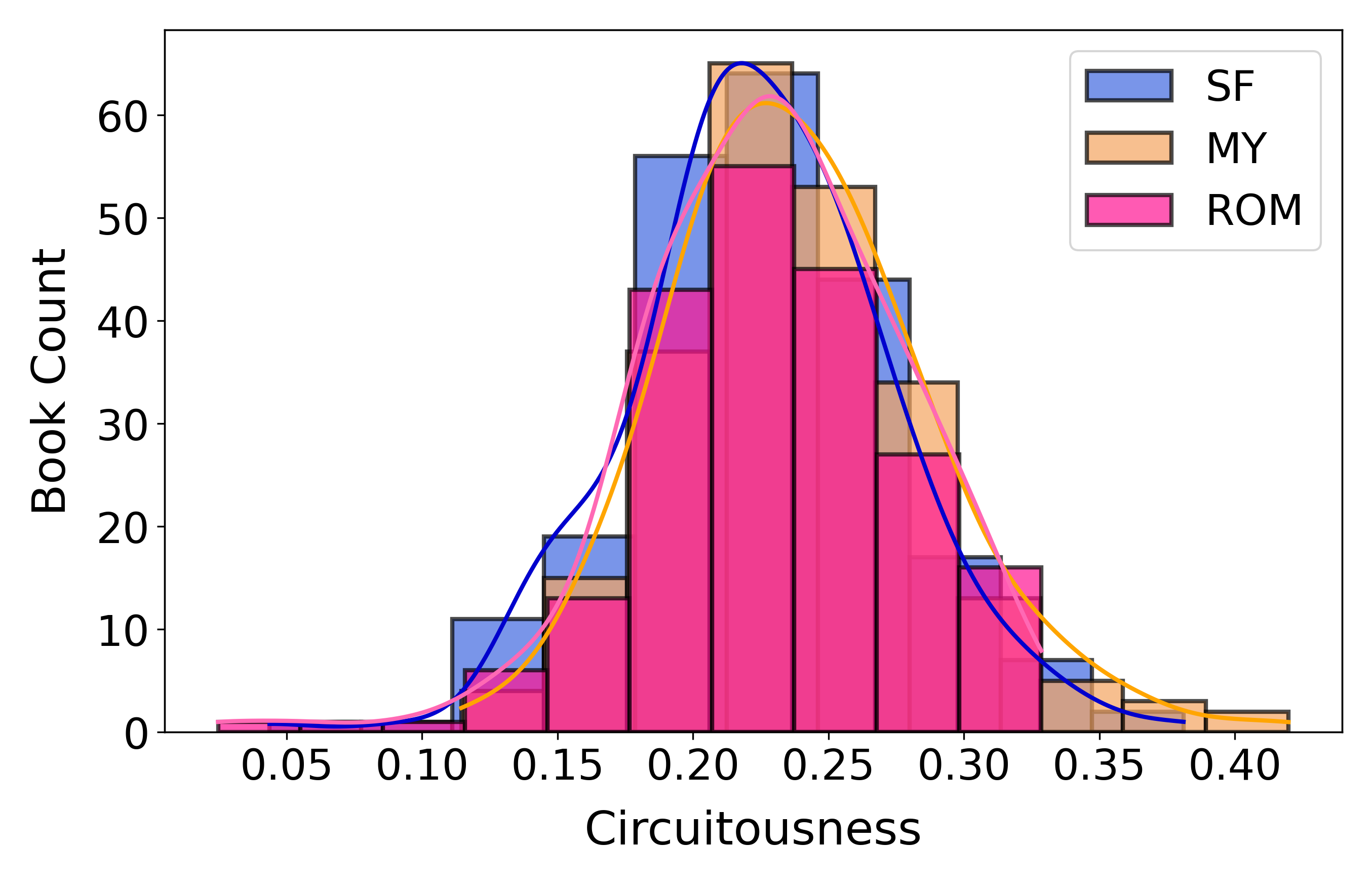}
  \includegraphics[width=0.45\linewidth]{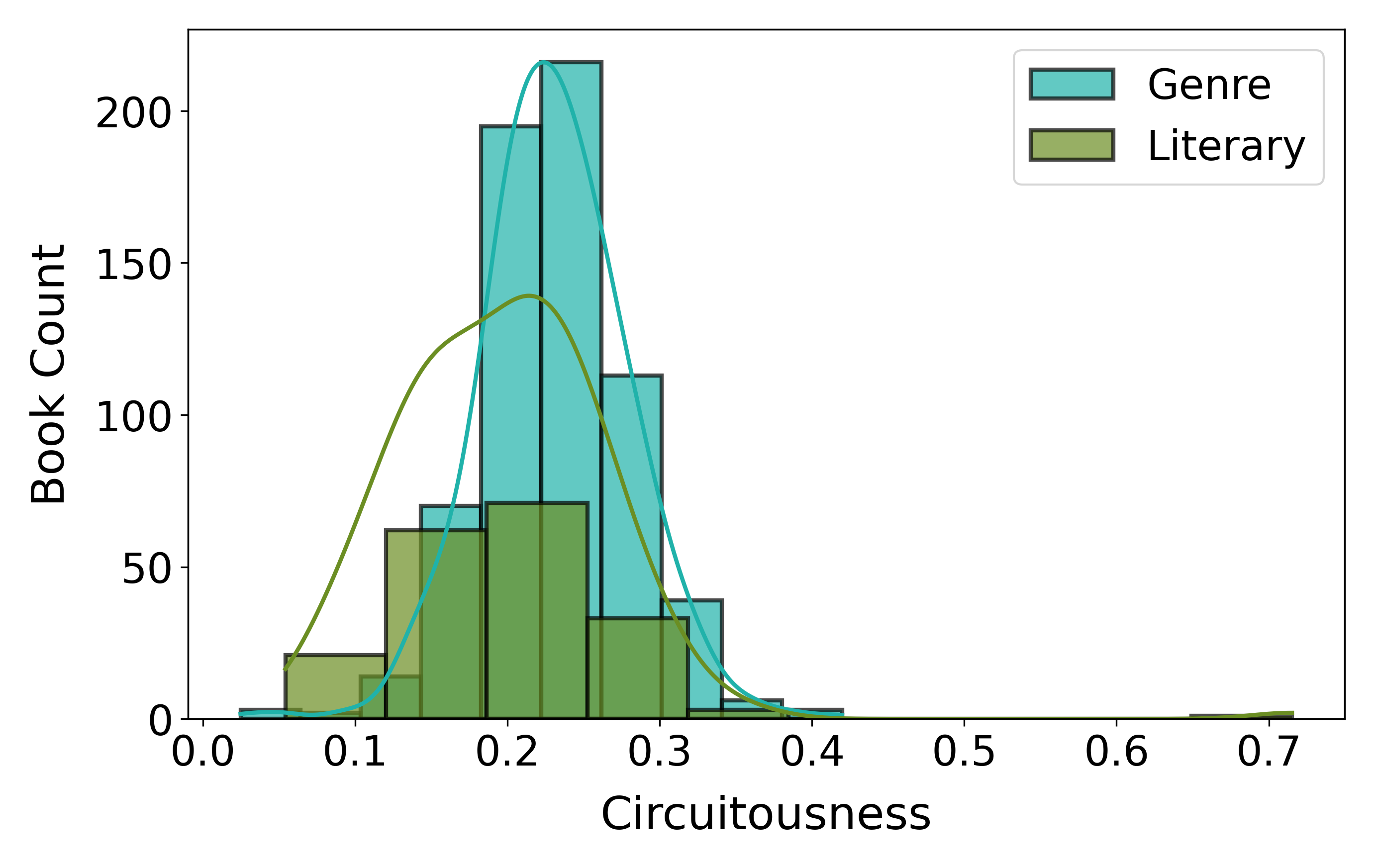}
  \centering
  \includegraphics[width=0.45\linewidth]{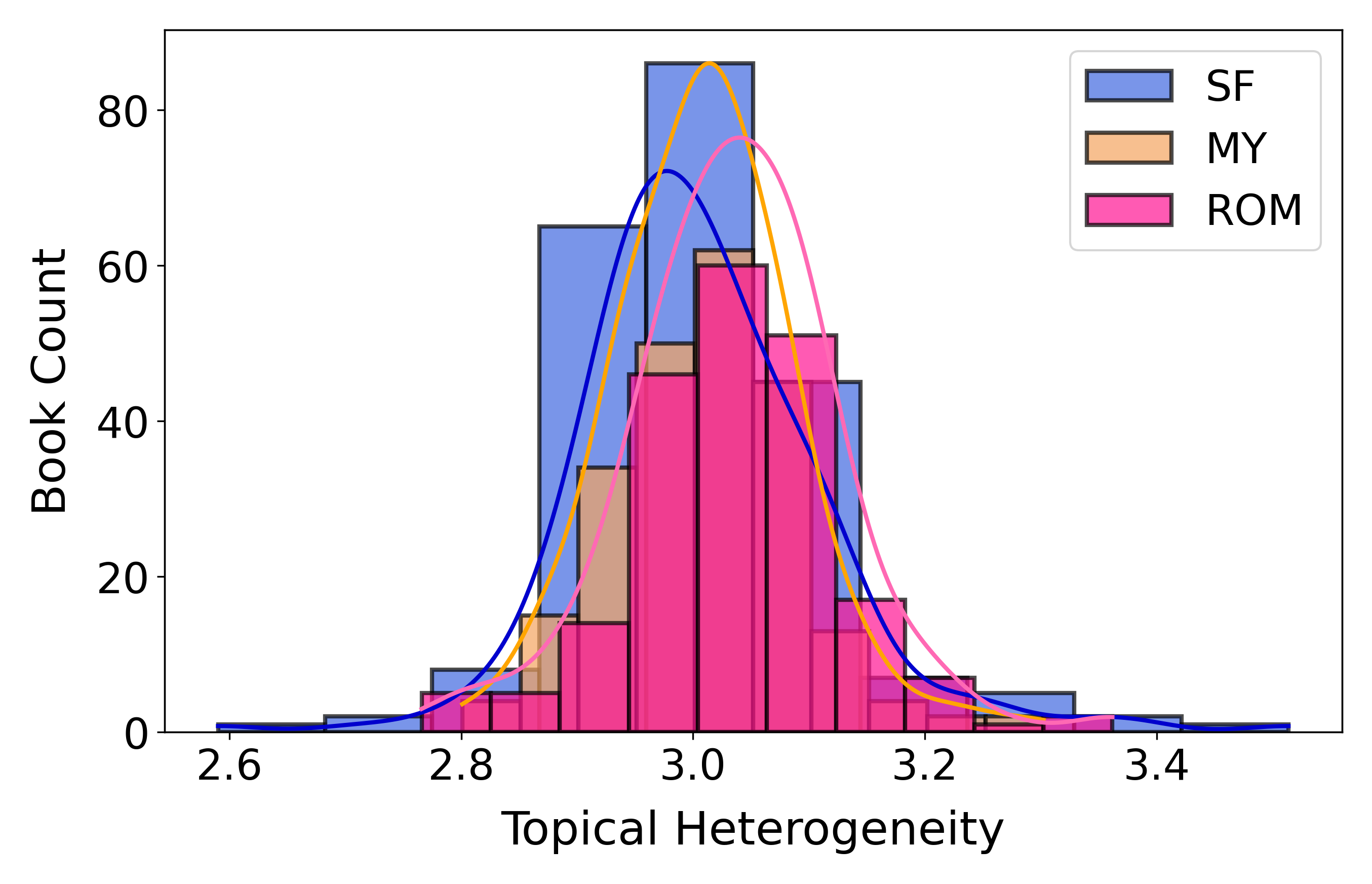}
  \includegraphics[width=0.45\linewidth]{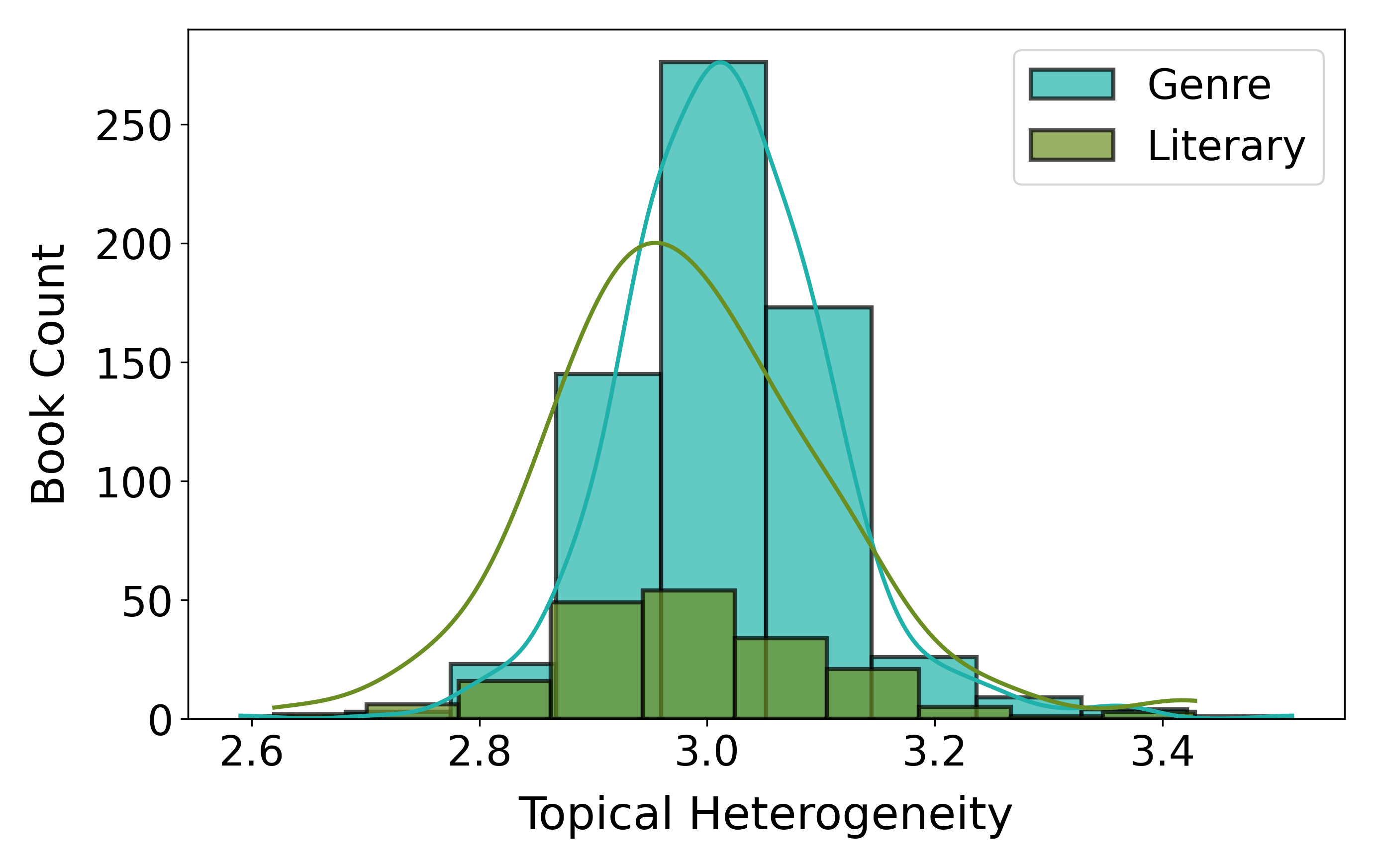}
  \centering
  \includegraphics[width=0.45\linewidth]{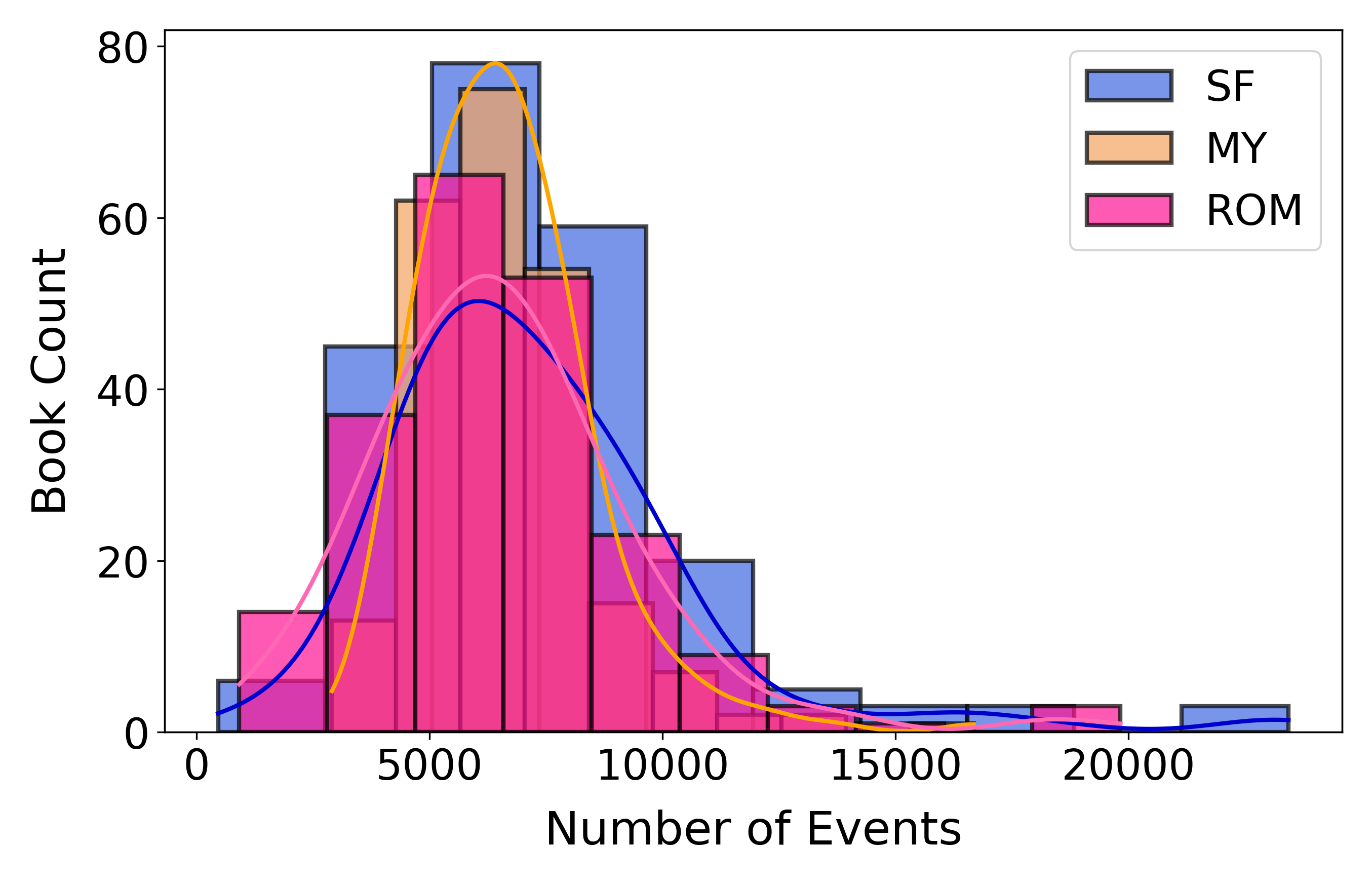}
  \includegraphics[width=0.45\linewidth]{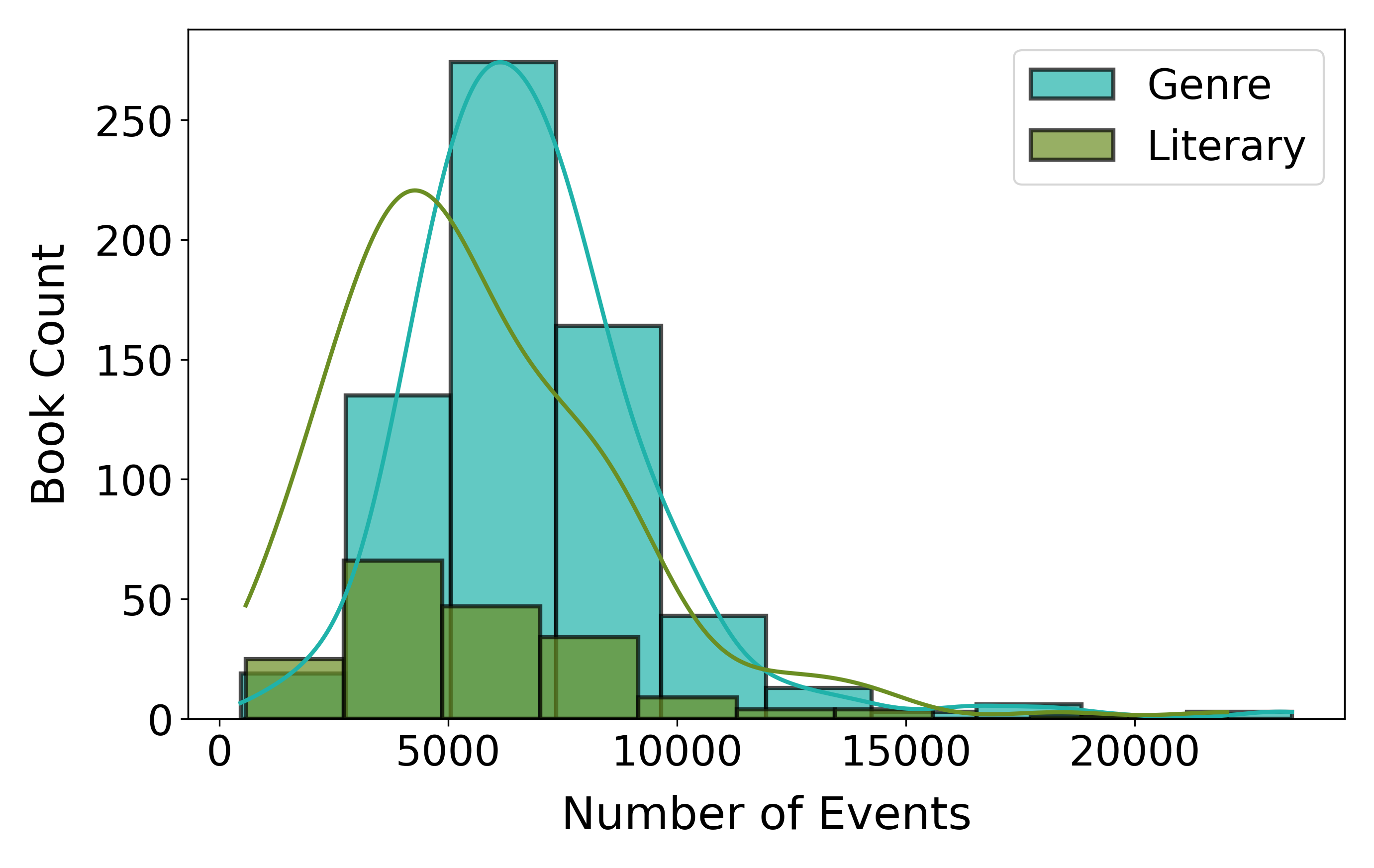}
    \caption{Narrative structure histograms}
  \label{fig:narrhist2}
\end{figure}

\begin{figure}[H]
  \centering
  \includegraphics[width=0.45\linewidth]{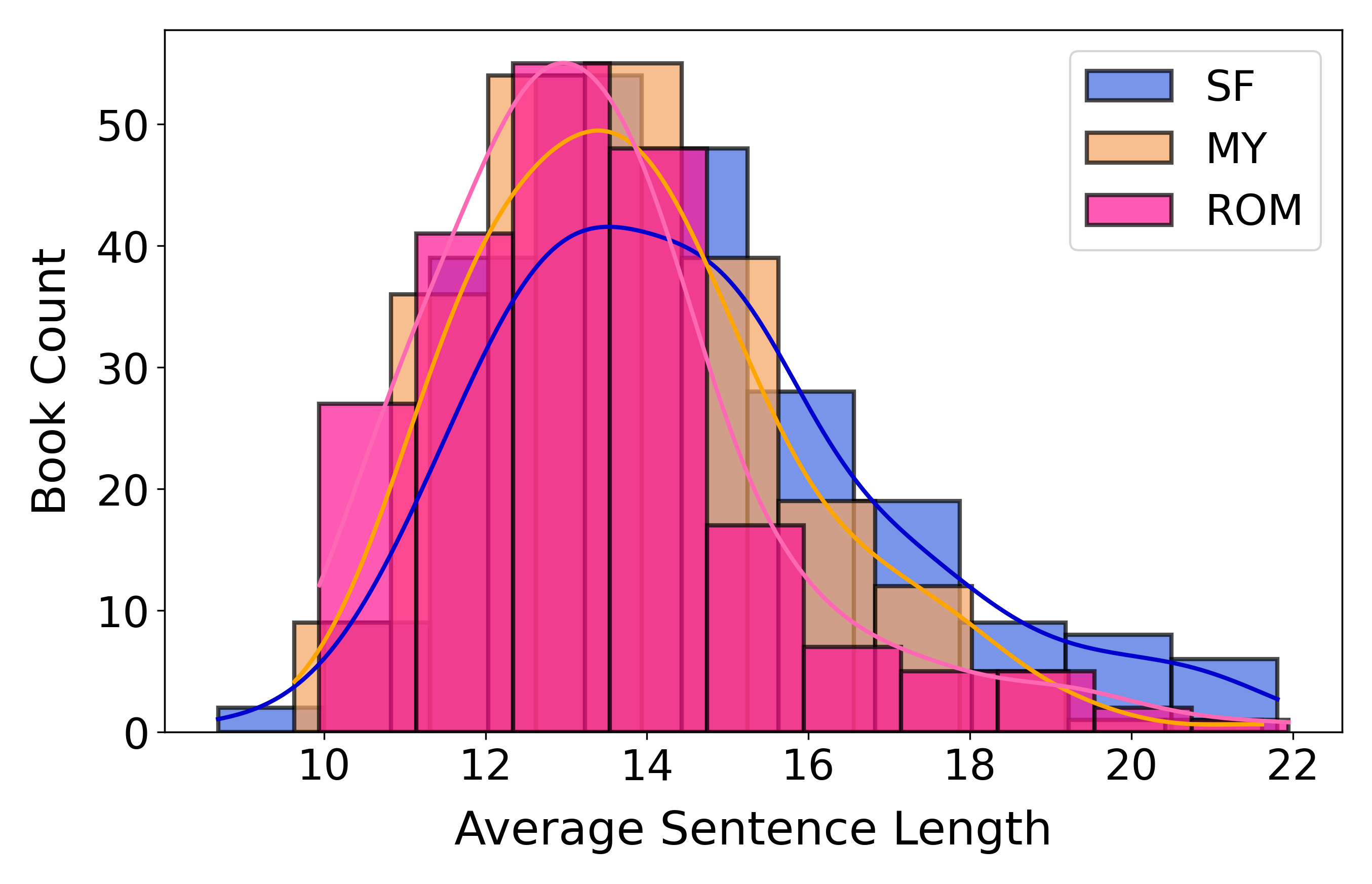}
  \includegraphics[width=0.45\linewidth]{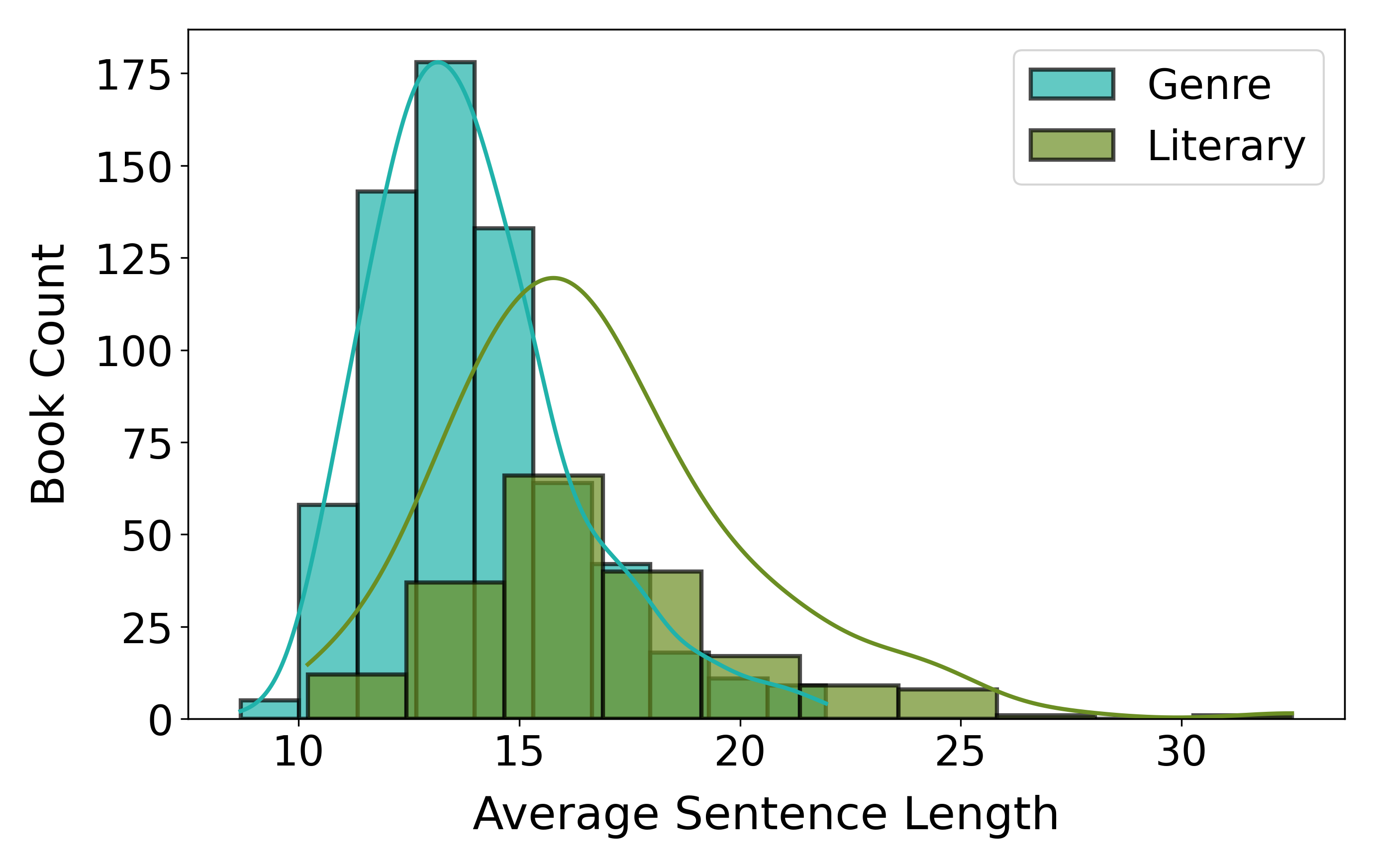}
  \centering
  \includegraphics[width=0.45\linewidth]{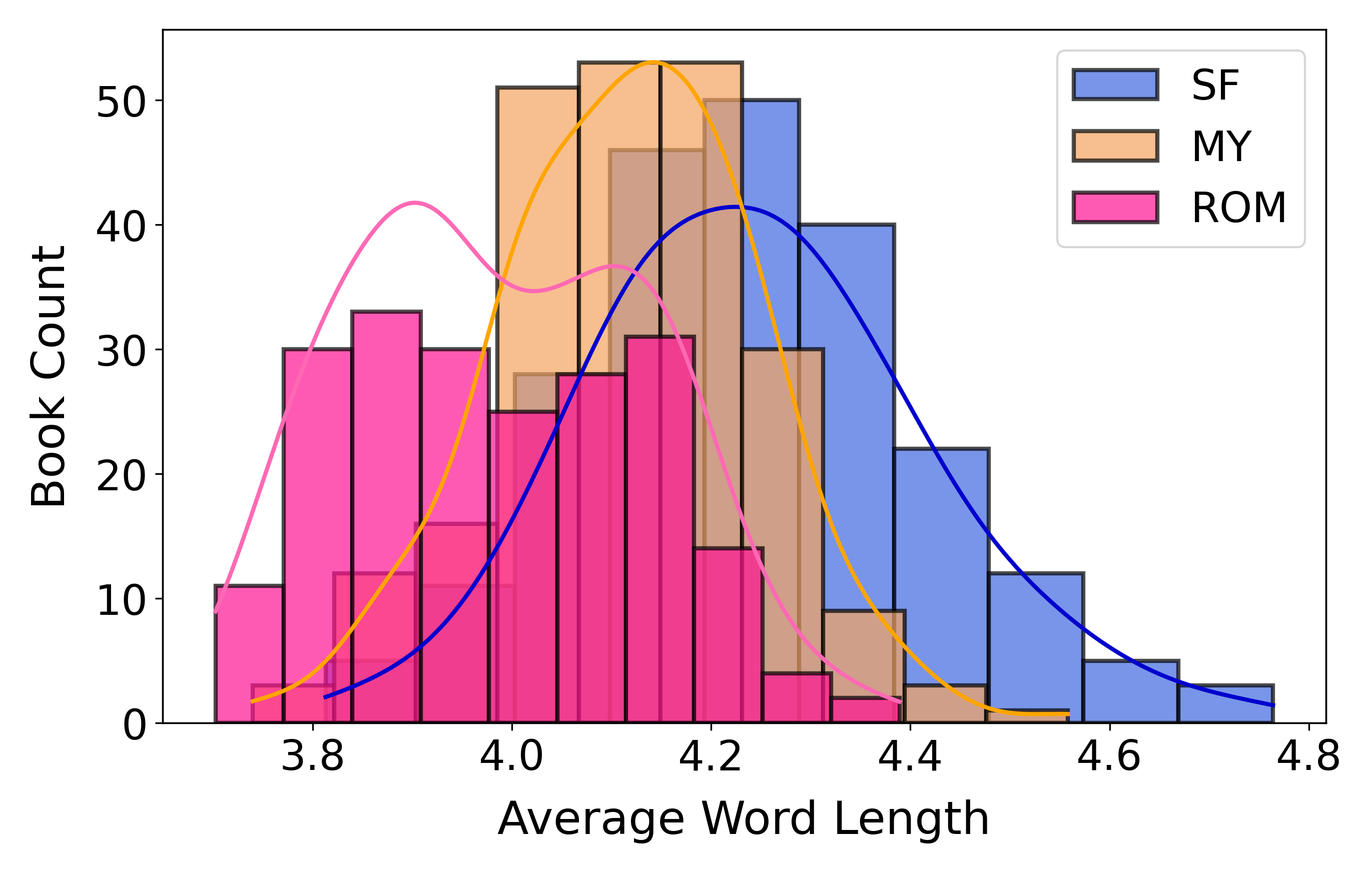}
  \includegraphics[width=0.45\linewidth]{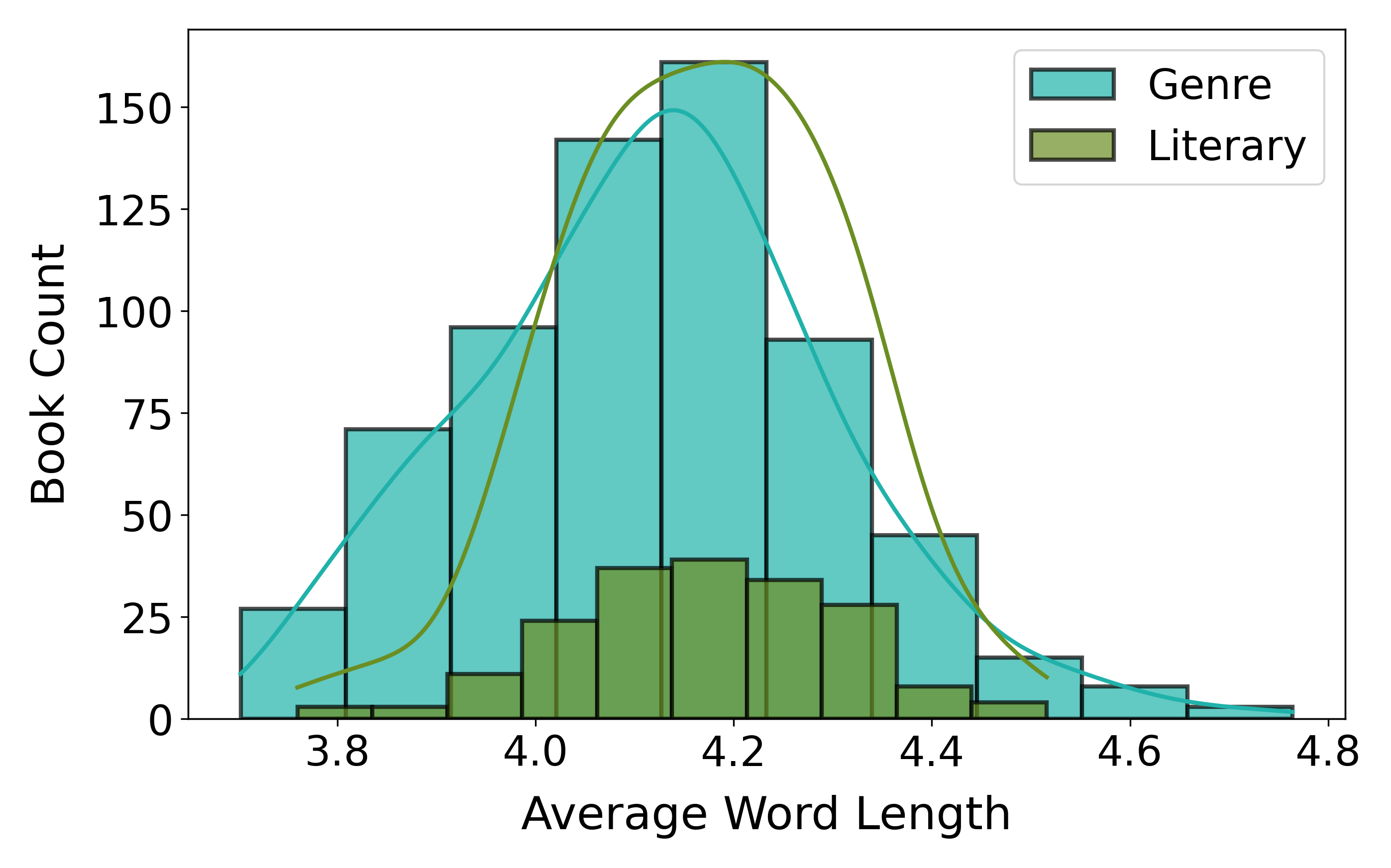}
  \centering
  \includegraphics[width=0.45\linewidth]{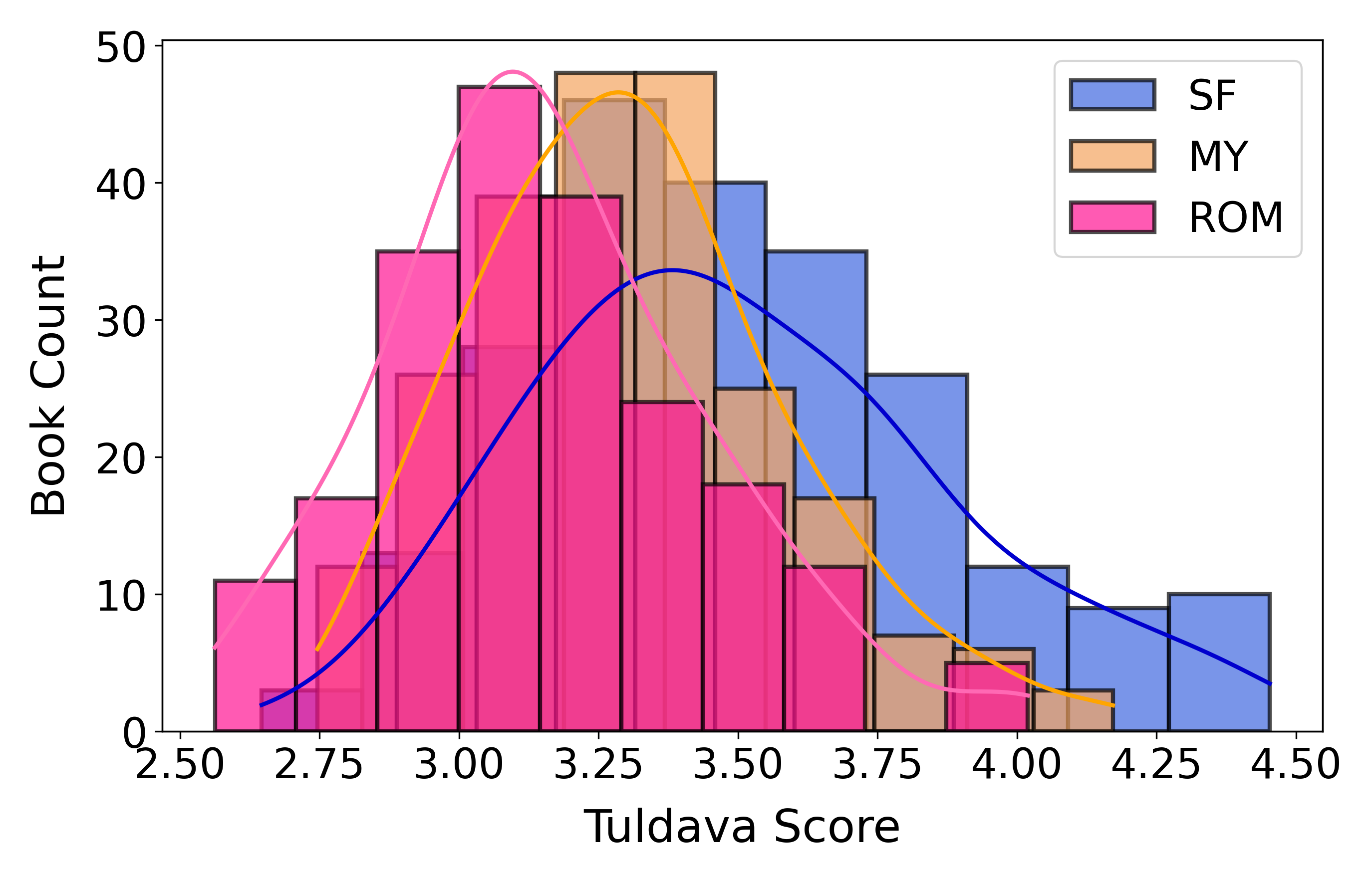}
  \includegraphics[width=0.45\linewidth]{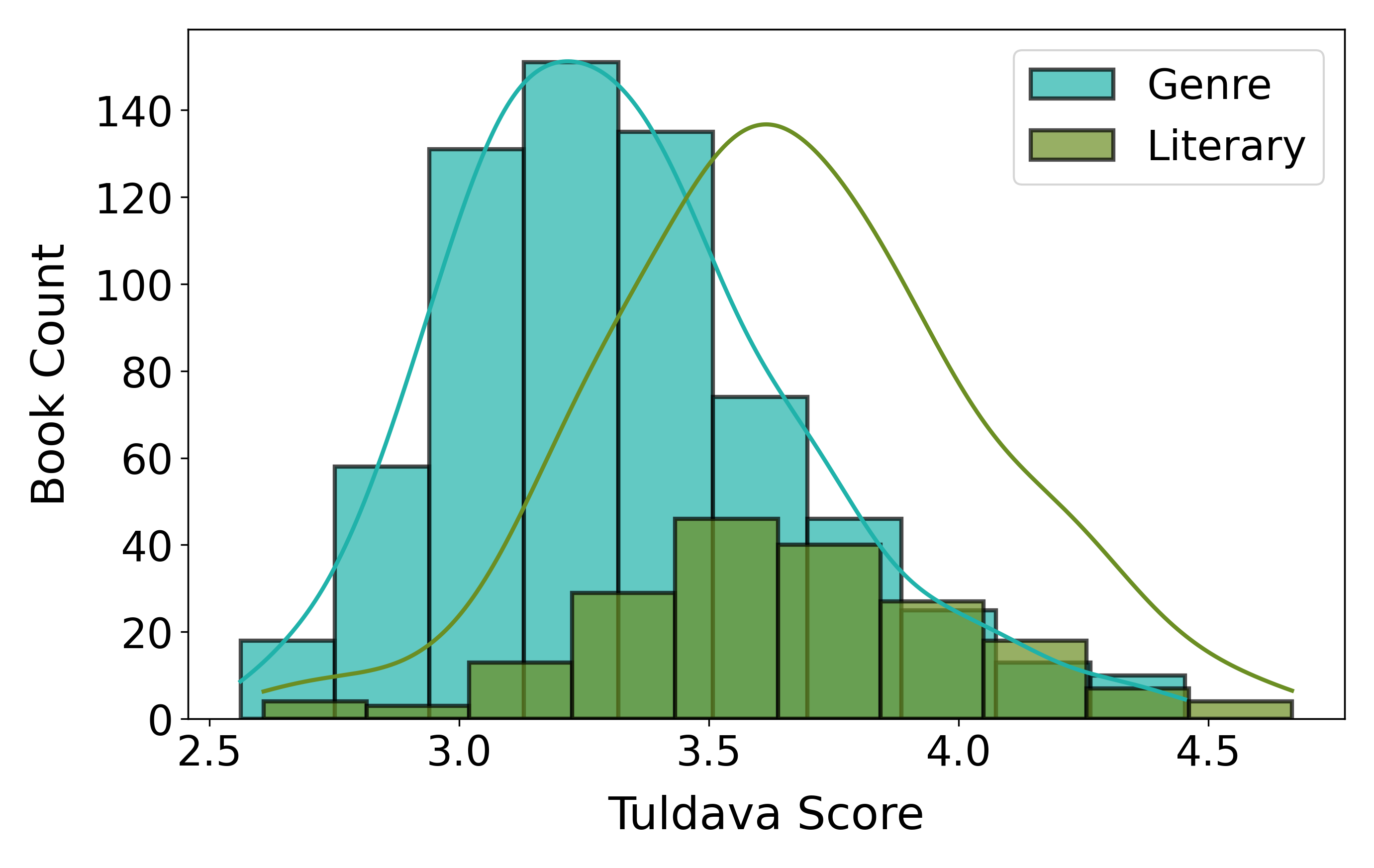}
  \centering
  \includegraphics[width=0.45\linewidth]{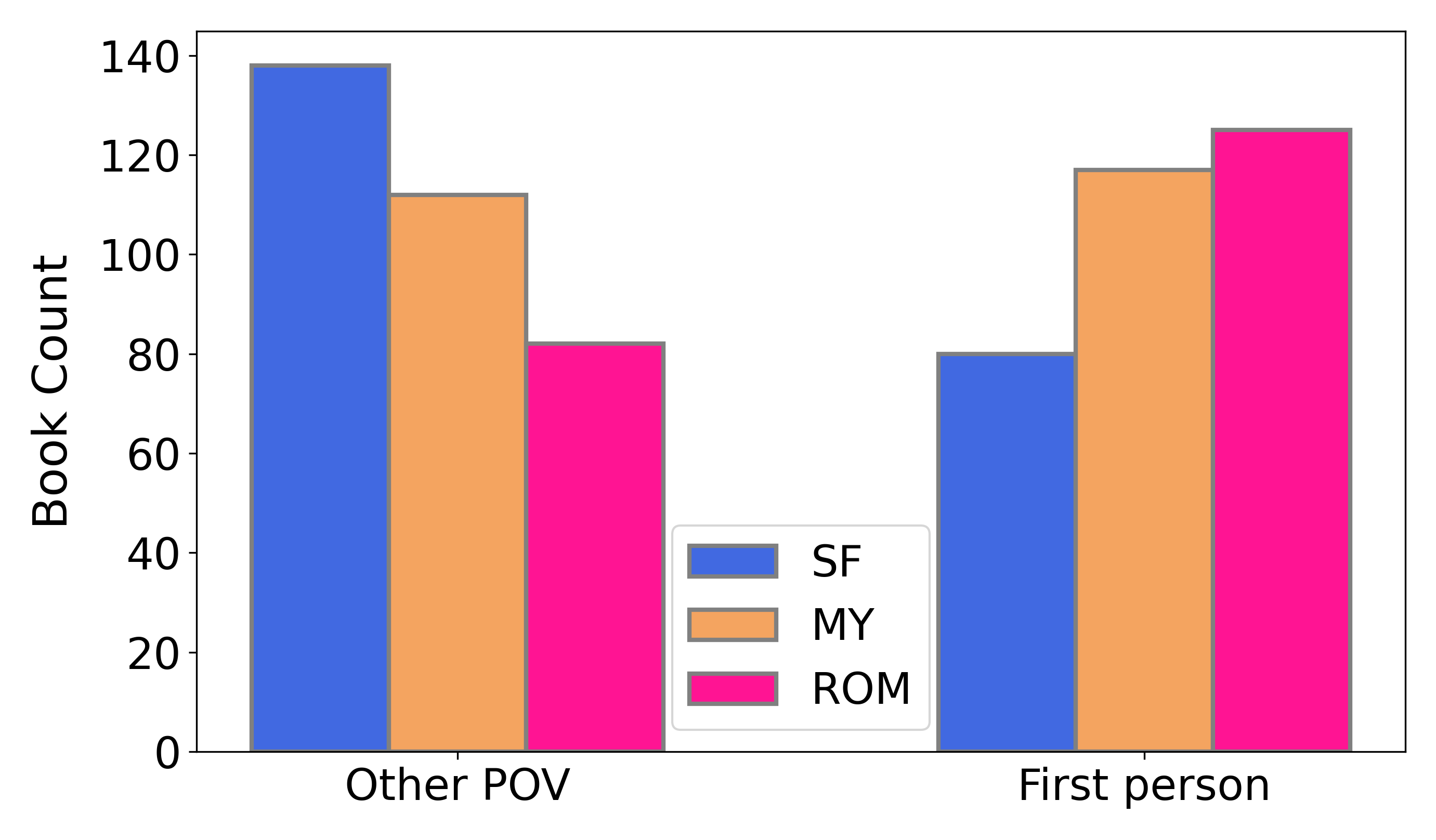}
  \includegraphics[width=0.45\linewidth]{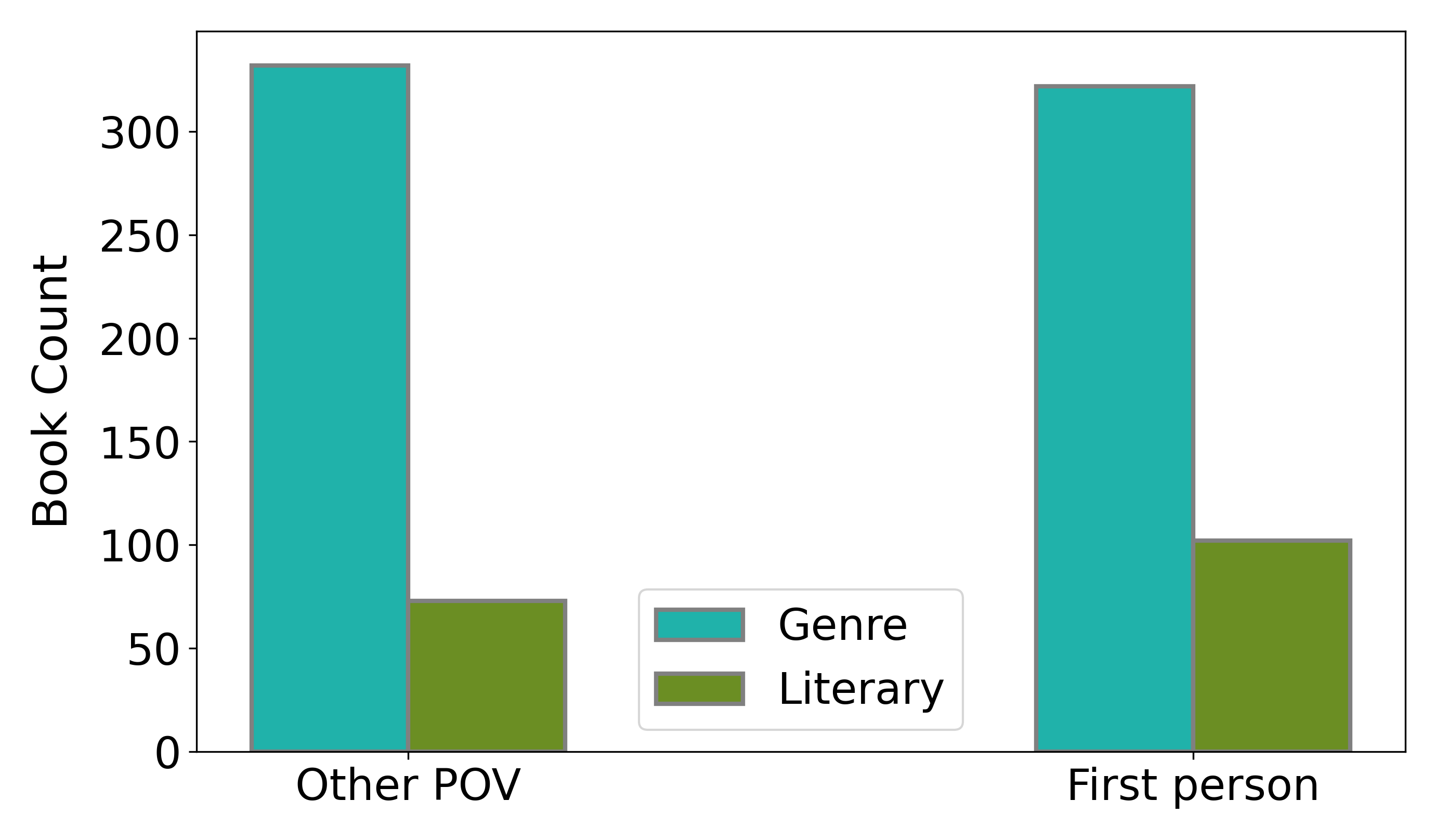}
    \caption{Language histograms}
  \label{fig:langhist3}
\end{figure}

\section{Additional Logistic Regression Output}
\label{appdx:logistic_output}
\begin{table}[H]
  \centering 
  \small
  \begin{tabular}{lc}
    \toprule
    \textbf{Feature} & \textbf{VIF} \\
    \midrule
    Narrative Pace & 4527199.22 \\
    Narrative Distance & 3408000.66 \\
    Circuitousness & 1867745.32 \\
    Topical Heterogeneity & 37.04 \\
    Event Count & 4.61 \\
    Character Count & 2.88 \\
    Protagonist Conc. & 1.41 \\
    Written in 1st Person & 1.57 \\
    Token Count & 23.96 \\
    Average Word Length & 11.64 \\
    Average Sentence Length & 24.65 \\
    Tuldava Score & 43.27 \\

    \bottomrule
  \end{tabular}
  \captionof{table}{Variance Inflation Factors (VIFs) for features, rounded to two decimal places.}
  \label{tab:vif_features_rounded}
\end{table}

\begin{table} [H]
  \centering
  \small
  \begin{tabular}{lcccc}
    \toprule
    \textbf{Feature} & \textbf{Feature Coeff} & \textbf{Feature} $\boldsymbol{p}$ & \textbf{Interaction Coeff} & \textbf{Interaction} $\boldsymbol{p}$\\
    \midrule
    Narrative Pace & -0.54 & 0.00 & 0.60 & 0.00 \\
    Narrative Distance & -0.05 & 0.67 & 0.66 & 0.00 \\
    Circuitousness & -0.74 & 0.00 & 0.04 & 0.82 \\
    Topical Heterogeneity & 0.16 & 0.15 & 0.36 & 0.04 \\
    Event Count & -0.51 & 0.00 & -0.04 & 0.87 \\
    Character Count & -0.03 & 0.78 & 0.43 & 0.01 \\
    Protagonist Conc. & -0.26 & 0.03 & -0.44& 0.03 \\
    Written in in 1st Person & 0.94 & 0.00 & -1.15& 0.00 \\
    Token Count & -0.23 & 0.05 & 0.27 & 0.15 \\
    Average Word Length & -0.29 & 0.02 & 1.03& 0.00 \\
    Average Sentence Length & 0.35 & 0.07 & -0.74 & 0.00 \\
    Tuldava Scores & 0.57 & 0.00 & 0.76& 0.00 \\
    \bottomrule
  \end{tabular}
  \caption{Output of univariate regression, rounded to two decimal places. $p$ < 0.05 indicates significance.}
  \label{tab:logistic}
\end{table}

\begin{table} [H]
\centering
\small
\begin{tabular}{lccccc}
\toprule
\textbf{Parameter} & \textbf{Coefficient} & $\boldsymbol{p}$ & \textbf{CI Lower} & \textbf{CI Upper} & \textbf{Odds Ratio} \\
\midrule
Intercept & -1.23 & 0.00 & -1.74 & -0.73 & 0.29 \\
Female Author & -0.53 & 0.13 & -1.22 & 0.15 & 0.59 \\
Protagonist Concentration & -0.36 & 0.02 & -0.66 & -0.06 & 0.70 \\
\hspace{1em}+ Female Author & -0.16 & 0.52 & -0.66 & 0.33 & 0.85 \\
Average Word Length & -0.69 & 0.00 & -1.08 & -0.30 & 0.50 \\
\hspace{1em}+ Female Author & 0.76 & 0.01 & 0.16 & 1.35 & 2.13 \\
Average Sentence Length & 1.11 & 0.00 & 0.76 & 1.47 & 3.05 \\
\hspace{1em}+ Female Author & -0.03 & 0.90 & -0.54 & 0.48 & 0.97 \\
Token Count & 0.19 & 0.47 & -0.32 & 0.69 & 1.20 \\
\hspace{1em}+ Female Author & 0.65 & 0.14 & -0.20 & 1.49 & 1.91 \\
Narrative Pace & 0.18 & 0.34 & -0.20 & 0.56 & 1.20 \\
\hspace{1em}+ Female Author & 0.10 & 0.74 & -0.46 & 0.65 & 1.10 \\
Circuitousness & -0.58 & 0.00 & -0.96 & -0.20 & 0.56 \\
\hspace{1em}+ Female Author & -0.03 & 0.93 & -0.59 & 0.54 & 0.97 \\
Written in 1st Person & 0.05 & 0.88 & -0.62 & 0.72 & 1.05 \\
\hspace{1em}+ Female Author & -0.24 & 0.64 & -1.25 & 0.76 & 0.78 \\
Event Count & -0.40 & 0.15 & -0.94 & 0.14 & 0.67 \\
\hspace{1em}+ Female Author & -0.48 & 0.26 & -1.33 & 0.36 & 0.62 \\
\bottomrule
\end{tabular}
\caption{Multivatiate logistic regression results with interaction terms, rounded to two decimal places. $p$ < 0.05 indicates significance.} 
\label{tab:regression_outputs}
\end{table}

\section{Omega Squared Calculations}
\label{appdx:omega}
$\omega^2$ estimates the proportion of variance in the pairwise distances explained by group membership. The $\omega^2$ values for our $T^2_w$  tests are calculated using the following formula: 

\[
\omega^2_{\text{Tw2}} = 
\frac{
    \text{SS}_{\text{between}} - \text{df}_{\text{between}} \cdot \frac{\text{SS}_{\text{within}}}{\text{df}_{\text{within}}}
}{
    \text{SS}_{\text{between}} + \frac{\text{SS}_{\text{within}}}{\text{df}_{\text{within}}}
}
\]

where: \\
\[
\begin{aligned}
& \text{SS}_{\text{between}} = SST - SSW, \\
& \text{SS}_{\text{within}} = SSW = \frac{ss2_{11}}{n_1} + \frac{ss2_{22}}{n_2}, \\
& \text{df}_{\text{between}} = 1, \\
& \text{df}_{\text{within}} = N - 2, \\
& N = n_1 + n_2
\end{aligned}
\]
\\
The $\omega^2$ values for our $W^*_d$  tests are calculated using the following formula: 
\[
\omega^2_{\text{WdS}} = 
\frac{
    \text{SS}_{\text{between}} - \text{df}_{\text{between}} \cdot \frac{\text{SS}_{\text{within}}}{\text{df}_{\text{within}}}
}{
    \text{SS}_{\text{between}} + \frac{\text{SS}_{\text{within}}}{\text{df}_{\text{within}}}
}
\]

where: \\
\[
\begin{aligned}
&\text{SS}_{\text{within}} = \sum_{i=1}^{k} \frac{ss2_{ii}}{n_i}, \\
& \text{SS}_{\text{between}} = SST - \text{SS}_{\text{within}}, \\
& \text{df}_{\text{between}} = k - 1, \\
& \text{df}_{\text{within}} = N - k, \\
& N = \sum_{i=1}^{k} n_i \\
\end{aligned}
\]

\end{document}